\documentclass[a4paper,fleqn]{cas-sc}

\usepackage[numbers]{natbib}
\usepackage{xcolor}
\usepackage{graphics} % for pdf, bitmapped graphics files
\usepackage{epsfig} % for postscript graphics files
\usepackage{times} % assumes new font selection scheme installed
\usepackage{amsmath} % assumes amsmath package installed
\usepackage{amssymb}  % assumes amsmath package installed

\usepackage{amsthm}

\usepackage{algorithmic}
\usepackage[ruled]{algorithm2e}
\newtheorem{definition}{Definition}
\usepackage{multirow}
\usepackage{adjustbox}
\usepackage{hyperref}
\usepackage{booktabs}
\usepackage{float}
\usepackage{flushend} 

\newcommand{\figref}[1]{Figure~\ref{#1}}

\def\tsc#1{\csdef{#1}{\textsc{\lowercase{#1}}\xspace}}
\tsc{WGM}
\tsc{QE}
\begin{document}
\let\WriteBookmarks\relax
\def\floatpagepagefraction{1}
\def\textpagefraction{.001}

\shorttitle{Real-Time Reachability for the Safety Assurance of Autonomous Vehicles}    

\shortauthors{Musau, Hamilton, Manzanas Lopez, Robinette, Johnson}  

\title [mode = title]{An Empirical Analysis of the Use of Real-Time Reachability for the Safety Assurance of Autonomous Vehicles}  

%\tnotemark[] 

%\tnotetext[<tnote number>]{<tnote text>} 


% \author[Vanderbilt University]{Patrick Musau}[]
\affiliation[inst1]{organization={Vanderbilt University Department of Electrical and Computer Engineering},%Department and Organization
            addressline={2201 West End Ave}, 
            city={Nashville},
            postcode={37235}, 
            state={TN},
            country={USA}}
            
\author[inst1]{Patrick Musau}[orcid=0000-0002-0227-1336]

\cormark[1]


\ead{patrick.musau17@gmail.com}

\ead[url]{http://pmusau17.github.io/}

\author[inst1]{Nathaniel Hamilton}
\author[inst1]{Diego Manzanas Lopez}
\author[inst1]{Preston Robinette}
\author[inst1]{Taylor T. Johnson}

%\credit{<Credit authorship details>}

% \affiliation[Vanderbilt University]{organization={},
% %          citysep={}, % Uncomment if no comma needed between city and postcode

% \author[Vanderbilt University]{Patrick Musau}[]

% % Footnote of the second author

% % Email id of the second author
% \ead{}

% % URL of the second author

% % Credit authorship

\cortext[1]{Corresponding author}



\begin{abstract}
Recent advances in machine learning technologies and sensing have paved the way for the belief that safe, accessible, and convenient autonomous vehicles may be realized in the near future. Despite tremendous advances within this context,  fundamental challenges around safety and reliability are limiting their arrival and comprehensive adoption. Autonomous vehicles are often tasked with operating in dynamic and uncertain environments. As a result, they often make use of highly complex components, such as machine learning approaches, to handle the nuances of sensing, actuation, and control. While these methods are highly effective, they are notoriously difficult to assure. Moreover, within uncertain and dynamic environments, design time assurance analyses may not be sufficient to guarantee safety. Thus, it is critical to monitor the correctness of these systems at runtime. One approach for providing runtime assurance of systems with components that may not be amenable to formal analysis is the simplex architecture, where an unverified component is wrapped with a safety controller and a switching logic designed to prevent dangerous behavior. In this paper, we propose using a real-time reachability algorithm for the implementation of the simplex architecture to assure the safety of a 1/10 scale open source autonomous vehicle platform known as F1/10. The reachability algorithm that we leverage (a) provides provable guarantees of safety, and (b) is used to detect potentially unsafe scenarios. In our approach, the need to analyze an underlying controller is abstracted away, instead focusing on the effects of the controller's decisions on the system's future states. We demonstrate the efficacy of our architecture through a vast set of experiments conducted both in simulation and on an embedded hardware platform.
\end{abstract}


\begin{highlights}
\item Providing provable assertions of safety at runtime is crucial for autonomous systems.
\item Autonomous systems are tasked with operating in dynamic and uncertain environments.
\item Reachability analysis is adept in analyzing the correctness of uncertain systems.
\item The Simplex Architecture can be used to assure the safety of complex components.
\item Simplex approaches can be used to assure the safety of machine learning components.
\end{highlights}

% Keywords
\maketitle%
\section{Introduction}
\label{sec:aij_introduction}

For decades, the vision of deploying autonomous vehicles ubiquitously has enraptured technology enthusiasts, researchers, and corporations. The prevailing conviction is that there are relatively few technologies that hold as much promise as autonomous vehicles (AVs) in bringing about safe, accessible, and convenient transportation. Despite demonstrated success through efforts such as the Defense Advanced Research Projects Agency's (DARPA) Grand Challenges \cite{driverlesschallenges,DARPA_Challenge}, and the emergence of high profile autonomous vehicle companies such as Alphabet's Waymo, Argo AI, Aptiv, Zoox, General Motors' Cruise, Tesla, Aurora, and Intel Corporation's Mobileye, the consensus remains that there are serious technical and safety challenges to be resolved.

The two fundamental challenges widely regarded as limiting the arrival and widespread adoption of AVs are safety and reliability \cite{Majumdar2017}. Reasoning about safety requires an understanding of the joint dynamics of computers, networks, and physical dynamics in uncertain and variable environments, making it a notoriously difficult problem \cite{Yurtsever2019}. To handle the complexities of their environments, many AVs make use of \emph{Machine Learning} (ML) components such as artificial neural networks to decipher the information observed from an ever-evolving configuration of on-board sensors \cite{Yurtsever2019}. However, despite the impressive capabilities of these components, there are reservations about using them within safety critical settings. This is primarily due to the difficulty of interpreting the inner workings of these models, which prevents any meaningful explanation of their behavior from being made.

Utilizing a model that lacks transparency with respect to how its decisions are made within a system that is safety critical, constitutes the highest form of technical debt \cite{Sculley2015} and, as a result, the last several years have witnessed a significant increase in developing methods that seek to reason about the safety and robustness of machine learning methods \cite{VariationalMIT2018,Liu2019,xiang20118survey}. Unfortunately, while numerous works have been proposed over the past few years for the formal analysis of machine learning methods, the vast majority of these efforts have not been able to scale to the complexity found in real world applications, where models, such as neural networks, may contain millions or even billions of parameters \cite{BallesterGoogLeNet,SimonyanVeryDeep}. Further exacerbating this challenge are the unanticipated environmental conditions that cannot be captured at design time. Testing, while rather effective, is also infeasible, as this requires a prohibitively large amount of tests to be executed in order to demonstrate a sufficient amount of reliability \cite{Beg2017}. 

Since design-time testing and formal analysis are not sufficient alone to demonstrate the safety of complex systems, verifying systems at runtime is often required. These regimes are classically referred to as \textit{runtime verification} or \textit{runtime assurance} approaches, and they broadly consist of observing the execution of a system at operation-time and checking whether relevant safety properties are preserved. The system model under consideration may take on many forms, including models of the physical dynamics of the system or even models of the underlying software governing its behavior. These models are then used to consider lightweight yet rigorous considerations of presupposed formal properties \cite{PETTERSSON200573,DesaiRV2017,Deshmukh2015,Masson2018,dunlap2021Safe}.

A crucial question that must be answered when a runtime assurance approach to verification is used, is what happens when a problem is discovered by a monitor \cite{Sokolsky2012}. Many runtime verification approaches classically involve the ability of invoking recovery actions in response to safety or property violations. Within this context, one of the most popular runtime assurance architectures is the \textit{Simplex Architecture}, and it has demonstrated significant success in enabling the assurance of systems with components that may be too complex, or too large, for complete design-time analysis \cite{SetoSimplex}. This makes this regime particularly attractive for the assurance of machine learning and AI-based components. In this framework, an unverified component is wrapped with a safety controller and a switching logic designed to transfer control to the safety controller in the event of property or safety violations \cite{Bak2014}. A useful analogy for this architecture is a driving instructor's car with two steering wheels and two sets of brakes. As long as the instructor is capable of intervening in dangerous situations, the capricious student is allowed to drive \cite{Bak2014}. 

Typically, in simplex architectures, the switching logic is primarily designed either from a control theoretic perspective through the solution of linear matrix inequalities (LMI) \cite{SetoCaseStudy2000}, or using a formal analysis hybrid-systems reachability technique \cite{Bak2009Simplex}. As Bak et al. note, it is easy to design a safe decision logic; one can simply always use the safety controller \cite{Bak2014}. However, this is unsatisfactory since the performance responsibilities of the system might be forfeited or unreasonably delayed \cite{Bak2014}. The key challenge in this regime is to design a switching logic that allows the dynamic capabilities of the unverified complex controller to be exploited as much as possible without compromising safety. 

In this paper, we extend the real-time reachability algorithm from \cite{Bak2014,Johnson2016} to design a simplex architecture for a 1/10 scale autonomous racing car called the F1/10 platform. The central idea behind our framework lies in computing the set of reachable states of the F1/10 system and ensuring that it never enters unsafe states as it navigates an environment. Specifically, this entails checking that the vehicle's trajectories are free from collisions with both static and dynamic obstacles within its environment. Rather than performing an analysis of the underlying controller governing the behavior of the system, the crux of our approach lies in monitoring the influence of a controller's decisions on the overall evolution of the system by reasoning about the set of reachable states over a finite-time horizon. This set can then be used to forecast potential collisions. Thus, this safety checking procedure forms the basis of the switching scheme in our simplex architecture. In our work, the nature of the underlying controller is immaterial, and our experiments consider a variety of control strategies ranging from machine learning, path tracking, and gap following regimes. 

One of the key benefits of utilizing reachability analysis to construct our simplex architecture is that reachability analysis is quite adept at handling uncertainty. This makes the approach particularly attractive for autonomous systems, whose correct operation is dependent on an effective treatment of uncertain sensor measurements, predictions about the behavior and intent of dynamic environmental participants, and the modeling assumptions defining its control regimes. Bearing the above in mind,  reachability techniques can be used to compute \textit{all} the possible states that a system may attain from large bounded sets of initial conditions, disturbances and system parameter variations  \cite{Asarin2007}. Thus, as Asarin et al. note in their work, such an analysis "provides knowledge about the system with a completeness or coverage that a finite number of simulations cannot deliver," \cite{Asarin2007}. We evaluate the merits of using reachability methods  for the safety assurance of the F1/10 platform both in simulation and on an embedded hardware platform using a variety of controllers, number of obstacles, and runtime configurations. Furthermore, we present an analysis of the effects of various sources of uncertainty on the conservativeness of our safety regime. Finally, we also present a robust runtime characterization of the real-time reachability regime that forms the basis of our work.

\subsection{Statement of Contributions}

The contributions of this article can be summarized as follows. (1) We present a runtime verification technique that abstracts away the need to analyze the nature of the underlying controller governing the behavior of our system, and instead focuses on the effects of the control decisions of these controllers on the overall system during operation. This is accomplished by obtaining the set of reachable states of the system over a finite time horizon and checking for potential collisions with objects within the environment. The approach presented in this paper has the ability to reason about safety in the presence of both static and dynamic obstacles. (2) Leveraging the reachability framework, we implement a simplex control architecture in order to maintain safety during operation. (3) We present a safety analysis of a diverse set of controllers through a series of experiments with varying speeds and number of opposing vehicles in order to explore the tradeoffs of our approach. (4) We present a rigorous empirical analysis, accounting for various classes of uncertainty within our regime. Our analysis includes a study of the effects of uncertainty on the conservativness of our safety regime. (5) Finally, we present a runtime characterization of the real-time reachability regime, enabling our work over a broad set of experiments. 

A preliminary version of this work appeared at the 2022 IEEE Conference on Assured Autonomy (ICAA). In this enhanced and extended version, we provide the following additional contributions (1) An extension of our safety framework to account for dynamic obstacles, (2) further commentary on the physical dynamics models used within our approach and a deeper discussion of handling uncertainty, (3) an empirical study of the effects of uncertainty on our overall safety regime, and (4) additional experimental results including an evaluation of our approach in contexts where two 1/10 scale embedded open-source autonomous vehicle testbeds interact within a racing context.

\section{Related Work}

The increasing ubiquity of software in numerous domains, particularly in safety critical domains, has heightened the need to ensure the correct and reliable operation of deployed software. Over the last several years, there has been a wealth of approaches proposed towards proving the correctness of autonomous systems prior to fielding them in their respective operational design domains \cite{Maler14,Tomlin2003,Doyen2018,Yang2017,Clark2013,Kwon2018}. However, very few approaches exist that can provide strict formal guarantees about their behavior, due to the challenges associated with reasoning about large and complex systems that are tasked with operating in dynamic and uncertain environments. Moreover, these systems often leverage machine learning components to deal with the diverse data obtained from the system's sensors. Applying formal assurance techniques to machine learning systems has only been considered recently and poses unique challenges \cite{UrbanFormalMethodsML2021}. While there has been significant progress within this realm, there is still a significant gap between the machine learning models that these approaches can handle, and the models deployed in state-of-the-art systems.

In a similar vein, while significant efforts have been devoted to developing approaches that can deal with the complexities of autonomous systems, only a few of them can be leveraged at runtime \cite{Angelo2020,Bu2011,BU2020,Lin2020,Ti2014,Bajcsy2019Provably,holmes2020reachable,Althoff2014,LeungReach2020,Huang2014,Desai2018,Desai2017,mitsch}. The motivation for utilizing runtime assurance approaches stems from a recognition that in certain environments, complete or even partial verification may be infeasible due to the well-known state-explosion problem \cite{Valmari1998} and the reality that the complete analysis of certain components may be infeasible at design time. As an example, for an autonomous vehicle, it is imperative that collisions are avoided while the system carries out its high-level goals. This requires monitoring the vehicle's state during operation, as design time considerations cannot feasibly consider all the possible scenarios that the system may encounter \cite{Butler1993}. In light of these challenges, this has led to the rise of  \textit{runtime assurance}, \textit{runtime-verification} and \textit{simplex} strategies, and it is within this context that we present our work.

Runtime assurance (RTA) and runtime verification (RV) methods broadly consist of techniques that allow for observing the execution of a system at runtime and checking whether relevant correctness properties are preserved. An intuitive explanation of these regimes can be summarized as follows. RTA techniques are tasked with ensuring the correct operation of systems with untrusted components, while runtime verification techniques monitor a system against presupposed formal properties at runtime \cite{PETTERSSON200573,DesaiRV2017,Deshmukh2015,Masson2018,Akametalu2014,mitsch,Daws1998,Phan2020}. The distinction here is that while runtime assurance techniques may often utilize verification results, they may often also employ statistical techniques such as anomaly detection \cite{boursinos2020trusted} or simulation based strategies \cite{TranSimulation2019}. One such example is the work by Allen et al.\cite{Allen2014}, in which they utilize various machine learning techniques, mainly support vector machines and linear regression models, to approximate the solution to the real-time computation of the set of reachable states of an underlying system. Their work is capable of being applied in low-resource, real-time environments, but suffers from the downside that theoretical guarantees cannot be made using these techniques. Only statistical guarantees may be obtained. However, the approach demonstrated impressive results in improving state-of-the-art execution times in the assurance process by four orders of magnitude.

Within the context of runtime-assurance approaches, the simplex architecture has been used widely in the research literature to provide guarantees for systems with unverified components. The contexts in which it has been applied to include aerospace systems \cite{SetoSimplex,SetoCaseStudy2000,dunlap2021Safe}, fleets of remote controlled cars \cite{Crenshaw2007}, industrial embedded infrastructure \cite{Bak2009Simplex,Yang2017}, and distributed mobile robotics applications \cite{Desai2018,Tran2020}. In \cite{Lin2020}, the authors utilize a reachability regime to guarantee the safety of an autonomous vehicle that makes use of a reinforcement learning controller for a way-point-following task. Most similar to this work in \cite{Tran2020}, the authors utilize a real-time reachability approach to verify that a group of quadcopters executing a distributed search mission is free from collisions. Their approach is implemented in simulation and can theoretically deal with over 64 quadcopters. These works primarily deal with developing provably correct motion planners, while the focus of the work presented in this paper, is abstracting away the nature of an underlying controller, and instead reasoning about the consequences of its actions on overall system safety.

Finally, in recent years, researchers have begun to integrate traditionally non-real-time verification approaches within real-time systems by making these approaches more amenable to real-time execution. These include viability kernel approaches that determine if a set of states remain within a predefined region \cite{Gurriet2018,Althoff2014}, as well as Hamilton-Jacobi reachability (HJR) techniques that can deal with dynamical systems with general nonlinear dynamics in uncertain environments \cite{Herbert2019,Bajcsy2019Provably,bansal2020hamiltonjacobi,Fisac2017,Chen2016,dhinakaran2017hybrid}. One of the major benefits of Hamilton-Jacobi reachability is that it allows for the specification of an optimal control problem characterized by a differential game, where a controller must maintain system safety under the influence of disturbances \cite{Akametalu2018}. Therefore, this framework allows for a robust analysis of safety under uncertainty. Specifically, \cite{Bajcsy2019Provably} and \cite{Akametalu2014} were able to implement these techniques on both simulation and hardware platforms. However, both papers used their respective reachability results for safe motion planning, and their approach does not possess rigorous real-time guarantees. The work contained in this paper, however, primarily deals with the construction and implementation of a safety assurance architecture based through use of a real-time reachability regime.

What distinguishes the methods presented in this article is that they possess real-time guarantees and have been extended from \cite{Johnson2016} to consider the safety of a set of controllers under varying levels of uncertainty. Moreover, this work considers the safety question in the presence of dynamic and static obstacles and is evaluated over a wide range of experiments. We realize that there are numerous other interesting works present within this area, and to the best of our knowledge, the aforementioned works are the most relevant to the methods presented within this paper.   

\section{Preliminaries}
\label{sec:aij_Background}
\subsection{The Simplex Architecture}

As modern autonomous systems grow in complexity, so do the challenges in assessing their reliability and correctness \cite{Bak2009Simplex}. Moreover, any arguments about the reliability and safety of the system rely on assertions about the individual components that make it up \cite{Bak2014}. However, in recent years, with the growth of increasingly autonomous systems \cite{Jha2020}, individual components may be designed using machine learning methods, such as neural networks, that are opaque to traditional formal analysis. Despite the recent years' surge in the development of formal analysis techniques for these types of models \cite{Liu2019,xiang20118survey}, most techniques are incapable of dealing with the scale of models deployed in state-of-the-art systems.

One paradigm for dealing with untrustworthy components is the \emph{simplex architecture} \cite{SetoSimplex, Bak2014, SetoCaseStudy2000}. In the simplex architecture, the unverified component, or \emph{complex controller}, is wrapped with a \emph{safety controller} and a switching logic used to ensure safety~\cite{Bak2014}. A useful analogy for this architecture is a driving instructor's car with two steering wheels and two sets of brakes. As long as the instructor is capable of intervening in dangerous situations, the capricious student is allowed to drive. Typically, the complex controller has better performance with respect to the design metrics, whereas the safety controller is designed with simplicity and verifiability in mind. Thus, by using this architecture, one can utilize the complex controller while still maintaining the formal guarantees of the safety controller. The key challenge when designing a system with the simplex architecture is properly designing the switching logic \cite{Johnson2016}. One must be able to clearly delineate safe states from unsafe states. 

In a typical implementation of the simplex architecture, the switching logic is primarily designed either from a control theoretic perspective through the solution of \emph{Linear Matrix Inequalities} (LMI) \cite{SetoCaseStudy2000}, or using a formal analysis hybrid-systems reachability technique \cite{Bak2009Simplex}. In this paper, our simplex design requires computing the set of reachable states online through the use of a real-time reachability algorithm for short time horizons.

\begin{figure}[!htbp]%
  \centering
  \includegraphics[width=0.9\linewidth]{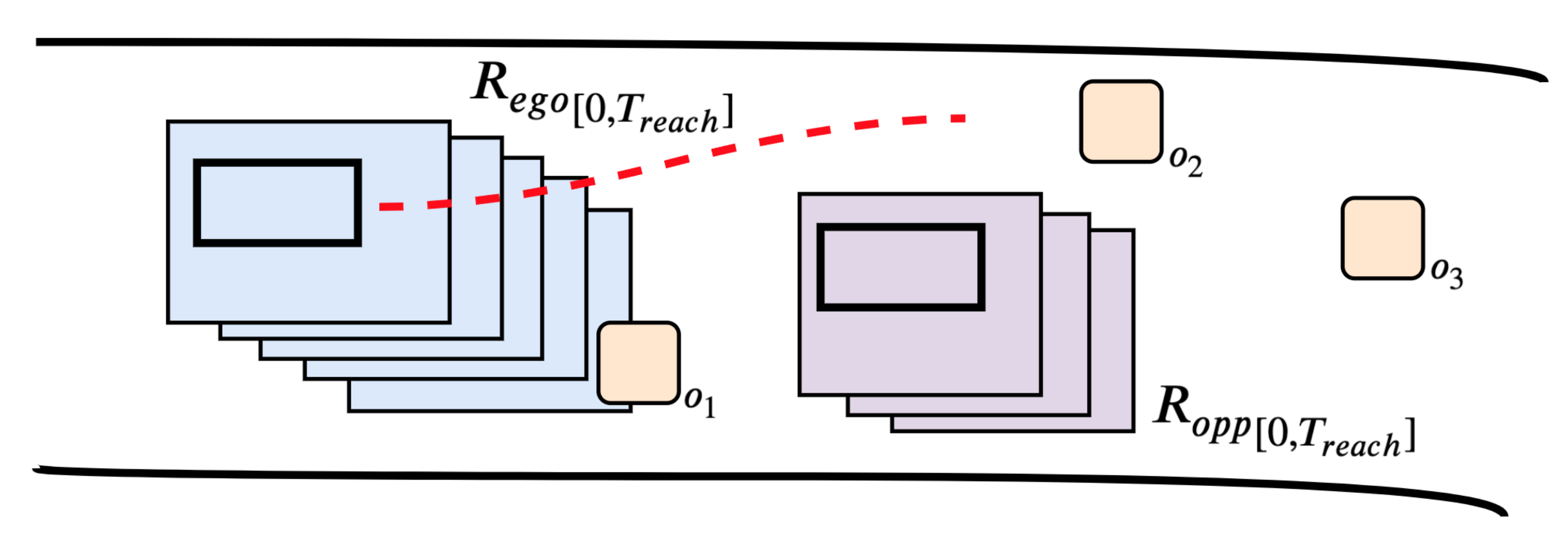}
  \caption{Overview of our runtime safety assurance framework. In this figure, the blue rectangles correspond to the reachable set of the ego vehicle, while the purple rectangles correspond to the reachable set of a dynamic opponent. Static obstacles are shown in orange, and the racetrack boundaries are the curved solid black lines. The red dotted line corresponds to the trajectory that would be obtained through the exclusive use of the safety controller. In the above figure, the reachable set of the ego vehicle, $\mathnormal{R_{ego}}_{[0,T_{reach}]}$, projects the effects of using a control action issued by the complex controller leveraged by the system, while the reachable set of the dynamic opponent is obtained by assuming that the opponent vehicle will maintain its velocity and direction over a short time horizon $\mathnormal{R_{opp}}_{[0,T_{reach}]}$. If the reachable set of the ego vehicle intersects with any obstacle, $o_1$, in the environment, or with the reachable set of an opponent vehicle, then our simplex approach switches to using a safety controller optimized to avoid collisions (red trajectory). }
  \label{fig:aij_overview}
\end{figure}%

\subsection{Reachability Analysis}

Reachability analysis is a model checking technique that involves rigorously computing the set of all states that a system can attain over a finite time horizon, and it is commonly used as a method for demonstrating that a system satisfies relevant safety properties \cite{Asarin2007}. One of the major strengths of these approaches is they are able to provide knowledge about an underlying system with a level of completeness that a finite number of simulation analyses cannot deliver \cite{Asarin2007}. Primarily, because the set of reachable states obtained using these approaches can describe the system's trajectories from all possible initial conditions, and under all admissible disturbances and variations in the parameter values of the underlying model \cite{Asarin2007}. In deriving such a set, the safety assurance problem often consists of determining whether there is an intersection between the reachable set of a system and a set of \textit{undesirable states}. As an example, for an autonomous vehicle this analysis can be leveraged to investigate whether the vehicle remains within lane boundaries, and if static and dynamic obstacles are avoided as the vehicle navigates within its environment \cite{Althoff2014}.

Generating the set of reachable states involves a combination of numerical analysis techniques, graph algorithms, and computational geometry \cite{Asarin2007,Asarin2003} and there is a rich set of literature and software tools available for the reachability analysis of systems with continuous, discrete, and hybrid dynamics. While, in this article, we confine our focus to those with continuous dynamics, the reachability analysis of discrete systems has been extensively considered since the early 1960s and is a well studied problem \cite{Desoer,Rakovic} 

\begin{figure}[!htbp]%
  \centering
  \includegraphics[width=\linewidth]{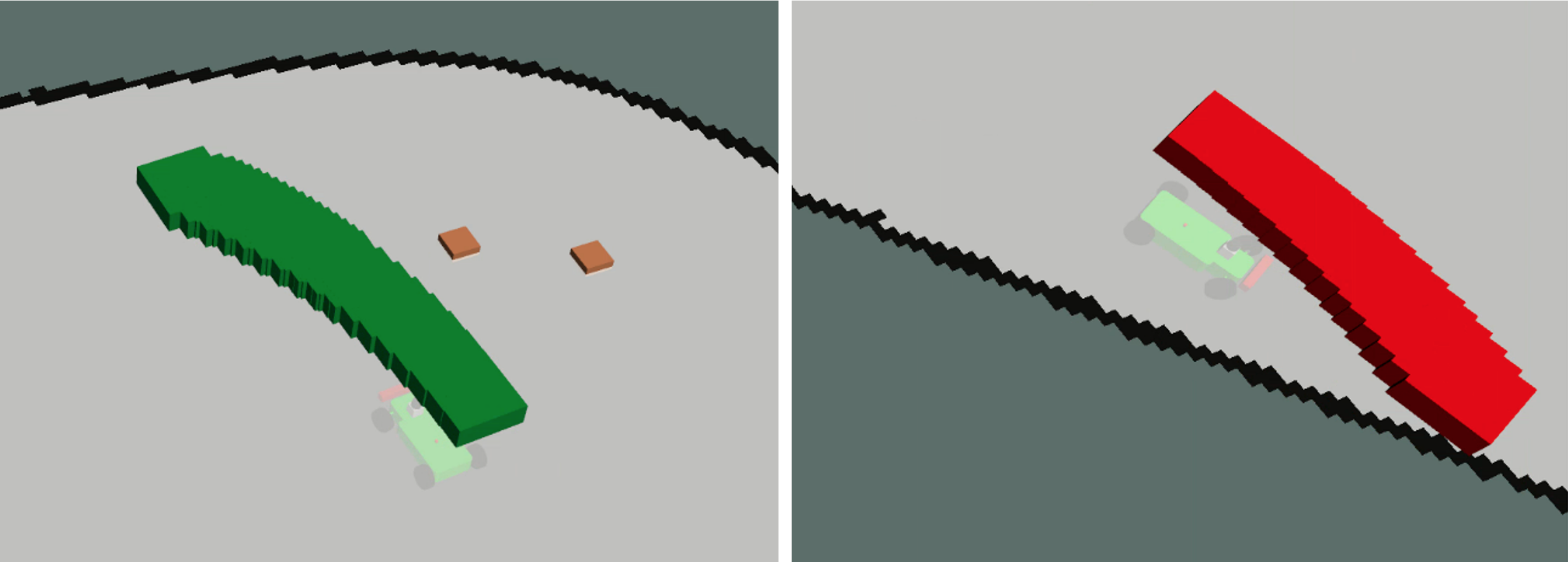}
  \caption{Visualization of the set of reachable states derived by projecting a control action forward over a finite time horizon in our simulation evnironment. For illustration purposes, we display only a subset of the hyper-rectangles in the above images. \textit{Left}: (green boxes) Example of an action labeled as safe, since there are no intersections between the reachable set and obstacles in the vehicle's environment or the racetrack walls (black). \textit{Right:}  (red boxes) This example corresponds to an unsafe scenario, as following the issued control action would result in a collision between the vehicle and the racetrack boundaries. In the above images, the orange squares represent the location of cones and their corresponding bounding box.}
  \label{fig:aij_reachset}
\end{figure}%

Traditionally, reachability methods have been executed offline, at design-time, because they are computationally intensive endeavors \cite{AlthoffCORA2015,Asarin2007,Chen2012}. However, in \cite{Bak2014,Johnson2016}, Bak et al. and Johnson et al. present a reachability algorithm based on the seminal mixed face-lifting algorithm \cite{dang2000}, capable of running in real-time on embedded processors. The algorithm is implemented as a standalone C-package that does not rely on sophisticated (non-portable) libraries, recursion, or dynamic data structures and is amenable to the anytime computation model in the real-time scheduling literature. In this regime, each task produces a partial result that is improved upon as more computation time is added \cite{Johnson2016}. The standalone nature of this approach allowed us to utilize this scheme in implementing our safety architecture on an embedded hardware platform.

\subsection{Safety Architecture}

The controllers in our experiments are designed to sample sensor data and compute control actions at fixed time intervals, as typically done in the control community. During each control period, we take the corresponding control action and compute the reachable set of states into the future as defined by the current state, assumptions around disturbance and uncertainty, and a specified finite-time horizon. An example of this computation is shown in \figref{fig:aij_reachset}. We assume a fixed control action throughout the reachable set computation.\footnote{We discuss the merits of this assumption in Section \ref{sec:aij_discussion_limitations}.} Based on the obtained reachable set, we determine if the system will collide with objects in its environment and, if necessary, switch to a safety controller optimized for obstacle avoidance. If the system falls back to using a safety controller, we only allow a switch back to the complex controller if the complex controller has demonstrated safe behavior for a fixed number of control periods.\footnote{In our experiments, we allowed a switch back to the safety controller after 30 control periods. This corresponds to 1.5 seconds using a $20Hz$ control period.} This prevents arbitrary switching and incorporates a sense of hysteresis into our control strategy. Additionally, by not switching back until consistently safe behavior has been demonstrated, we enforce a notion of dwell time, which reduces instabilities caused by switching too frequently. An overview of our approach is shown in Figure~\ref{fig:aij_overview}.

\subsection{Handling Uncertainty}

\label{sec:aij_uncertainty_analysis}

As with all model-based approaches, the quality of our safety declarations, and thereby the results of our reachability computations, are highly dependent on the quality of the models of the underlying system and environment. It is imperative that our derived model is a good representation of reality. Otherwise, any predictions about the behavior of our system may be invalid. The reality, however, is that deriving good representations of each of these elements is quite challenging \cite{SeshiaTowards2016}. Developing an exact model of a system, for example, is extremely difficult due to the presence of complex physical interactions that may be hard to describe precisely \cite{Akametalu2018}. These interactions include phenomena such as drag forces, non-observable external disturbances, friction, non-deterministic model parameters, non-observable states, and other stochastic elements \cite{Akametalu2018}. 

In many cases, things that cannot be modeled explicitly are often aggregated together as uncertainty, and rigorous analyses aimed at quantifying levels of uncertainty in the underlying model are frequently conducted \cite{Akametalu2018,ORELLANA2021589}. Sources of uncertainty within a system can broadly be classified into two categories, \textit{Aleatoric Uncertainty}, and \textit{Epistemic Uncertainty}. Aleatoric uncertainty, also referred to as irreducible random uncertainty, is characterized by the natural variation of physical systems due to random effects \cite{DotyUncertainty}. Epistemic Uncertainty, however, refers to systematic uncertainty, that is attributed to a lack of knowledge of the dynamics of the system \cite{Eyke2021}. As an example, epistemic uncertainty can be attributed to a lack of knowledge of how to model quantities that are hard to measure or represent, and unlike aleatoric uncertainty, it can be reduced by more comprehensive experimentation and modeling \cite{Eyke2021}. 

Before making use of a model within model-based design approaches, it is imperative to identify and define assumptions about all the known sources of uncertainty in the system, and if through rigorous analyses one can obtain bounds on the uncertainty associated with a model's parameters, then one can conduct exhaustive analyses of the behavior of the system under a large set of bounded parameter variations and initial conditions \cite{Beg2017}. In doing so, the system designer can gauge whether the system satisfies the required levels of consistency, robustness, and quality governed by its requirements under the assumptions about the underlying levels of uncertainty \cite{ORELLANA2021589}. 

While stochastic simulation techniques, such as the Monte Carlo Paradigm, have allowed for efficient explorations of the parameters of formal models, the number of simulations needed to gain full confidence in an underlying system is prohibitively large \cite{Beg2017}. As Beg et al. note, in general, the total number of Monte Carlo Simulations aimed at matching a model to experimental data would need to be increased one hundred-fold to achieve an additional decimal place of precision. Thus, theoretically, having total confidence in the results of Monte Carlo analyses would require an infinite number of simulations \cite{Beg2017}. 

Within this context, set-based reachability regimes have displayed significant success in rigorously capturing the behavior of a system under a large set of initial conditions, disturbances, and system parameter values \cite{Asarin2003}. Thus, reasoning the correctness of the model can be reduced to verifying whether the set of all reachable states satisfy key properties across all possible model parametrizations \cite{Beg2017}.

\section{Problem Formulation}
\label{sec:aij_problem_formulation}

This section reviews modeling dynamical systems, and then presents this work's main contribution, a runtime assurance framework leveraging real-time reachability and the simplex architecture.

\subsection{System Dynamics Model}

Suppose the dynamics of the system can be described by an ordinary differential equation (ODE) of the form: 
\begin{align}
    \dot{\textbf{x}}=f(\textbf{x},\textbf{u},\textbf{d})
    \label{eq:aij_ODE}
\end{align}
\noindent where $f\colon \mathbb{R}^n \rightarrow \mathbb{R}^n$ describes the dynamics of the system, $\textbf{x} \in \mathbb{R}^n$ is the state vector, $\textbf{u} \in \mathbb{R}^m$ is input to the system, and $\textbf{d} \in \mathbb{R}^n$ is a disturbance input. Assuming that $f$ is globally Lipschitz continuous, a solution to (\ref{eq:aij_ODE}) describing the evolution of the system with initial condition $\textbf{x}_0 \in \mathbb{R}^n$, initial input $\textbf{u}_0$, and disturbance $\textbf{d}_0$, is any differentiable function $\psi(t)$, where $\psi \colon \mathbb{R}^+ \rightarrow \mathbb{R}^n$, such that $\psi(0) = \textbf{x}_0,\textbf{u}_0,\textbf{d}_0$ and $\dot{\psi}(t) = f(\psi(t))$ \cite{Puri1994}. Under our Lipschitz assumptions, the solution to the above differential equation is unique.

Suppose now that we wish to consider a family of solutions for the dynamics of a particular system, in order to characterize the uncertainty in the underlying model. In this realm, we can formulate the dynamics as a \textit{differential inclusion}. A differential inclusion can be written as:
\begin{align}
    \dot{\textbf{x}} \in \textit{F}(\textbf{x})
    \label{eq:aij_differential_inclusion}
\end{align}
\noindent where \textit{F} is a set-valued map from $\mathbb{R}^n$ to $\mathbb{R}^n$. That is $\textit{F}(\textbf{x}) \subset \mathbb{R}^n$. Maintaining our Lipschitz assumptions, a solution to (\ref{eq:aij_differential_inclusion}), with initial condition $\textbf{x}_0 \in \mathbb{R}^n$, initial input $\textbf{u}_0$, and disturbance $\textbf{d}_0 \in \mathbb{R}^n$ is any differentiable function $\psi_1(t)$, where $\psi_1 \colon \mathbb{R}^+ \rightarrow \mathbb{R}^n$, such that $\psi_1(0) = \textbf{x}_0,\textbf{u}_0,\textbf{d}_0$ and $\dot{\psi}_1(t) \subset F(\psi(t))$ \cite{Puri1994}. Whereas in (\ref{eq:aij_ODE}), the solution to the ODE had a unique solution, in this realm a differential inclusion has a family of solutions.

Bearing the above in mind, many classes of uncertainty, such as environmental uncertainty and modeling discrepancies, can be modeled as differential inclusions \cite{Gonzalez2020}. In this realm, one way of describing the uncertainty with respect to an underlying model is allowing the state, input, and parameters to be described by sets. As an example, if we assume that we can obtain bounds on the set of disturbances $\mathnormal{D}$, that represent all the things that are not explicitly modeled by $f$, then the behavior of the system resulting from interactions with any admissible disturbance can be rigorously obtained via the solution of the differential inclusion with $\textbf{d} \in \mathnormal{D}$ \cite{Gonzalez2020}.

% \noindent where \textit{F} is a set-valued map which associates any point $\textbf{x} \in \mathbb{R}^n $, under disturbance $\textbf{d} \in \mathbb{R}^n$  with a set-valued map \textit{F}(\textbf{x}), given by:

% \begin{align}
%     \bigcup\limits_{d \in \mathnormal{D}}^{} \ f(\textbf{x},\textbf{u},\textbf{d}).
% \end{align}

% \noindent Thus, under this regime, a system that begins at some initial state $\textbf{x}_0$ produces compact bundles of trajectories that describe the set of all trajectories of the system under all admissible disturbances \cite{DangMaler1998}.

\subsection{Online Reachability Computation}
\label{sec:aij_reachability_regime}
Before outlining the reachability framework leveraged in our work, let us define two key terms.% \smallskip

\begin{definition}[REACHTIME]
The reachtime, $T_{reach}$, is the finite time horizon for computing the reachable set.
\end{definition}% \smallskip

\begin{definition}[RUNTIME]
 The runtime, $T_{runtime}$, is the duration of (wall) time given for constructing the reachable set.
\end{definition} % \smallskip

With that, deriving the set of reachable states for an underlying system works as follows. Assuming a dynamics model for the system, we utilize the mixed face-lifting algorithm proposed in \cite{DangMaler1998}, to compute the set of reachable states from the current time $t$ up until $(t+T_{reach})$. The mixed face-lifting approach utilized here is part of a class of methods that deals with \textit{flow-pipe construction} or \textit{reachtube computation} \cite{Johnson2016}. This is done using snapshots of the set of reachable states that are enumerated at successive points in time. To formalize this concept, we define the reachable set below.

\begin{definition}[REACHABLE SET]
 Given a system with state vector $\textbf{x} \in \mathbb{R}^n$, input vector $\textbf{u} \in \mathbb{R}^m$, disturbance vector $\textbf{d} \in \mathbb{R}^n$, and dynamics $\dot{\textbf{x}}=f(\textbf{x},\textbf{u}, \textbf{d})$, where the initial states $\textbf{x}_0 = \textbf{x}(0)$, disturbances $\textbf{d}_0 = \textbf{d}(0)$,  and inputs $\textbf{u}_0 = \textbf{u}(0)$ are bounded by sets, $\textbf{x}_0 \in \chi_0$, $\textbf{d}_0 \in \mathnormal{D}_0$, $\textbf{u}_0 \in \mathnormal{U}$. The reachable set of the system for a time interval $ t \in [0,T_{reach}]$ is:%
\begin{equation*}%
    \mathnormal{R}_{[0,T_{reach}]} = \big \{ \psi(\textbf{x}_0,\textbf{u}_0,\textbf{d}_0,t) \ \big| \ \textbf{x}_0 \in \chi_0, \ \textbf{u}_0 \in \mathnormal{U} \ \textbf{d}_0 \in \mathnormal{D}, \ t \in [0,T_{reach}] \big \},%
\end{equation*}%
\noindent where $\psi(\textbf{x}_0,\textbf{u}_0,\textbf{d}_0,t)$ is the solution of the ODE at time $t$ with initial state $\textbf{x}_0$ under control input $\textbf{u}_0$ and disturbance $\textbf{d}_0$.\footnote{Our assumption is that $f$ is globally Lipschitz continuous. This property guarantees the existence and uniqueness of a solution for every initial condition in $\chi_0$.}
\label{def:aij_reachabilty}
\end{definition}

 In practice, for systems with non-trivial continuous dynamics, obtaining the exact reachable set is often extremely difficult or undecidable. In fact, even for linear systems, obtaining the exact reachable set is only possible if the matrices that describe the differential equations possess a specific eigen-structure \cite{Asarin2007}. Such a structure is outlined in \cite{LAFFERRIERE}. Thus, for general nonlinear systems, deriving the reachable set involves obtaining a sound over-approximation of $\mathnormal{R}_{[0,T_{reach}]}$, such that the actual system behavior is contained within the over-approximation \cite{AlthoffCORA2015,Chen2012,dang2000}. There are a variety of set representations for accomplishing this task, however, the algorithm utilized in this work uses $n$-dimensional hyper-rectangles (``boxes'') to generate reachtubes \cite{Johnson2016}. Over long reachtimes, the over-approximation error resulting from the use of this representation can be problematic. However, for short reachtimes, it is ideal in terms of its simplicity and speed \cite{Bak2014}.

\begin{figure}[!htbp]%[tb]
  \centering
  \includegraphics[width=\linewidth]{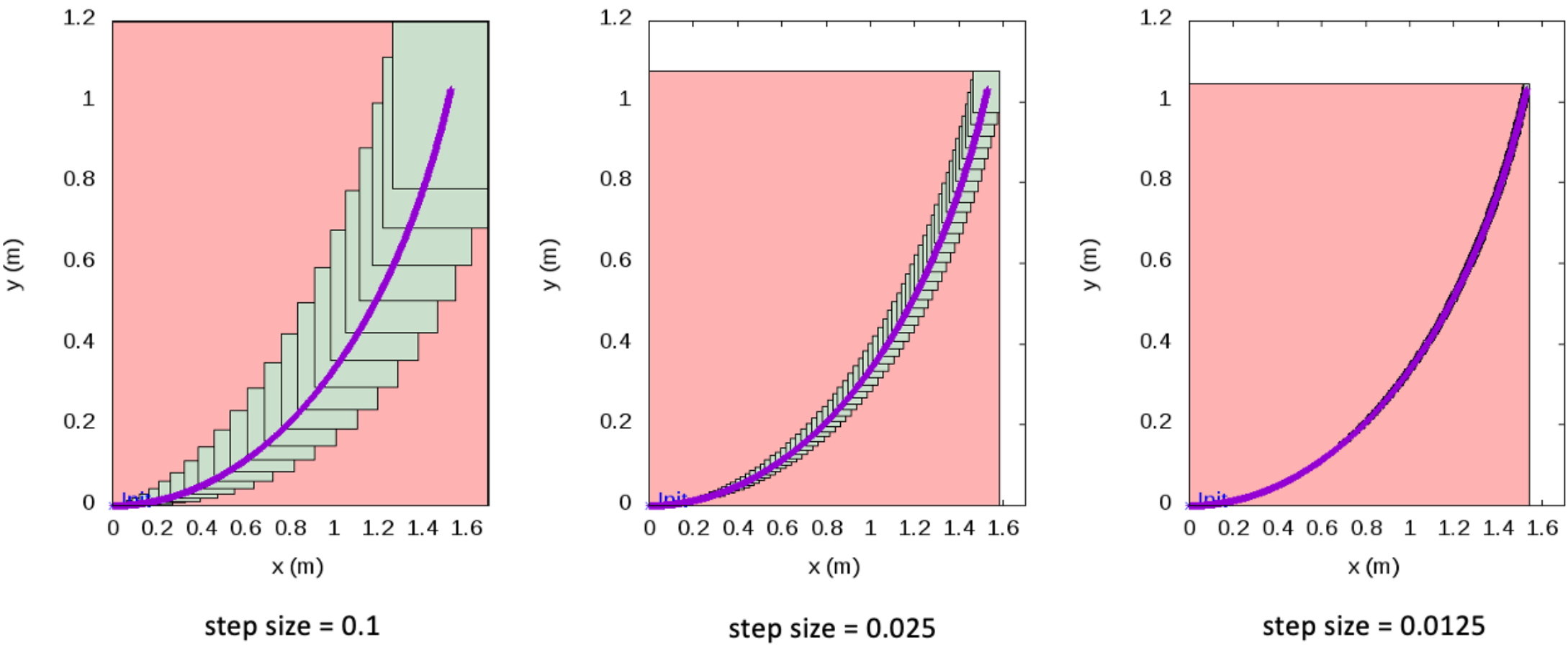}
  \caption{The real-time reachability algorithm always returns an over-approximation of the reachable set of states. The over-approximation error decreases with successive iterations, provided that there is enough runtime for re-computations. The above images demonstrate this aspect by simulating a left-hand turn control action for $T_{reach} = 2$ seconds. The green boxes represent the set of reachable states, the red rectangle represents the interval hull of the reachable states, and the purple points are points obtained from a simulation of the vehicle's dynamics.}
  \label{fig:aij_reach_refine}
\end{figure}%

The over-approximation error and the number of steps used in generating the reachable set can be controlled by a reachtime ($T_{reach}$) step size $h$. This parameter defines the level of discretization of the time interval $[0,T_{reach}]$ and can be used to tune the runtime of the reachability computation. Bak et al. leverage the step size to make the reachability algorithm amenable to the anytime computation model in the real-time scheduling literature \cite{Liu1991}. Thus, given a fixed runtime, $T_{runtime}$, we compute the reachable set $\mathnormal{R}_{[0,T_{reach}]}$, and if there is remaining runtime, we restart the reachability computation with a smaller step size. In both this work and \cite{Bak2014}, the step-size is halved in each successive iteration, leading to more accurate determinations of the reachable set. The relationship between the over-approximation error and the step size can be seen in \figref{fig:aij_reach_refine}. We refer readers to the following papers for an in depth treatment of these procedures \cite{dang2000,Bak2014,Johnson2016}.

\subsection{Safety Checking}
\label{sec:aij_safety_checking}

The computation of the reachable set allows us to reason whether the system under consideration will enter an unsafe situation in the future. Furthermore, by supposing a dynamics model for the dynamic obstacles within the environment, one can reason about potential future collisions. While the work described in \cite{Bak2014,dang2000,Johnson2016} made use of Lyapunov Stability theory in order to reason about the safety of systems, the following manuscript extended the approach to handle online collision avoidance queries. Thus, the safety checking in this work is formulated as checking whether an intersection between a set of unsafe states and $\mathnormal{R}_{[0,T_{reach}]}$ is empty. 

%Let $\Lambda$ represent the set of unsafe states.

We define the notion of safety considered in this work below.

\begin{definition}[SAFETY]
Let $\Lambda$ represent the set of unsafe states. A system is considered safe over the finite time horizon, $T_{reach}$, if  $\mathnormal{R}_{[0,T_{reach}]}  \cap \Lambda = \emptyset$.
\end{definition}%

The unsafe set, $\Lambda$, consists of all static obstacles within the environment, described by a bounding-box, the boundaries of the racetrack, characterized by a list of finely separated points, and the union of the reachable sets of the dynamic obstacles. These representations are then converted into their hyper-rectangle formulations that make up $\Lambda$.

\begin{definition}[UNSAFE SET]
Given a set of $N$ dynamic obstacles and a set $O$ of static obstacles within the environment, let $\mathnormal{R_i}_{[0,T_{reach}]}$ denote the reachable set of the $i$-\textit{th} dynamic obstacle. The set of unsafe states $\Lambda$ is:
\begin{equation*}
    \Lambda \ = \ O \ \cup \ \bigcup\limits_{i=1}^{N} \ \mathnormal{R_i}_{[0,T_{reach}]}.
\end{equation*}
\end{definition}

\figref{fig:aij_reachset} provides a visualization of the obstacles we considered in our simulation experiments, and \figref{fig:aij_lab_setup} displays one of the hardware experiments we conducted evaluating our approach. If there are no intersections between $\mathnormal{R}_{[0,T_{reach}]}$ and $\Lambda$, then we conclude that the system is safe. However, since our approach computes an over-approximation of $\mathnormal{R}_{[0,T_{reach}]}$, it may lead to conservative observations of unsafe behavior. This occurs when the error in the over-approximation of $\mathnormal{R}_{[0,T_{reach}]}$ results in intersections with the set of unsafe states, despite these intersections not occurring with the exact reachable set. By refining the reachable set in successive iterations, our regime seeks to mitigate the occurrence of falsely returning \emph{unsafe}.

There are two chief considerations in the anytime implementation of the safety checking procedure, which are (1) the overall soundness of our approach and (2) the real-time nature of our scheme. Satisfying both requirements constitutes the \emph{novel} extensions of the aforementioned algorithm. For the results of the verification to be sound, the safety checking process must be carried out in its entirety before a safety result is issued. At the same time, this requirement must be balanced alongside the real-time stipulation that tasks operate within pre-defined and deterministic time spans. Thus, our implementation ensures soundness properties while maintaining a low-likelihood of missing timing deadlines. 

Ideally, to ensure there were no missed deadlines, we would build our system in a \emph{Real-Time Operating System} (RTOS), which allows for the specification of task priorities, executing them within established time frames. However, our implementation does not make use of an RTOS and instead depends on native Linux and the Robot Operating System (ROS) to handle task management. To combat this shortcoming and reduce the number of missed deadlines, we estimate the time required to compute the next reachability loop. If our estimate exceeds the remaining allotted time, the process terminates. There is an inherent tradeoff between the conservativeness of our runtime estimates and the conservativeness of the resulting reachable set. In this work, we chose to maximize the number of iterations used in constructing the reachable set at the risk of occasionally missing deadlines. Our experiments demonstrate that we were successful in minimizing the number of missed deadlines during operation.

Let $k = T_{reach} / h$ denote the number of hyper-rectangles used in representing the reachable set.%$k$ is characterized by the following equation: $k = T_{reach} / h$.
\footnote{We begin the flow-pipe construction with an initial time step of $h = T_{reach}/10$.} Since each successive iteration decreases the step size by half, the number of hyper-rectangles that make up $\mathnormal{R}_{[0,T_{reach}]}$ doubles. Thus, the complexity of the safety checking process is $\mathcal{O}(2^k)$.\footnote{This analysis neglects consideration of the obstacles and the points used to represent the racetrack boundaries. Since these do not change between iterations, reasoning only about the hyper-rectangles is sufficient.} Therefore, we can estimate that a subsequent iteration of the algorithm will take twice as long as the current one and bloat this estimate to be conservative. 

A high-level overview of the reach-set construction and safety checking procedures defining our safety assurance framework is presented in Algorithm~\ref{alg:aij_algo_rtreach}. The $constructReachSet$ function is defined by Definition~\ref{def:aij_reachabilty}, and realized through hyper-rectangle-based mixed-face-lifting methods. Notably, we extend the original reach-set construction process outlined in \cite{Bak2014} to handle uncertain model parameters and set-based disturbances. The safety checking process is implemented as outlined in Section~\ref{sec:aij_safety_checking}. Computing the elapsed time, $computeElapsedTime()$, is done by leveraging functions of the underlying operating system. Finally, the function $estimateNextIterationRuntime(elapsedTime)$ is based on our requirement that the reach-set construction and safety checking process be carried out in their entirety, as outlined above.

\begin{algorithm}[htbp]%
\DontPrintSemicolon 
INITIALIZE{
\\
\textbf{Input:} $\chi_0,\mathnormal{U}, \mathnormal{D},T_{reach}, T_{runtime}$ \\
\textbf{Output:} $safe$ (boolean)
}

\vspace{2mm}

$elapsedTime = 0$\;
$T_{remaining} = T_{runtime}$\;
\While{$T_{remaining}>0$} {
    %\% Construct the Reachable Set\;
    $safe = $ true\;
    $\mathnormal{R}_{[0,T_{reach}]}= constructReachSet(\chi_0,\mathnormal{U}, \mathnormal{D},T_{reach},h)$\;
    %\% Check for intersections with unsafe set\;
    \If{$\mathnormal{R}_{[0,T_{reach}]} \cap \Lambda \neq \emptyset $}{
    $safe = $ false\;
    }
    $elapsedTime$ = $computeElapsedTime()$\;
    $nextIterationEstimate$ = $estimateNextIterationRuntime(elapsedTime)$\;
    $T_{remaining} = T_{runtime} -elapsedTime - nextIterationEstimate$\;
    $h = h /2$\;

}
\textbf{return:} $safe$
\caption{Safety Assurance Leveraging Real-Time Reachability}
\label{alg:aij_algo_rtreach}
\end{algorithm}%

\section{Experimental Overview}
\label{sec:aij_experimental_overview}

In this section, we detail the steps needed to implement our runtime assurance framework for the safety assurance of a 1/10 scale autonomous vehicle known as the F1/10. First, we construct a mathematical model of the F1/10 car's physical dynamics using system identification techniques. Next, we synthesize a series of controllers frequently used within autonomous racing whose control decisions will be analyzed at runtime. These controllers include a standard path tracking controller, a gap-following based collision avoidance controller, and a machine learning (ML controller) synthesized using imitation learning (IL). All the controllers use sensor information to determine the desired steering angle for the vehicle. At runtime, the mathematical model obtained through system identification is used within the reachability algorithm to reason about safety of the control actions selected by each of the controllers. Finally, the mathematical model of the F1/10 dynamics is augmented to include an interval based model of uncertainty and environmental disturbances. This allows us to reason about the effects that uncertainty imposes on the overall safety of the system. Finally, the simplex architecture provides the framework for ensuring safe operation of the F1/10 in the event of potential safety violations.

\subsection{The F1/10 Autonomous Platform}

\begin{figure}[htbp]%
  \centering
    % \hspace*{-8mm}  
    \includegraphics[width=0.5\linewidth]{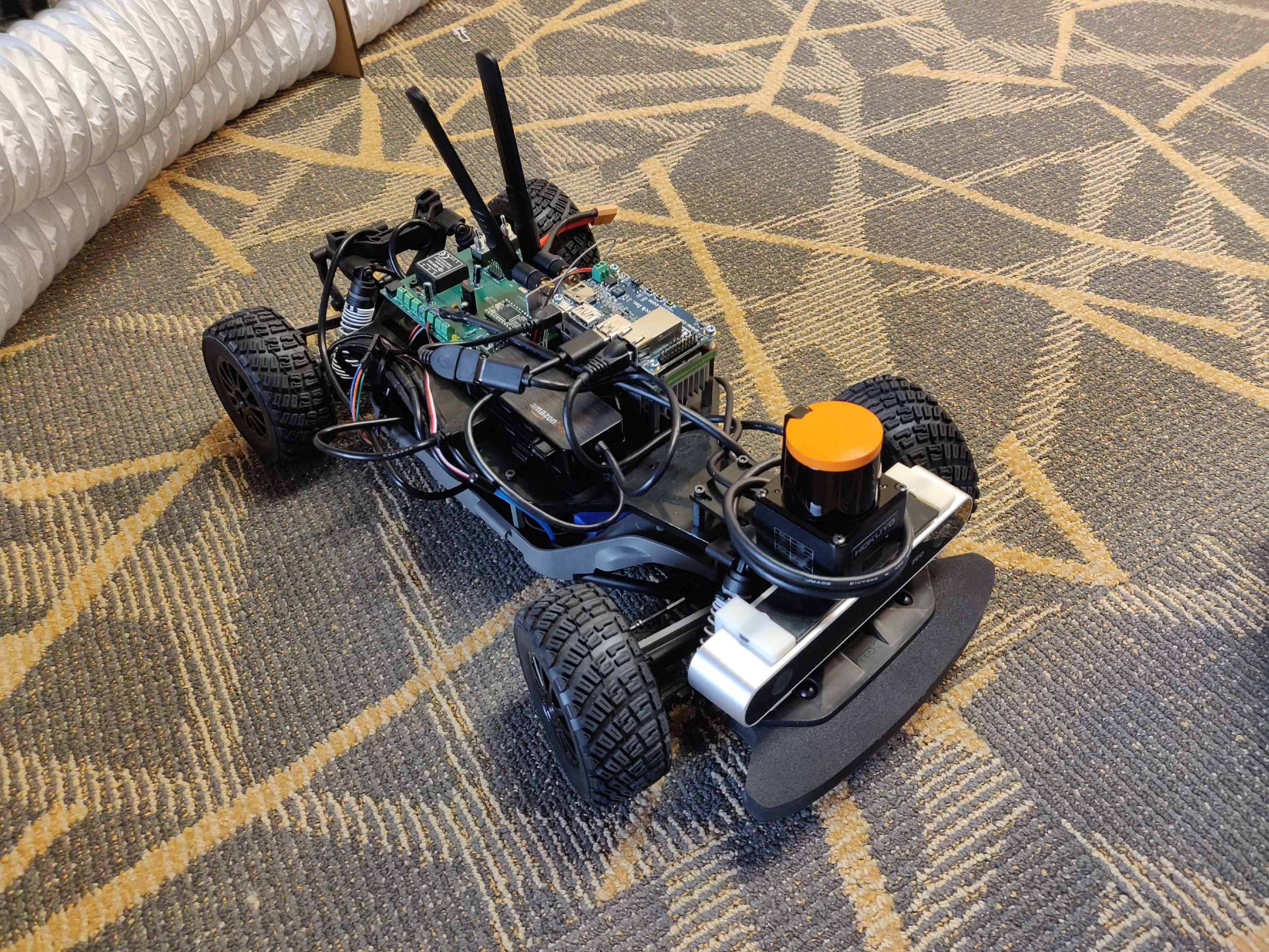}
  \caption{Visualization of our experimental F1/10 hardware platform. This platform is a one-tenth scale RC car that has been altered to entertain autonomous control inputs as well as support a sensor and compute architecture for autonomous decision-making. \cite{okelly2020}. }
  \label{fig:aij_f1tenth_hardware}
\end{figure}

The F1/10 platform of O'Kelly et al.~\cite{F1102019} was originally designed to emulate the hardware and software capabilities of full-scale autonomous vehicles. The platform is equipped with a standard suite of sensors such as stereo cameras, LiDAR (Light Detection And Ranging), and inertial measurement units (IMU). The platform uses an NVIDIA Jetson TX2 as its compute platform, and its software stack is built on the \emph{Robot Operating System} (ROS)\footnote{It is worth noting that ROS is not an operating system in the traditional sense but rather a meta-operating system that primarily provides the message passing interface for various components within robot software development \cite{Huang2014}.} \cite{ROS}. The result is a platform that allows researchers to conduct real-world experiments that investigate planning, networking, and intelligent control on a relatively low-cost, open-source test-bed \cite{F1102019}. A picture of the platform is shown in  \figref{fig:aij_f1tenth_hardware}. Additionally, to promote rapid prototyping and consider research questions around closing the simulation to reality gap\cite{Muratore2019}, Varundev Suresh et al. designed a Gazebo-based simulation environment \cite{Gazebo} that includes a realistic model of the F1/10 platform and its sensor stack \cite{varundev_ros_19}. We utilize this simulation environment for a number of experiments and training our controllers.

\subsection{System Identification and Model Validation}
\label{sec:aij_system_identification}
\begin{figure}[!htbp]%
  \centering
    % \hspace*{-8mm}  
    \includegraphics[width=0.9\linewidth]{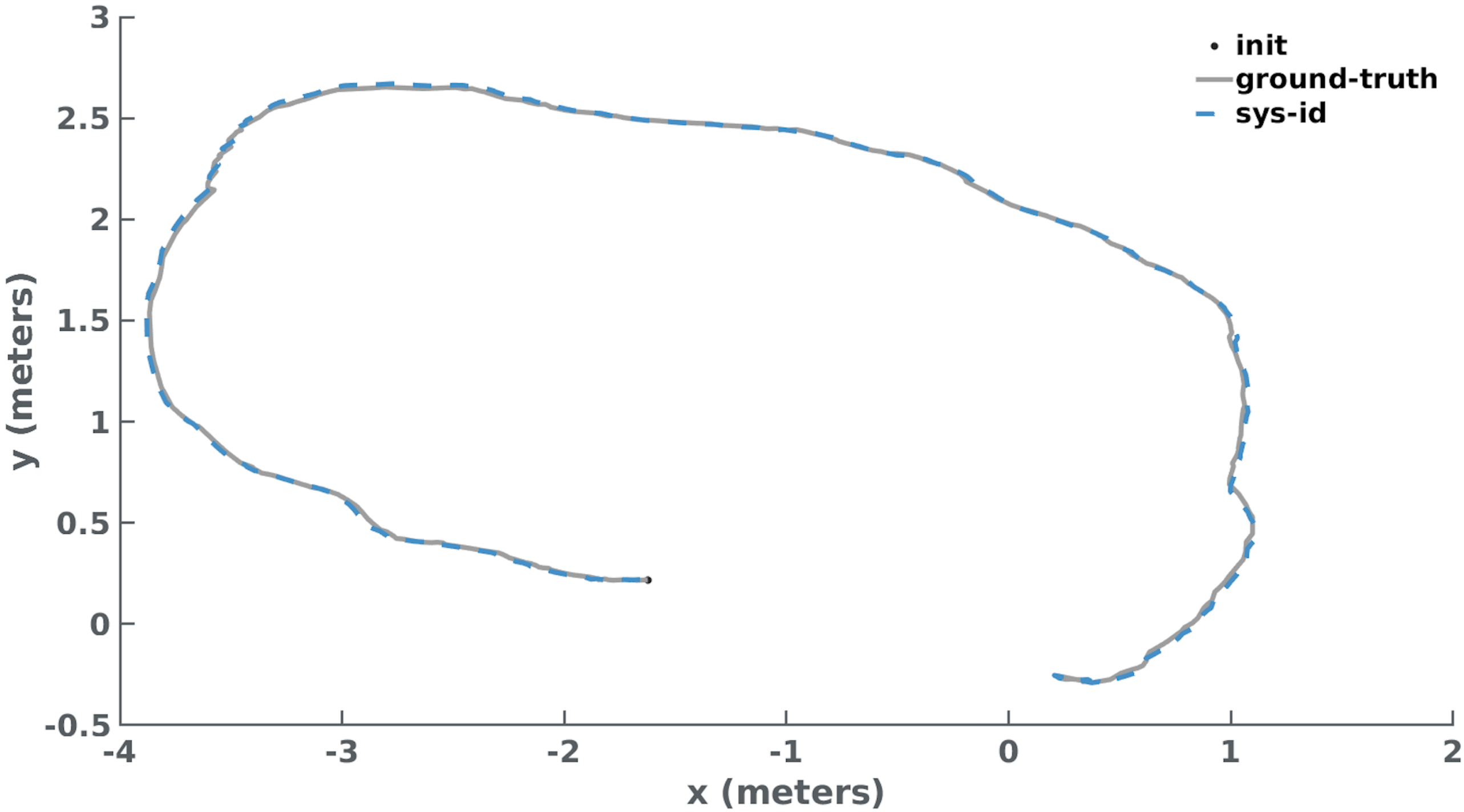}
  \caption{Vehicle Position (map frame). Illustrative example of an experiment used in validating the F1/10 Hardware Model. The model was validated using data collected from six experimental runs. \emph{Init} corresponds to the starting position of the vehicle in the considered experiment.}
  \label{fig:aij_validation}
\end{figure}%

The physical dynamics of the F1/10 vehicle can be modeled using a kinematic bicycle model \cite{Rajamani2012}, which is described by a set of four-dimensional nonlinear ordinary differential equations (ODEs). The kinematic bicycle model is characterized by relatively few parameters and tracks reasonably well at low speeds.\footnote{The kinematic bicycle model typically tracks well under $5 m/s$ \cite{ivanov2020case}} The model has four states: Euclidean positions $x$ and $y$, linear velocity $v$, and heading $\theta$. The dynamics are given by the following ODEs: 
\begin{equation}
    \begin{split}
        \Dot{x} & = v\cos(\theta +\beta)\\
    \Dot{y} & = v\sin(\theta + \beta)\\
    \Dot{v} & = -c_av +c_ac_m(u_v-c_h)\\
    \end{split}
    \quad
    \begin{split}
        \Dot{\theta} & = \frac{v\cos(\beta)}{l_f+l_r}\tan(\delta)\\
    \beta &= \tan^{-1}\Big(\frac{l_r\tan(\delta)}{l_f+l_r}\Big),
    \end{split}
    \label{eq:aij_kinematic_bicycle}
\end{equation}
\noindent where $v$ is the car's linear velocity, $\theta$ is the car's orientation, $\beta$ is the car's slip angle, $x$ and $y$ are the car's position, $u_v$ is the throttle input, $\delta$ is the steering input, $c_a$ is an acceleration constant, $c_m$ is a motor constant, $c_h$ is a hysteresis constant, and $l_f$ and $l_r$ are the distances from the car's center of mass to the front and rear respectively \cite{ivanov2020case}. For simplicity, since the slip angle is fairly small at low speeds, we assume that $\beta = 0$.

While the kinematic bicycle model is an effective model for describing vehicle dynamics, in order to describe the dynamics of the F1/10 precisely, it must be parametrized with respect to measured data obtained from experiments that characterize the behavior of the F1/10 system in various contexts \cite{SOHLBERG200811415}. The process of parametrizing an underlying theoretical model of a system with experimental data is often referred to as \textit{Grey-Box System Identification}, and in this work we utilized MATLAB's Grey-Box System Identification toolbox to obtain the acceleration constant, $c_a$, motor constant, $c_m$, and hysteresis constant, $c_h$, defining its dynamics. We obtained the following parameters for the simulation model of the F1/10: $c_a = 1.9569$, $c_m = 0.0342$, $c_h = -37.1967$, $l_f =0.225$, $l_r = 0.225$. The model was validated using six experimental campaigns %\footnote{The code used for system identification and model validation can be found at: \url{https://github.com/pmusau17/Platooning-F1Tenth/tree/master/src/race/sys_id}}
with an average \emph{Mean Squared Error} (MSE) of $0.003$. A sample experimental campaign is shown in \figref{fig:aij_validation}. For the hardware platform, we obtained the following parameters: $c_a = 2.9820$, $c_m = 0.0037$, $c_h = -222.1874$, $l_f =0.225$, $l_r = 0.225$, with a validation MSE of $6.75 \times 10^{-4}$.

The system identification results demonstrate that our model is reasonably accurate. However, as discussed in Section \ref{sec:aij_uncertainty_analysis}, developing an exact model is extremely difficult due to the presence of complex physical interactions that may be hard to describe precisely \cite{Akametalu2018}. Thus, to allow for the modeling of uncertainty, we can extend the dynamics presented in Equation~(\ref{eq:aij_kinematic_bicycle}) to allow each of the state variables and parameters that define the dynamics to be described by sets. Additionally, we allow for the modeling of bounded set based disturbances with respect to the velocity and orientation variables of our model in order to capture phenomena such as drag forces and friction that are not explicitly captured by the kinematic bicycle model. Therefore, the dynamics of our system become: 
\begin{equation}
    \begin{split}
        \Dot{\mathit{X}} & = \mathit{V}\cos(\Theta +\beta)\\
    \Dot{\mathit{Y}} & = \mathit{V}\sin(\Theta + \beta)\\
    \Dot{\mathit{V}} & = -C_a\mathit{V} +C_aC_m(u_v-C_h) + D_1\\
    \end{split}
    \quad
    \begin{split}
        \Dot{\Theta} & = \frac{V\cos(\beta)}{l_f+l_r}\tan(\delta) + D_2 \\
    \beta &= \tan^{-1}\Big(\frac{l_r\tan(\delta)}{l_f+l_r}\Big)
    \end{split}
    \label{eq:aij_Uncertainty}
\end{equation}
\noindent where the operations displayed above are set based operations, $\mathit{X}$,$\mathit{Y}$,$\mathit{V}$, and $\Theta$, are the equivalent set-based state variables, $D_1$ and $D_2$ denote the disturbance with respect to velocity and orientation, and $C_a, C_m$, and $C_h$ are sets that describe the uncertainty with respect to the acceleration constant, motor constant, and hysteresis constant defining our model.\footnote{As an example, one could use an interval to describe the uncertainty associated with one of the parameters defining the kinematic bicycle model. That is $C_a =  \{ c_a \in \mathbb{R} \colon \underline c_a \leq x \leq  \overline{c_a}  \}$, where $\underline c_a $ and $\overline{c_a}$ are the upper and lower bounds defining the acceleration constant $c_a$}

Accounting for increasing levels of uncertainty can lead to conservative behavior in the overall system that may potentially degrade system performance with respect to other objectives \cite{Akametalu2018}. One of the challenges that we investigate in this work is the tradeoff between the conservativeness of uncertainty estimates and the performance of our system. Specifically, what we measure is the percentage of actions issued by a controller that are labeled unsafe, as well as the growth in size of the set of reachable states of the system. The details are presented in Section~\ref{sec:aij_uncertainty_analysis}.

\subsection{Dynamic Obstacle Model}

Additionally, to reason about the safety of the system in the presence of dynamic obstacles, a model of their behavior is needed.  Specifically, we need to be able to compute how fast these agents are moving and the positions within the environment that they are likely to assume. This problem is frequently referred to as \textit{the obstacle tracking problem} within robotics and is a well studied and challenging topic within the autonomous vehicle, computer vision, and robotics literature \cite{YilmazObjectTracking}. In our experiments, we assume that the only dynamic obstacles present within the environment are the other vehicles participating in the race.

Typically, some assumptions are required to constrain the obstacle tracking problem to best suit the context of the application. As an example, if one wishes to model the behavior of an opponent vehicle within an autonomous racing scenario, then it is quite reasonable to model a dynamic agent using a kinematic bicycle model. However, the challenge in this context is predicting the behavior of the opponent vehicle, or in this case the set of control commands that an opponent agent is likely to use. Any predictions with respect to the behavior of the opponent will necessarily be characterized by a great deal of uncertainty. Therefore, one must carefully consider the fidelity of the models used to describe the behavior of dynamic agents. Models that are imprecise may lead to spurious declarations of unsafe behaviors, while models that are too rigid may be unable to capture the range of dynamic behaviors that agents can assume \cite{SeshiaTowards2016}.

In our framework, to avoid having to make distributional assumptions about the driving behavior of opponent vehicles, we assume that the obstacles can be described by a two-dimensional kinematic model and a corresponding bounding box. The assumptions made by this model are limited. Intuitively, our assumption is that dynamic agents will continue to follow their current trajectory over short time horizons. Thus, the equations describing the ODE are given as follows: 
\begin{align}%
\begin{split}
    \Dot{x} & = v_x,\\%
    \Dot{y} & = v_y,%
\end{split}
\label{eq:aij_opponent_dynamics}
\end{align}%
\noindent where $v_x$ and $v_y$ are the velocities in the $x$ and $y$ direction, respectively. Additionally, we make the assumption that we have access to the position and velocity of the other race participants. Similar to the kinematic bicycle model, the dynamics given by Equation~(\ref{eq:aij_opponent_dynamics}) can be formulated as a differential inclusion, in order to capture the uncertainty associated with measuring the position and velocity of dynamic obstacles using the vehicle's onboard sensors. 

\subsection{Controller Implementation}
\label{sec:aij_controllers}

One of the motivations in utilizing real-time reachability is that it abstracts away the need to analyze the underlying controller and instead focuses on the effects of control decisions on the system's future states. Thus, the nature of the underlying controller can vary quite significantly. This is particularly useful when the controller is a complex machine learning component such as a neural network that may be characterized by billions of parameters. To demonstrate the potential use cases for our methods as well as provide a broad picture of the application of our safety regime, we considered three different controllers within our experiments. These controllers include a standard path tracking controller, a gap-following based collision avoidance controller, and a machine learning controller synthesized via imitation learning (IL).

In this section, we provide a high-level introduction to the construction of the controllers used within our experiments.

% \todo{Rewrite this section: In this section we describe three controllers that are widely used in the F1/10 races and are considered in the case studies that follow.} 

\subsubsection{Pure Pursuit Controller}

The first controller considered within our experiments makes use of the Pure Pursuit algorithm. The Pure Pursuit algorithm is a widely used path-tracking algorithm that was originally designed to calculate the arc needed to get a robot back onto a path \cite{Coulter-1992-13338}. It has shown great success being used in numerous contexts, and in this work we utilize it to design a controller that allows the F1/10 vehicle to follow a pre-defined path along the center of the racetrack.

\subsubsection{Gap Following Controller}

Obstacle avoidance is an essential component of a successful autonomous racing strategy. Gap following approaches have shown great promise in dealing with dynamic and static obstacles. They are based on the construction of a gap array around the vehicle used for calculating the best heading angle needed to move the vehicle into the center of the maximum gap \cite{okelly2020}. In this work, we utilize a gap following controller called the ``disparity extender'' by Otterness et al. that won the F1/10 competition in April 2019 \cite{otterness_2019}. 

\subsubsection{Vision Based Imitation Learning} 

As modern data-driven and machine learning methods have become increasingly scalable and efficient, these methods have begun to be routinely used within autonomous applications. Particularly within perception tasks, machine learning models are frequently used to gain a semantic understanding of the objects within a vehicle's environment \cite{Devi2020ACS}. Unfortunately, these models are notoriously difficult to analyze \cite{Liu2019,xiang20118survey}.

One area of machine learning that has enjoyed wide success is Imitation Learning (IL). IL seeks to reproduce the behavior of a human or domain expert on a given task \cite{Hussein2017ImitationL}. These methods fall under the branch of \textit{Expert Systems} in AI, which has seen a surge in interest in recent years. The increased demand for these approaches is spurred on by two main motivations. (1) In many settings, the number of possible actions needed to execute a complex task is too large to cover using explicit programming. (2) Demonstrations show that having prior knowledge provided by an expert is more efficient than learning from scratch \cite{Hussein2017ImitationL}. While these approaches have demonstrated great efficacy in fixed contexts, there are concerns regarding their ability to generalize to novel contexts where the operating conditions are different from those seen during training, providing a need for effective runtime verification like the one explained in this work \cite{Hussein2017ImitationL}.

Since the seminal work of Krizhevsky et al. \cite{AlexNet2012} in the ImageNet Large Scale Recognition Challenge, \emph{Convolutional Neural Networks} (CNNs) have revolutionized the field of computer vision. Within the context of autonomous vehicles, CNNs have demonstrated efficacy for driving tasks such as lane following, path planning, and control, simultaneously, by computing steering commands directly from images \cite{bojarski2016end}.  In this work, we utilized the CNN architecture,  DAVE-2, initially proposed by Bojarski et al. to drive a 2016 Lincoln MKZ, to control the F1/10 model. The data we used to train DAVE-2 was collected from a set of simulation experiments where the sensor-action pairs were generated by a path tracking controller optimized to keep the F1/10 in the center of the track in the absence of obstacles. Such an environment is shown in \figref{fig:aij_reachset}.

Our main motivation in featuring a machine learning controller based on IL is to highlight the monitoring of a component that may be too complex to analyze. Beyond the seminal work of Bojarski et al. \cite{bojarski2016end}, machine learning components are not typically used for control tasks in autonomous systems beyond dealing with systems with unknown dynamics. However, there is a significant amount of promising research in this realm \cite{Tran2019}.

\subsection{ROS Simplex Architechture}
\label{section:aij_simplex}
\begin{figure}[!htbp]%
  \centering
  \includegraphics[width=\linewidth]{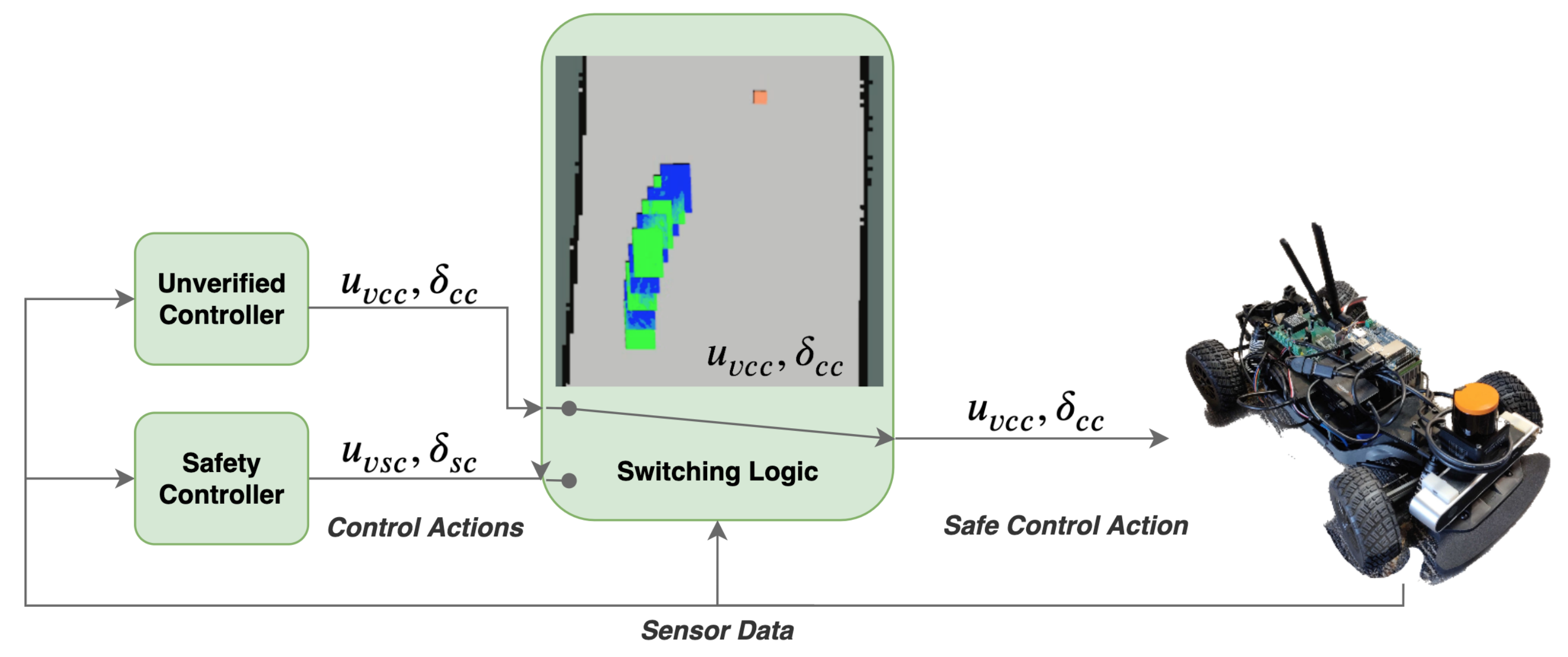}
  \caption{Overview of the simplex architecture deployed on the F1/10 system described in Section~\ref{section:aij_simplex}. The switching logic consists of monitoring the intersection between the reachable set of the F1/10 and the positions of static and dynamic obstacles within the environment. In the above figure, $u_{vcc}, \delta_{cc}$, corresponds to the control action issued by the complex, or high performance controller, while $u_{vsc}, \delta_{sc}$ corresponds to the control action issued by the safety controller. The reachability regime, uses $u_{vcc}, \delta_{cc}$ to determine the set of states that the vehicle will assume over $T_{reach}$. In the above figure, the alternating blue and green rectangles correspond to the intermediate reachable states defining the vehicle's trajectory. Since there are no intersections with obstacles in the environment, the control action issued by the complex controller can safely be used by the F1/10.}
  \label{fig:aij_simplex_arch}
\end{figure}

Our simplex architecture for the F1/10 is designed using ROS \cite{ROS}, and an overview of the design is shown in \figref{fig:aij_simplex_arch}. 

There are two considerations that play a major role in designing the simplex architecture: (1) the finite time horizon, $T_{reach}$, over which we are reasoning about safety and (2) the amount of time, $T_{runtime}$, allocated for the computation of the reachsets. In our experiments, we use $T_{reach} = 1.0s$ and $T_{runtime} = 25ms$, unless otherwise specified. These values were determined considering the empirical results of how long it took the F1/10 to come to a stop at speeds less than $1.5m/s$ and the control period, $20Hz$, which the reachability computation needs to finish within in order to not miss a deadline.\footnote{We limit velocities to $1.5 m/s$ because a lap on our physical track is approximately $13.08m$. Races held by the F1/10 community are around $30-50m$ per lap with larger distances between the track walls, allowing for much faster operating speeds. The rules are described in more detail here: \textbf{https://f1tenth.org/misc-docs/rules.pdf}}

Within this architecture, the primary sensors we rely on are a LiDAR and Stereo Labs' Zed Depth Camera. The messages from the LiDAR are published at $40Hz$, and the camera messages are published at $20Hz$. Additionally, we rely on odometry information, published at $40Hz$, to ascertain the state of the F1/10 vehicle. In our design, we decouple the control of the car's steering and throttle control. The controllers evaluated in our work are primarily concerned with the steering control, $\delta$, and the throttle control is designed to maintain a constant speed, $u_v$.  The primary reason we elected to use a constant speed in our experiments was to be able to evaluate the performance of each controller with respect to a single metric, thereby making comparisons across controllers easier.

In the traditional simplex architecture, both the decision module and the safety controller must be verified for the system to be verifiably safe \cite{Bak2014}. While this is straightforward for relatively simple controllers, it is significantly more challenging for many classes of controllers, especially when real-time execution is considered \cite{ivanov2020case}. 
However, the main focus of this work is evaluating the use of the reachability algorithm as a switching logic for the simplex architecture. Thus, we opted not to develop a ``formally verified'' safety controller. Instead, we selected a controller based on a gap-following algorithm optimized to avoid collisions with obstacles. A detailed description of the gap-following algorithm can be found in the following report \cite{otterness_2019}. It was primarily selected due to its robust collision avoidance ability and simplicity.

% \todo{How do I want to write this section? Primary focus here should be dynamic obstacle safety and uncertainty so I can do three controllers, pure pursuit (should have no obstacle avoidance capabilities, disparity extender, machine learning controller (end-to-end), I think this is good. Will show three different style controllers, but this disparity extender is the safety controller, so maybe two machine learning ones?}

\section{Experimental Evaluation}

Having described the details of our reachability algorithm, dynamics modeling, controller construction, notion of safety, and simplex architecture, we now present the experimental evaluation of our proposed approach. This section is concerned with four major experimental themes. Our first set of experiments presents a safety analysis of the controllers presented in Section~\ref{sec:aij_controllers} under a diverse set of experimental scenarios. The second study describes the implementation of the simplex architecture, described in Section~\ref{section:aij_simplex}, aimed at eliminating collisions and ensuring safety. Next, we present a runtime characterization of our reachability regime, evaluated using two separate platforms. Finally, we conclude this section with a presentation of a set of experiments analyzing the impact of various classes of uncertainty on the overall performance of our safety regime. 

The experiments presented here were conducted both in simulation on the physical F1/10 hardware platform. The aim of the simulation experiments was to promote reproducibility for those without hardware access and allow for consideration of a vast set of experiments. The hardware trials validate our claims that our safety regime admits minimal resource requirements.\footnote{The simulation artifacts can be found at:  \url{https://zenodo.org/record/6418817}.}

\subsection{Controller Safety Analysis}
\label{sec:aij_controller_safety_analysis}

The first set of experiments that we considered were concerned with how the controllers, discussed in Section~\ref{sec:aij_controllers}, performed in a variety of contexts. Specifically, what we were interested in was the portion of actions labeled as safe during a particular experiment. Consequently, to solely examine this metric, the following experiments did not make use of the simplex architecture that we described previously. To evaluate the performance of the various controllers, we considered a sizeable diversity of experiments with respect to the speed set-point %utilized by the controller tasked with commanding the F1/10,
$u_v$, the presence and configuration of obstacles, and the number of opponents present within the racetrack.\footnote{We limited the number of vehicles present within the racing environment to two or three vehicles, primarily because the simulator's performance decreases significantly with each additional car. The real-time factor for our Gazebo simulations, which is a parameter that communicates how fast time in the simulation environment is running relative to real-time, was on average about 0.61 on a desktop with 32 GB RAM and Intel Core i7-7700. Our future work aims to improve on this by utilizing a simulator with lower resource requirements.} The opposing vehicle's speed was set at $0.5m/s$ and utilized the disparity extender for navigation.\footnote{The reason we selected this speed was to guarantee that an overtaking action would be considered at least once during each experiment.} Our assumption is that the other vehicles in the environment are collision averse\footnote{This assumption however does not prohibit aggressive driving maneuvers}. Each experimental configuration was studied over five experiments consisting of a minute in length. A summary of these experiments is presented in Tables \ref{tab:aij_simulation_no_simplex}  and \ref{tab:aij_simulation_simplex}.

In general, the overall safety of each controller decreased as the number of static and dynamic obstacles present within the racing environment increased. The same effect was observed with an increase in the speed set-point utilized by each controller. Beyond these general trends, the overall performance of each controller varied significantly. For instance, since the pure pursuit controller was tasked with following a predefined path along the center line of the racing environment, it was in general the safest controller with respect to the number of actions labeled safe during an experiment. Provided that an opposing vehicle did not cross its path, or there were no obstacles located on the path that the pure pursuit controller was tasked with tracking, there was a low risk for collisions. In contrast, the pure pursuit controller performed poorly in experiments with obstacles, and it experienced the highest number of collisions among the controllers we studied. 

The vision based controller, which was trained to mimic the behavior of the pure pursuit controller, displayed similar results in the experiments that we considered. However, its performance was much less robust. In some scenarios, the vision based controller out-performed the pure pursuit controller. As an example, the vision based controller displayed an ability to deal with dynamic and static obstacles at low speeds. At the same time, its use also resulted in numerous collisions in scenarios without obstacles at speeds of greater than 0.5 m/s. Thus, the vision based controller was the least tolerant to speed changes in the experiments that we considered. While broad generalizations based on these results cannot be made, these results embolden our belief that monitoring machine learning components is of utmost importance.

Finally, the results of the experiments with the disparity extender were the most idiosyncratic. One of the most peculiar results of these experiments is that the use of the disparity extender displayed relatively low levels of safety despite being designed for collision avoidance. However, this observation can be explained by the greedy nature of its design, as described in \cite{otterness_2019}. Since the underlying goal of the disparity extender is to situate itself within the direction of the maximum gap, in practice, this design causes it to frequently steer the vehicle close to the racetrack boundaries as well as the other race participants. Despite the number of unsafe declarations issued by our reachability regime, the disparity extender was by far the most consistent controller across all the experiments that we conducted, and it displayed the greatest robustness to the presence of static and dynamic obstacles as well as higher speeds. Moreover, it experienced the lowest rate of collisions amongst the controllers that we evaluated. 

\begin{table}%[htbp!]%[]
    \renewcommand{\arraystretch}{1.3}
    \centering
    \caption{Controller Safety Analysis Without Use of the Simplex Architecture: Simulation Platform}
   
\resizebox{\linewidth}{!}{
    \begin{tabular}{ccccccc}
%\toprule
\multicolumn{1}{c}{} &
\multicolumn{1}{c}{} & 
\multicolumn{1}{c}{} & 

\multicolumn{2}{c}{\textbf{Obstacles}} & \multicolumn{2}{c}{\textbf{No Obstacles}} \\
\multicolumn{1}{c}{Controller}  & \multicolumn{1}{c}{Opponents} & \multicolumn{1}{c}{$u$ (m/s)} & \multicolumn{1}{c}{(\%) Safe Actions} & \multicolumn{1}{c}{Collision Frequency (\%)}  & \multicolumn{1}{c}{(\%) Safe Actions} & \multicolumn{1}{c}{Collision Frequency (\%)}  \\
\hline

Dispartiy Extender   &                   2 &      0.5 &          87.53 $\pm$   0.50 &                  0.0 &          87.64 $\pm$   0.88 &                  0.0 \\
                     &                     &      1.0 &          71.02 $\pm$   1.21 &                  0.0 &          69.21 $\pm$  13.32 &                 20.0 \\
                     &                     &      1.5 &          61.61 $\pm$  11.26 &                 60.0 &          71.71 $\pm$   1.27 &                 20.0 \\
                     &                   3 &      0.5 &          90.99 $\pm$   0.43 &                  0.0 &          92.17 $\pm$   0.62 &                  0.0 \\
                     &                     &      1.0 &          63.56 $\pm$   1.06 &                  0.0 &          65.84 $\pm$   1.58 &                  0.0 \\
                     &                     &      1.5 &          65.11 $\pm$   0.79 &                  0.0 &          66.89 $\pm$   1.16 &                 20.0 \\
Pure Pursuit         &                   2 &      0.5 &          96.56 $\pm$   0.29 &                100.0 &         100.00 $\pm$   0.00 &                  0.0 \\
                     &                     &      1.0 &          72.35 $\pm$   1.42 &                100.0 &          90.43 $\pm$  11.93 &                 20.0 \\
                     &                     &      1.5 &          82.86 $\pm$   4.77 &                 40.0 &          94.85 $\pm$   0.33 &                  0.0 \\
                     &                   3 &      0.5 &          96.79 $\pm$   0.16 &                100.0 &         100.00 $\pm$   0.00 &                  0.0 \\
                     &                     &      1.0 &           7.89 $\pm$   2.87 &                 80.0 &          12.14 $\pm$  12.19 &                100.0 \\
                     &                     &      1.5 &          33.86 $\pm$  33.06 &                100.0 &          73.27 $\pm$  34.94 &                 40.0 \\
Vision Based Network &                   2 &      0.5 &          95.36 $\pm$   2.83 &                 20.0 &          95.16 $\pm$   5.24 &                 20.0 \\
                     &                     &      1.0 &          75.14 $\pm$   9.97 &                100.0 &          16.85 $\pm$   6.09 &                100.0 \\
                     &                     &      1.5 &          62.47 $\pm$   3.49 &                100.0 &          24.53 $\pm$   2.26 &                100.0 \\
                     &                   3 &      0.5 &          97.49 $\pm$   0.56 &                  0.0 &          99.71 $\pm$   0.33 &                  0.0 \\
                     &                     &      1.0 &          37.13 $\pm$  25.66 &                 80.0 &          35.99 $\pm$   8.47 &                100.0 \\
                     &                     &      1.5 &          15.96 $\pm$   9.86 &                100.0 &          11.09 $\pm$   1.99 &                100.0 \\
\end{tabular}}
 \label{tab:aij_simulation_no_simplex}
\end{table}%

\begin{table}%[htbp!]%[]
    \renewcommand{\arraystretch}{1.3}
    \centering
    \caption{Controller Safety Analysis Using Simplex Architecture: Simulation Platform}
   
\resizebox{\linewidth}{!}{
    \begin{tabular}{ccccccc}
%\toprule
\multicolumn{1}{c}{} &
\multicolumn{1}{c}{} & 
\multicolumn{1}{c}{} & 

\multicolumn{2}{c}{\textbf{Obstacles}} & \multicolumn{2}{c}{\textbf{No Obstacles}} \\
\multicolumn{1}{c}{Controller}  & \multicolumn{1}{c}{Opponents} & \multicolumn{1}{c}{$u$ (m/s)} & \multicolumn{1}{c}{(\%) Safe Actions} & \multicolumn{1}{c}{Collision Frequency (\%)}  & \multicolumn{1}{c}{(\%) Safe Actions} & \multicolumn{1}{c}{Collision Frequency (\%)}  \\
\hline

Dispartiy Extender   &                   2 &      0.5 &          17.47 $\pm$   0.11 &                100.0 &          17.61 $\pm$   0.24 &                100.0 \\
                     &                     &      1.0 &          25.11 $\pm$   0.72 &                  0.0 &          24.73 $\pm$   0.97 &                  0.0 \\
                     &                     &      1.5 &          15.76 $\pm$   0.40 &                  0.0 &          15.87 $\pm$   0.56 &                  0.0 \\
                     &                   3 &      0.5 &          74.63 $\pm$   3.48 &                  0.0 &          71.51 $\pm$   5.38 &                  0.0 \\
                     &                     &      1.0 &          20.69 $\pm$   0.28 &                  0.0 &          21.12 $\pm$   0.75 &                  0.0 \\
                     &                     &      1.5 &          16.93 $\pm$   7.95 &                  0.0 &          13.39 $\pm$   0.27 &                  0.0 \\
Pure Pursuit         &                   2 &      0.5 &          93.95 $\pm$   0.11 &                  0.0 &         100.00 $\pm$   0.00 &                  0.0 \\
                     &                     &      1.0 &          74.37 $\pm$   1.13 &                  0.0 &          88.33 $\pm$   1.46 &                  0.0 \\
                     &                     &      1.5 &          62.52 $\pm$   1.95 &                  0.0 &          59.94 $\pm$   0.54 &                  0.0 \\
                     &                   3 &      0.5 &          94.17 $\pm$   0.11 &                  0.0 &         100.00 $\pm$   0.00 &                  0.0 \\
                     &                     &      1.0 &          60.50 $\pm$  17.63 &                  0.0 &          67.24 $\pm$  18.35 &                  0.0 \\
                     &                     &      1.5 &          21.04 $\pm$   0.34 &                  0.0 &          50.77 $\pm$   0.88 &                  0.0 \\
Vision Based Network &                   2 &      0.5 &          80.24 $\pm$   4.62 &                 20.0 &          91.71 $\pm$   4.68 &                  0.0 \\
                     &                     &      1.0 &          47.74 $\pm$   3.81 &                 40.0 &          43.36 $\pm$  13.78 &                 40.0 \\
                     &                     &      1.5 &          27.89 $\pm$  11.16 &                 80.0 &          29.88 $\pm$   6.18 &                 60.0 \\
                     &                   3 &      0.5 &          93.02 $\pm$   2.13 &                  0.0 &          97.78 $\pm$   1.24 &                  0.0 \\
                     &                     &      1.0 &          33.66 $\pm$   3.74 &                  0.0 &          40.69 $\pm$   0.26 &                  0.0 \\
                     &                     &      1.5 &          18.83 $\pm$   3.12 &                  0.0 &          38.25 $\pm$  12.81 &                  0.0 \\
\end{tabular}}
 \label{tab:aij_simulation_simplex}
\end{table}%

\begin{figure}[!htbp]
  \centering
    % \hspace*{-8mm}  
    \includegraphics[width=0.5\linewidth]{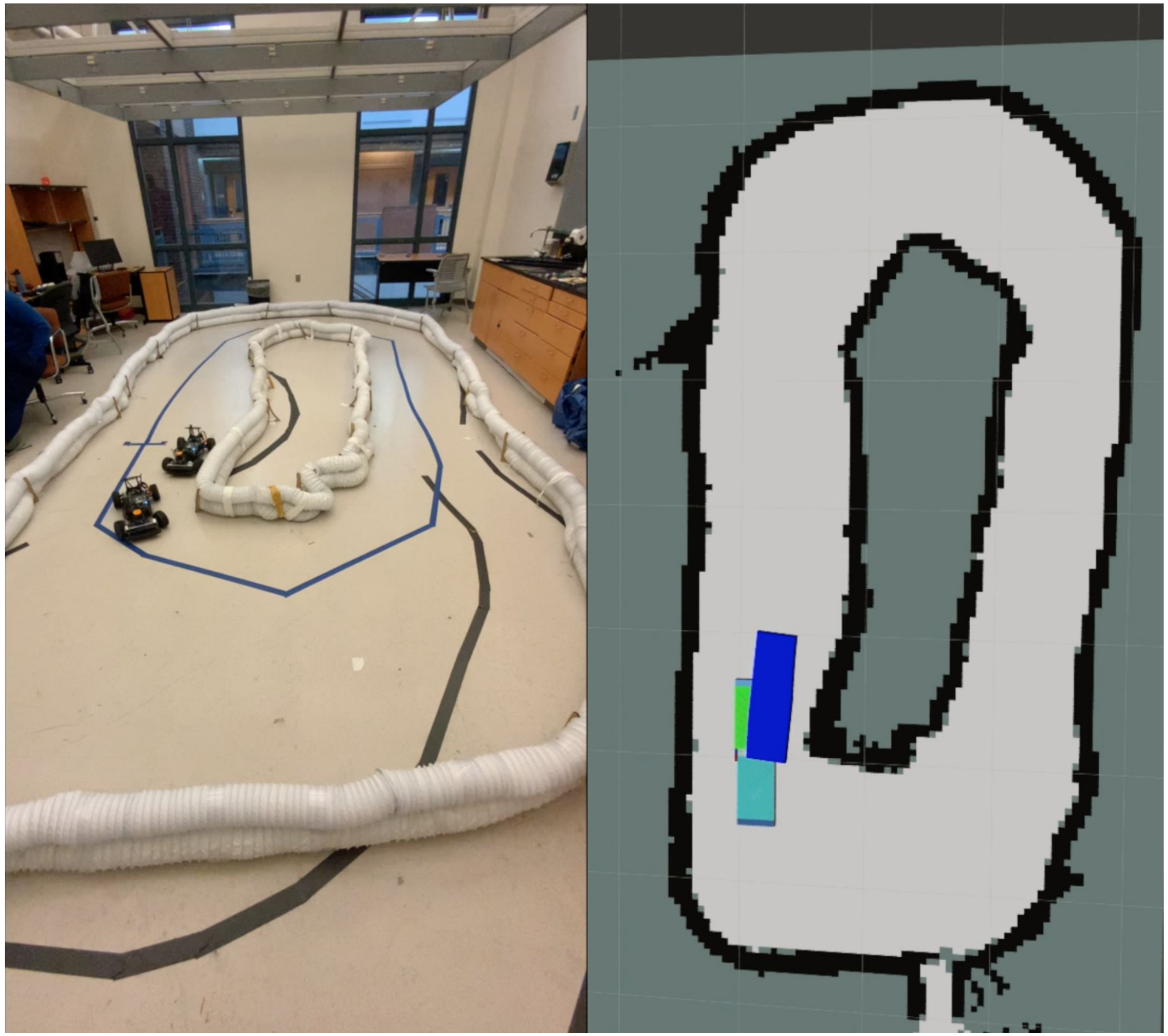}
    % \vspace{-1em}
   \caption{Example of one of the hardware experiments we conducted evaluating the efficacy of our safety regime in multi-agent racing settings. In the above image, the reachable set of the ego vehicle (dark blue) intersects with the reachable set of an opponent vehicle (light blue). This overlap corresponds to an action that would be labeled unsafe and a switch to the safety controller in our simplex architecture would occur.}

  \label{fig:aij_lab_setup}
\end{figure}

\subsection{Mitigating Collisions via Simplex}
\label{section:aij_experimental_simplex}

In the previous section, we were primarily concerned with the number of safe actions produced by a controller during an experiment, and due to the lack of the implementation of any mitigation strategies in the event of a potential safety violation, collisions occurred with varying levels of frequency depending on the nature of the controller. Preventing the occurrence of these collisions warrants the use of our simplex architecture, and here we present the results of utilizing our architecture under the same set of scenarios considered in Section~\ref{sec:aij_controller_safety_analysis}.  

Using our simplex architecture, we were able to completely eliminate collisions from occurring when the pure pursuit controller was used to control the F1/10 vehicle. Moreover, the number of collisions that occurred when the other two controllers were utilized also meaningfully decreased, as shown in Table~\ref{tab:aij_simulation_no_simplex}. However, in practice, to provably eliminate all collisions from occurring within our experiments, one must also verify the decision module and the safety controller utilized within the simplex regime. Within our work, the safety controller that we utilized was a conservative version of the disparity extender.\footnote{The underlying logic of the safety controller is the same as the disparity extender. However, we limit the maximum speed of this controller to be $0.3m/s$ . Additionally, if the distance between an obstacle and the F1/10 falls below $0.5m$, the vehicle stops completely.} Thus, while in practice it is quite effective in eliminating collisions, it is not perfect.

As stated earlier in this paper, developing a formally verified safety controller is quite challenging in practice. The challenge within this realm is developing controllers whose behavior is not exceedingly limited and whose analysis is feasible. However, the main focus of this work is evaluating the use of the reachability algorithm as a switching logic for the simplex architecture. Thus, rather than developing a provably safe simplex regime, our assumption is that our underlying safety controller is safe. We refer the interested reader to works such as \cite{ivanov2020case,Bak2014} for an in-depth discussion of developing provably safe controllers.

Finally, one of the greatest challenges in considering the question of safety within the context of dynamic obstacles is that collision avoidance strategies may not be enough to guarantee safety. For example, consider a scenario where a dynamic agent abruptly swerves into the path of the ego vehicle, thereby causing all the future trajectories of the ego vehicle to result in a collision. These scenarios are known as \textit{Inevitable Collision States} (ICS) \cite{Fraichard2007} and in recent years, there has been significant work towards avoiding ICS. Notably, one popular strategy of dealing with ICS is the use of reachability regimes which inherently satisfy two of the three conditions needed to avoid ICS. Specifically, reachability regimes inherently reason about the underlying dynamics of the ego vehicle, which is the first requirement of avoiding ICS. Secondly, they can be used to reason about the future state of the environment through the computation of the reachable sets of dynamic agents. This is the second condition for eliminating ICS. The final condition, requires reasoning about the previous two items over an infinite time horizon, which is infeasible in practice \cite{Fraichard2007}. However, one can meaningfully decrease the probability of a collision through the selection of a sufficiently long time horizon. Thus, the regime presented in this work can meaningfully be used to mitigate ICS.

Bearing the above in mind, we were able to eliminate all collisions in subsequent experiments by significantly increasing the reachtime horizon used in the reachability regime. However, doing so caused the overall use of the safety controller to increase significantly. While this may be acceptable in certain contexts, it is not always a desirable solution.

\subsection{Real-time Characterization of Reachability Regime}

One of the main benefits of the reachability regime presented in this work is that it admits minimal resource requirements and possesses real-time guarantees \cite{Bak2014}. In this section, we present, a runtime characterization of the real-time reachability regime outlined in Section~\ref{sec:aij_reachability_regime}. We ran our experiments on platforms running Linux (Ubuntu 16.04 LTS). The runtime analysis of the simulation experiments was conducted on a Dell XPS-15 (9570) with 32GB RAM, a six-core Intel Core i7-8750 $@ 4.1\textrm{GHz}$ processor, and an Nvidia GeForce GTX 1050Ti 4GB graphics card. The hardware experiments were evaluated on a Jetson TX2 with a Dual-core Nvidia Denver 64-bit CPU (ARM), a quad-core ARM A57 Complex, and an NVIDIA Pascal Architecture GPU with 256 CUDA cores. This latter configuration validates our claims that our safety architecture admits minimal resource requirements. 

For benchmarking purposes, we recorded the mean execution-times (Mean ET) of our real-time reachability algorithm, as well as the average number of iterations utilized in constructing the reachable set (Mean Iters). While a discussion of upper bounds on execution times typically involves a discussion of the \emph{Worst-Case Execution Time} (WCET), we instead report the \emph{Maximal Observed Execution Times} (MOET). In general, the WCET is unknown or difficult to derive without the use of static analysis proofs \cite{Reinhard2008}. Since our safety regime relies on ROS, which is highly dynamic and distributed, it is prohibitively difficult to perform an exhaustive exploration of the space of all execution times and thus derive the WCET.  However, we provide a rough proxy of the WCET by reporting the MOET \cite{Reinhard2008} in Tables~\ref{tab:aij_simulation_experiments} and \ref{tab:aij_hardware}. Additionally, we report the percentage of missed deadlines (PMD) that result from our soundness requirements; as we execute on a regular operating system and not an RTOS, this is possible, and performing the runtime measurements may result in variance due to changing load and scheduling. We demonstrate that this value is low across all experiments.

In the tables that follow, all summary statistics are reported alongside their corresponding standard deviations. Each configuration was evaluated using 30 experiments that were a minute in length.

\begin{table*}[!htpb]
\renewcommand{\arraystretch}{1.3}
\caption{Analysis of Wall-Time and Speed Variation Simulation Platform}
\centering
% \resizebox{0.8\linewidth}{!}{
\begin{tabular}{cccccc}% cccc}
    $u_v$ (m/s) & $T_{runtime}$ & MOET (ms) & Mean ET (ms) & Mean Iters &  PMD(\%) \\
    \hline
    
     0.5 & 10 &  10.71 $\pm$    0.97 &    7.00 $\pm$   0.13 &           3.95 $\pm$   0.03  & 0.13 $\pm$   0.15 \\
         & 25 &  28.49 $\pm$    5.23 &   15.42 $\pm$   0.88 &           5.07 $\pm$   0.12 & 2.62 $\pm$   2.71 \\
     1.0 & 10 &  11.10 $\pm$    1.25 &    6.85 $\pm$   0.16 &           4.08 $\pm$   0.23 & 0.27 $\pm$   0.24 \\
         & 25 &  28.36 $\pm$    0.77 &   14.86 $\pm$   0.14 &           5.06 $\pm$   0.10 & 2.83 $\pm$   1.64 \\
    %  1.5 & 10 &  11.36 $\pm$    0.60 &    6.83 $\pm$   0.16 &           4.04 $\pm$   0.14 & 0.77 $\pm$  1.13 \\
    %      & 25 &  28.19 $\pm$    0.53 &   15.00 $\pm$   0.25 &           5.06 $\pm$   0.03 & 3.45 $\pm$   1.59 \\
\end{tabular}
% }
\label{tab:aij_simulation_experiments}
\end{table*}%

\begin{table*}[!htbp]%
\renewcommand{\arraystretch}{1.3}
\caption{Analysis of Wall-Time and Speed Variation: Jetson TX2}
\label{tab:aij_hardware}
\centering
% \resizebox{0.8\linewidth}{!}{
\begin{tabular}{ccccccc}
 $u_v$ (m/s) & $T_{runtime}$ & MOET (ms) & Mean ET (ms) & Mean Iters &  PMD(\%) \\
\hline

     0.5 & 10 &  10.87 $\pm$   0.36 &    5.07 $\pm$  0.53 &           5.39 $\pm$  0.83 &  0.58 $\pm$  0.29 \\
         & 25 &  24.42 $\pm$   0.49 &   10.41 $\pm$  0.63 &           7.56 $\pm$  0.97 &  0.00 $\pm$  0.00 \\
     1.0 & 10 &  14.36 $\pm$   5.74 &    5.56 $\pm$  0.58 &           5.12 $\pm$  0.39 &  1.61 $\pm$  1.18 \\
         & 25 &  31.60 $\pm$  17.81 &    9.36 $\pm$  0.79 &           8.69 $\pm$  1.16 &  0.03 $\pm$  0.06 \\

\end{tabular}
% }%
\end{table*}%

The experiments demonstrated that Mean ET of our regime fell well within our desired $T_{runtime}$. Our estimates of future iterations of the reachability computations were quite conservative in nature. Though a few deviations were observed as displayed by the MOET, these did not significantly impact the performance of our approach. Rather, they are a result of our requirement that the safety checking process complete.

The runtime characterization of our approach done on the F1/10 hardware platform took on the same structure as the simulation experiments. Our experimental setup is shown in \figref{fig:aij_lab_setup}, and a video demonstration of the results is available online.\footnote{\url{https://youtu.be/3jPKucx4AF4}}. Similar to the simulation trials, the experiments demonstrated that the Mean ET of our regime fell well within our desired $T_{runtime}$. However, compared to the simulation experiments, our hardware experiments demonstrated a large decrease in the execution time of the approach. However, this result can be explained by the size of our racetrack and the frequency of unsafe action declarations that occurred during the experiments.\footnote{The racetrack considered in the simulation experiments is much larger than our real-world racetrack. In simulation, the narrowest point occurs when the walls have a separation of 2 meters. In contrast, our real-world track is slightly over 2 meters at its widest point. Thus, the frequency of unsafe actions declared by our approach in small spaces is significantly larger in our hardware experiments.}

Within the real-time reachability algorithm, the reachable-set computations terminate whenever an unsafe state is detected. The algorithm then restarts with half the step-size in order to refine the over-approximation error and determine if the ``unsafe'' declaration is spurious. This strategy continues until the step size falls below a pre-specified threshold specified to guarantee numerical stability. On the hardware platform, this threshold was met consistently, which demonstrates the frequency of unsafe declarations. This observation was explored more delicately in an earlier version of this work \cite{Musau2022}, and for brevity we refer readers to the aforementioned paper for a more in-depth discussion of this phenomenon. However, in general, experiments with higher levels of safety utilize fewer iterations in constructing the reachable set and generally have higher execution times. 
The numerical stability termination conditions are also discussed in more detail in~\cite{Johnson2016}.

\subsection{Uncertainty Analysis}

In Section~\ref{sec:aij_uncertainty_analysis}, we discussed how developing exact models of dynamical systems can be challenging due to the presence of complex physical interactions that may be difficult to explicitly model \cite{Akametalu2018}. While these interactions can frequently be lumped together as uncertainty, one of the challenges in system design is allowing for rigorous estimates of the bounds of uncertain parameters without prohibitively impacting the performance of the system \cite{Akametalu2018}. Intuitively, accounting for more uncertainty in the models of a system and its environment inherently leads to more conservative control strategies aimed at ensuring safety. In this section, we present an analysis of the effects of uncertainty on the conservativeness of the reachability computations that form the basis of our runtime assurance framework.  

Our experiments are primarily concerned with the effects of two classes of uncertainty: uncertainty with respect to our derived model of the system and uncertainty related to sensing, localization, and the environment. For context, this section summarizes over 6000, minute long experiments, that provide an enlightening analysis of the effects of uncertainty on our reachability regime.

\subsubsection{Model Uncertainty}

The difficulty of deriving models that are an accurate representation of the real world is the defining cause of uncertainty with respect to the structure and parameters of a system's models \cite{BERNARDO20162151}. In our experiments, the types of model uncertainty that we considered were set-based disturbances with respect to the velocity and orientation variables of our model, as well as uncertainty around the acceleration, motor constant, and hysteresis constants defining our model. Since the underlying reachability approach leverages hyper-rectangles for the reachable set computations, we can model phenomena such as drag forces and friction components that are not explicitly captured by the kinematic bicycle model using intervals. As an example, a friction parameter could be described by a real-valued interval:
\begin{align}
    \big[d_1\big] = \big[ \underline d_1 , \ \overline{d_1} \big] \ = \{ d_1 \in \mathbb{R} \colon \underline d_{1} \leq d_1 \leq  \overline{d_1}  \}
    \label{eq:aij_disturbance_interval}
\end{align}
\noindent which is a connected subset of $\mathbb{R}$,  and $\underline d_1$ and $\overline{d_1}$ are the lower and upper bounds of the additive and diminutive frictional components affecting the system's velocity such that, $\underline d_1 \leq \overline{d_1}$ \cite{ARCH16:Implementation_of_Interval_Arithmetic_CORA}. The same intuition applies to the disturbances applied to the steering of the F1/10 model. In this context, the equations of motion of the system are characterized by Equation~\ref{eq:aij_Uncertainty}.

Similarly, in contexts where the system dynamics are a combination of a partial theoretical model and parameters obtained from data, which corresponds to a standard grey-box system identification problem, there is typically some uncertainty associated with the parameters that characterize the underlying system \cite{LJUNG2004399}. The space of possible model structures as well as parameters that may characterize an underlying system may be extremely rich, and there is a large body of work aimed at characterizing this uncertainty \cite{LJUNG2004399,DotyUncertainty}. Thus, to explore this form of uncertainty in our work, we allowed the parameters defining the kinematic bicycle model to lie within an interval. As an example, one can imagine that the motor or acceleration constant defining the evolution of the velocity of our vehicle could vary depending on the environmental conditions that the system is tasked with operating within. 

\begin{table}[htbp]%[]
    \centering
    \caption{Uncertainty Analysis of Reachability Computations}
   
\resizebox{\linewidth}{!}{
    \begin{tabular}{ccccc}
%\toprule
\multicolumn{1}{c}{} 
& \multicolumn{2}{c}{Ground Truth} & \multicolumn{2}{c}{Particle Filter Localization} \\
\multicolumn{1}{c}{Parameter Uncertainty (\%)}   & \multicolumn{1}{c}{Controller Usage (\%)} & \multicolumn{1}{c}{Reachset Size (Median Area $A$)} & \multicolumn{1}{c}{Controller Usage (\%)} & \multicolumn{1}{c}{Reachset Size (Median Area $A$)}  \\
\hline
0.0  &  79.95 &   2.90 & 77.29 &  10.15 \\
5.0  &  78.27 &   4.88 & 72.92 &  12.43 \\
10.0 &  80.12 &   7.45 & 73.01 &  16.23 \\
15.0 &  76.26 &  10.56 & 65.16 &  20.30 \\
20.0 &  63.39 &  14.55 & 57.62 &  23.54 \\
25.0 &  56.42 &  19.02 & 52.24 &  28.28 \\
30.0 &  45.33 &  24.05 & 41.57 &  34.44 \\
35.0 &  38.53 &  30.82 & 32.54 &  40.47 \\
40.0 &  36.69 &  39.39 & 17.60 &  591.92 \\
45.0 &  37.69 &  47.71 &  1.61 &  2443.18 \\
\end{tabular}}
 \label{tab:aij_uncertainty_analysis}
\end{table}%

We evaluated the effects of uncertainty in our regime in two key ways. The first was an analysis of the growth of the size of the derived reachable set with respect to uncertainty. The second analysis was an investigation of the percentage of time in which the complex controller was utilized during an experimental run with increasing levels of uncertainty. To measure the growth in size of the set of reachable states for the F1/10 system with respect to increasing levels of uncertainty, we computed the median total area of the flow-pipe representing the set of trajectories that the vehicle could assume under the current control action. Since the safety checking task in our work was concerned with a two-dimensional intersection problem, one can compute the measure of the total area of a flow-pipe by computing the area of each intermediate hyper-rectangle describing the euclidean positions $x$ and $y$ of the vehicle over the finite-time horizon, $t \in [0,T_{reach}]$. In this way, one can investigate the relationship between uncertainty and the conservativeness of the reachability computations. 

We now formally define our notion of the total area of a flow-pipe representing the vehicle's trajectories. This notion can be extended to the concept of a volume in higher dimensions. Let $\mathnormal{R}_t \subset \mathnormal{R}_{[0,T_{reach}]}$ represent the reachable set of the system at time $t \in [0,T_{reach}]$. In our context, since we utilize hyper-rectangles as the set representation used to characterize flow-pipes, $\mathnormal{R}_t$ can be described by a Cartesian product of intervals $[x] \times [y]$. Here, $[x]$, and $[y]$ are intervals describing the euclidean position of the vehicle. Finally, let $w([x])$ represent the width of the interval $[x]$, where $w([x]) =  \overline{x} - \underline x$. Here $\underline x, \ \overline{x}$ are the left and right bounds defining $[x]$ such that $\underline x \leq \overline{x}$ \cite{Xiang2021TNLS}. Then the total area, $A$, of the flow-pipe, $\mathnormal{R}_{[0,T_{reach}]}$, can be defined as follows: 
\begin{equation}
    A = \sum_{t=0}^{T_{reach}} w([x]) w([y]).
    \label{eq:aij_area}
\end{equation}
\noindent There are three things to note in equation (\ref{eq:aij_area}). Firstly, the number of intermediate reachable sets $\mathnormal{R}_t$ is defined by the step size, $h$, used in the reachability computations, as described in Section~\ref{sec:aij_reachability_regime}. Secondly, since we had to bloat each intermediate hyper-rectangle representing the vehicle's trajectory to capture the true size of the F1/10 system, the area of each intermediate reachable set is necessarily non-zero. Finally, many of the intermediate reachsets overlap, which leads to an inflated estimate of the area that a flow-pipe assumes. However, our focus is on the relative growth of the flow-pipes under uncertainty. Thus, our over-approximation of the area in this context is sufficient.

\begin{figure}[htbp]%
  \centering
    % \hspace*{-8mm}  
    \includegraphics[width=\linewidth]{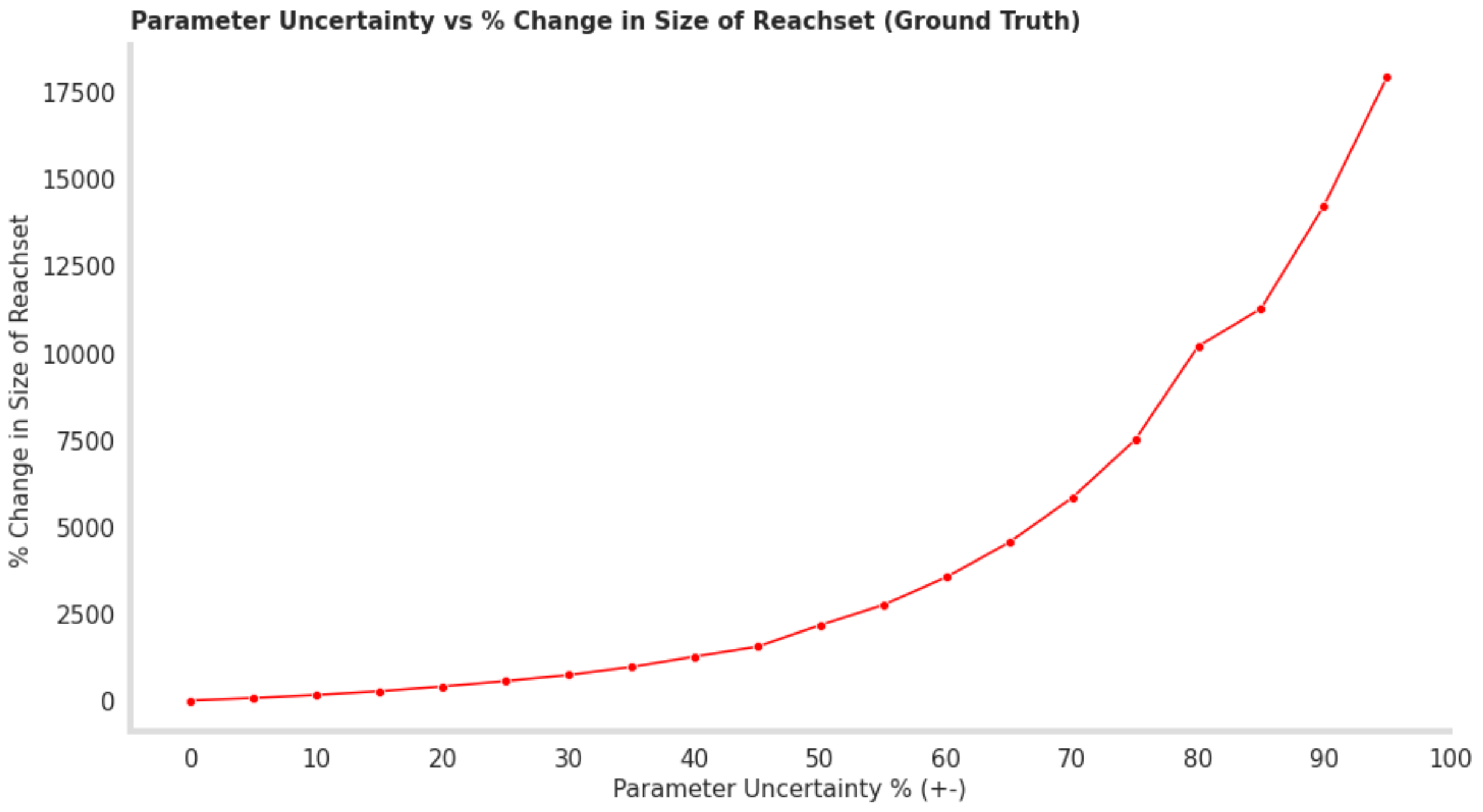}
  \caption{Relationship between the level of parameter uncertainty in the vehicle dynamics and the size of the reachable set describing the future behavior of the vehicle}
  \label{fig:aij_parameter_uncertainty}
\end{figure}%

\begin{figure}[htbp]%
  \centering
    % \hspace*{-8mm}  
    \includegraphics[width=\linewidth]{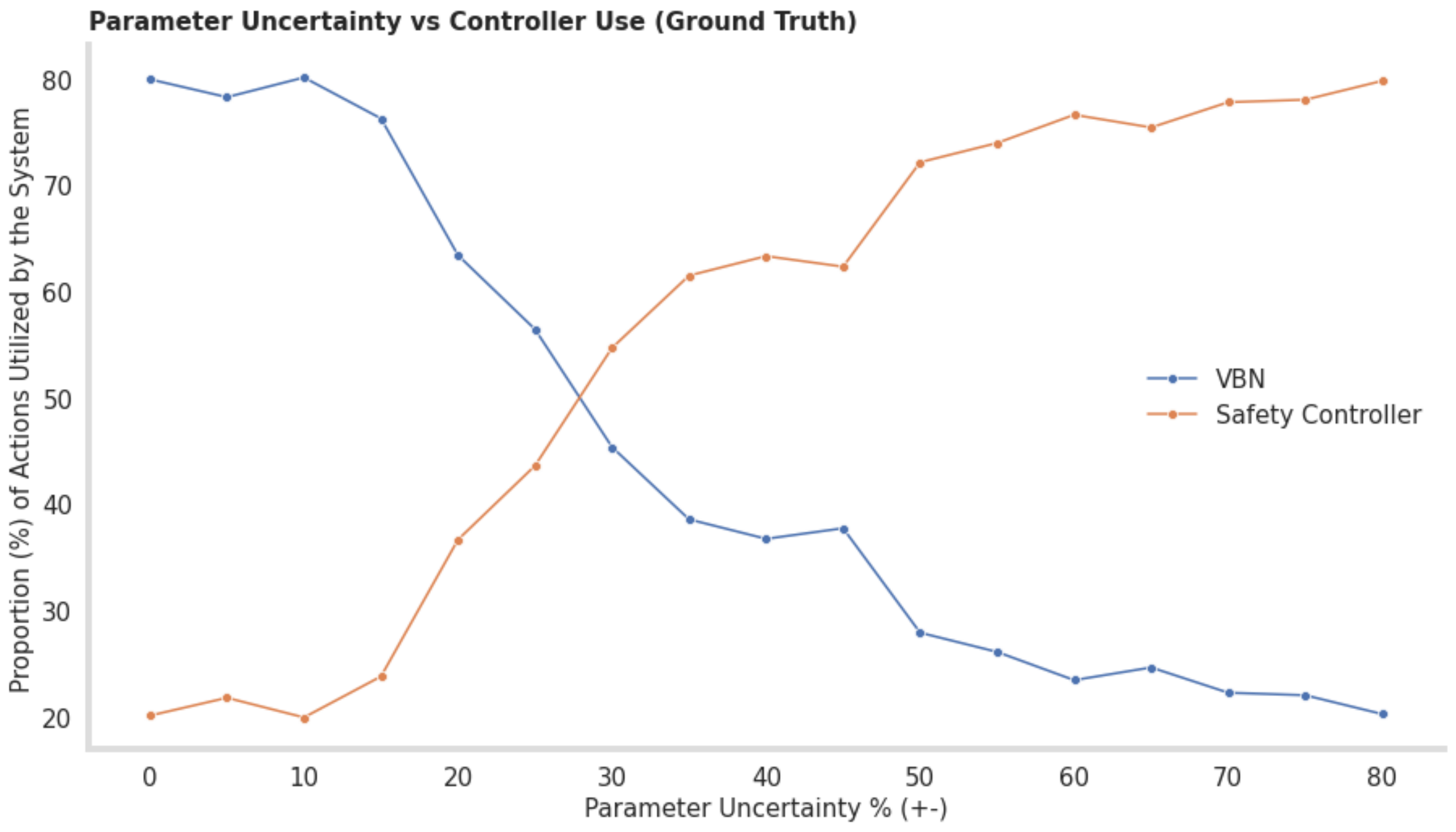}
  \caption{Relationship between the level of parameter uncertainty in the vehicle dynamics, and percentage of the time in which the vision based machine learning controller was utilized during an experimental run (Controller Usage).}
  \label{fig:aij_parameter_uncertainty_safety}
\end{figure}

\begin{figure}[!htbp]%
  \centering
    % \hspace*{-8mm}  
    \includegraphics[width=0.90
    \linewidth]{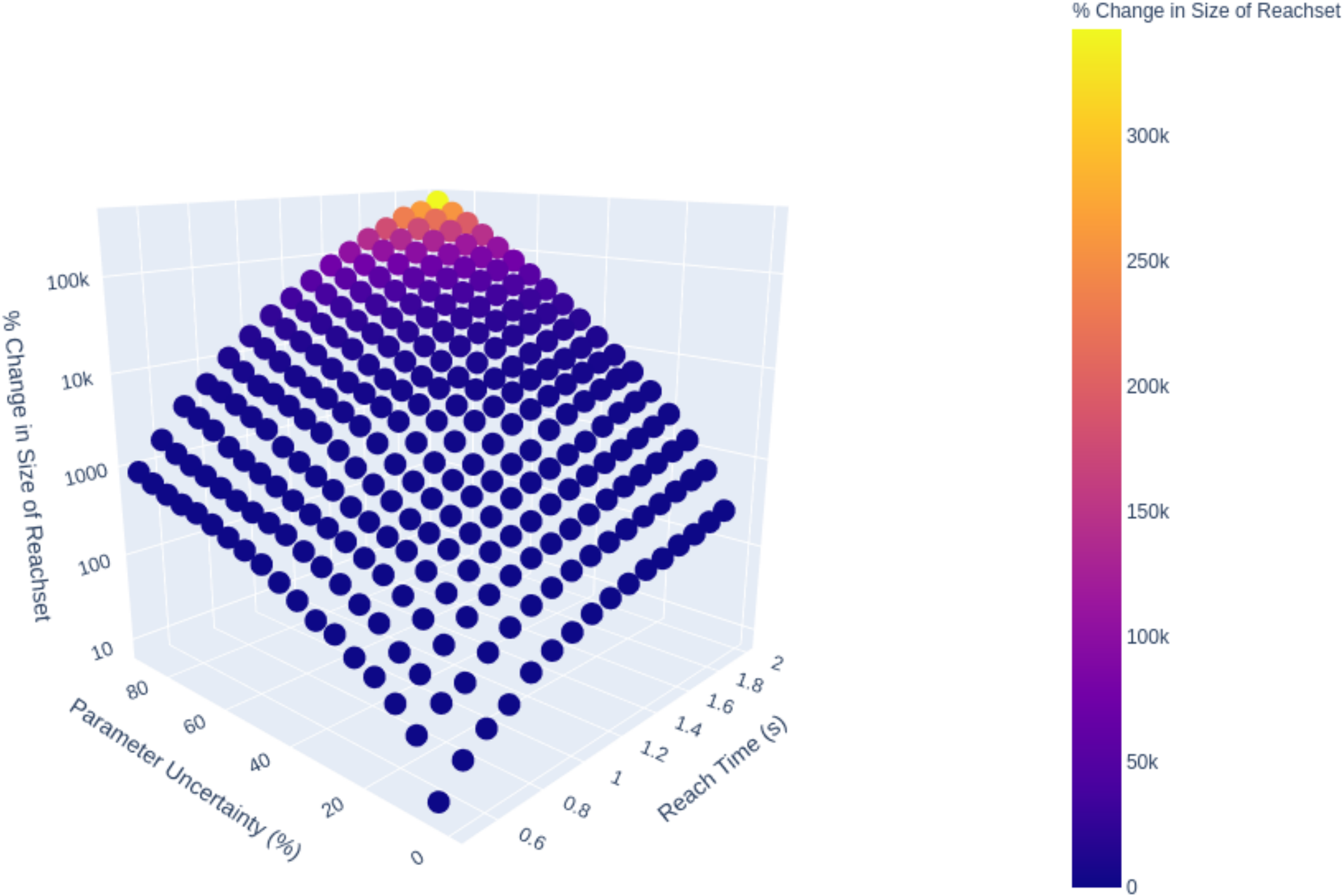}
  \caption{Relationship between the level of parameter uncertainty in the vehicle dynamics, and the size of the reachable set describing the future behavior of the vehicle. The interested reader can interact with the above figure using the following link: \url{
tinyurl.com/8wxx2xnm}.}
  \label{fig:aij_parameter_uncertainty3D}
\end{figure}

\figref{fig:aij_parameter_uncertainty} displays the growth in size of the reachable sets with respect to increasing levels of parameter uncertainty, while \figref{fig:aij_parameter_uncertainty_safety} displays the relationship between parameter uncertainty and the conservativeness of our safety regime. Table~\ref{tab:aij_uncertainty_analysis} presents the data shown in the aforementioned figures. From the figures, one can see that there is an exponential relationship between the growth in the size of the reachable set and increasing levels of uncertainty. The same is true of the relationship between uncertainty and the use of the safety controller. This relationship is further exacerbated if long time horizons are utilized in the underlying reachability approach, and \figref{fig:aij_parameter_uncertainty3D} presents a visualization of our investigation of how $T_{reach}$ impacts the conservativness of our reachset estimations. One  way of mitigating the growth of the size of the derived reachable set computations is by utilizing short time horizons. However, this may not always be feasible in all contexts.

\subsubsection{Modeling Sensor, Localization and Situational Uncertainty}

The data obtained from the sensors onboard autonomous vehicles possess inaccuracies that must be accounted for in the computations aimed at building a higher level understanding of the vehicle's surroundings \cite{Macfarlane2016}. One such example are inaccuracies or constraints related to the resolution of a particular sensor's measurements. Typically, significant testing allows for estimations of the variance of the measurements obtained from particular sensors in different contexts \cite{Macfarlane2016}. Moreover, these analyses often include descriptions of how various sensors perform in the context of varying weather conditions, temperatures, and other scenarios of interest \cite{Macfarlane2016}. 
These analyses will then inform how sensor observations are used within the control stack of the vehicle.

\begin{figure}[!htbp]%
  \centering
    % \hspace*{-8mm}  
  \includegraphics[width=0.85\linewidth]{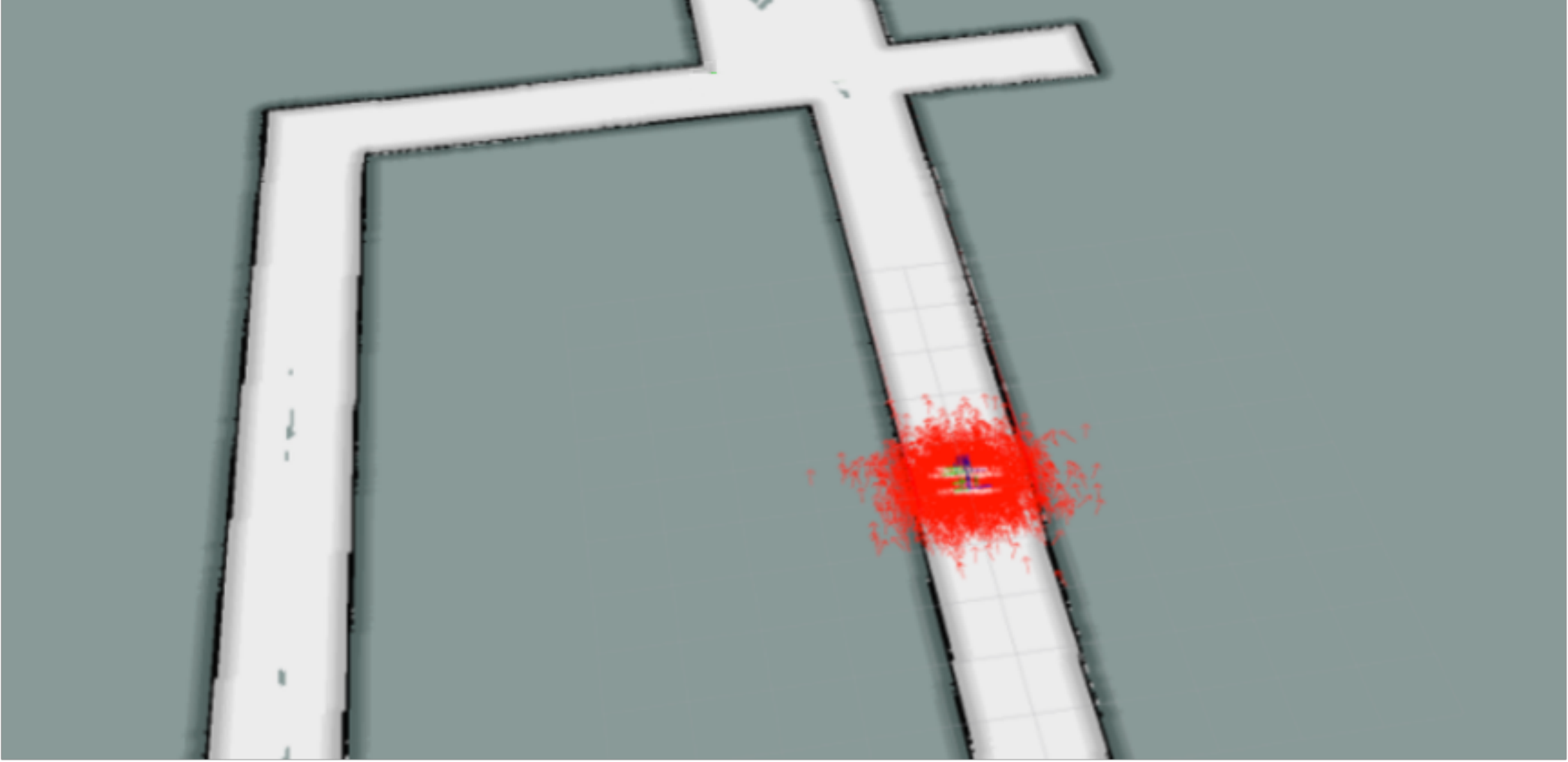}
  \caption{GPU-based particle filtering for position and orientation estimation, developed by Walsh et al. \cite{Walsh2017}. Each arrow represents a position and orientation estimate produced by the algorithm.}
  \label{fig:aij_sensor_uncertainty}
\end{figure}%

Bearing the above in mind, arguably the most salient problems within this context are characterizing the propagation of sensing errors, as they relate to the system's basic measure of its position in space and its environment \cite{Macfarlane2016}. Localization systems for autonomous vehicles are frequently based on measurements from a variety of sensors, and in state-of-the-art systems, estimations of the vehicle's position with respect to an underlying map must be accurate to within $10 cm$ or better \cite{Macfarlane2016}. This problem has received significant attention within the research literature, where localization subsystems often make use of algorithms such as particle filters that allow for estimations of the state of the system defined by probability densities that are a function of motion models and sensor information. Thus, they are quite adept at handling and estimating the uncertainty associated with the vehicle’s understanding of its environment \cite{Walsh2017}.

Finally, autonomous vehicles must be able to effectively handle interactions with other moving objects and vehicles within its environment \cite{Macfarlane2016}. These may include pedestrians, animals, and bicycles, whose behavior the vehicle may have limited knowledge about \cite{Macfarlane2016}. While effective methods for detecting, classifying, and tracking objects exist \cite{Girao2016}, many of these approaches make use of deep learning and probabilistic modeling in order to characterize the behavior of moving objects. Thus, there is an inherent uncertainty in the description of the vehicle's environment.

\begin{figure}[!htbp]%
  \centering
    % \hspace*{-8mm}  
    \includegraphics[width=0.79\linewidth]{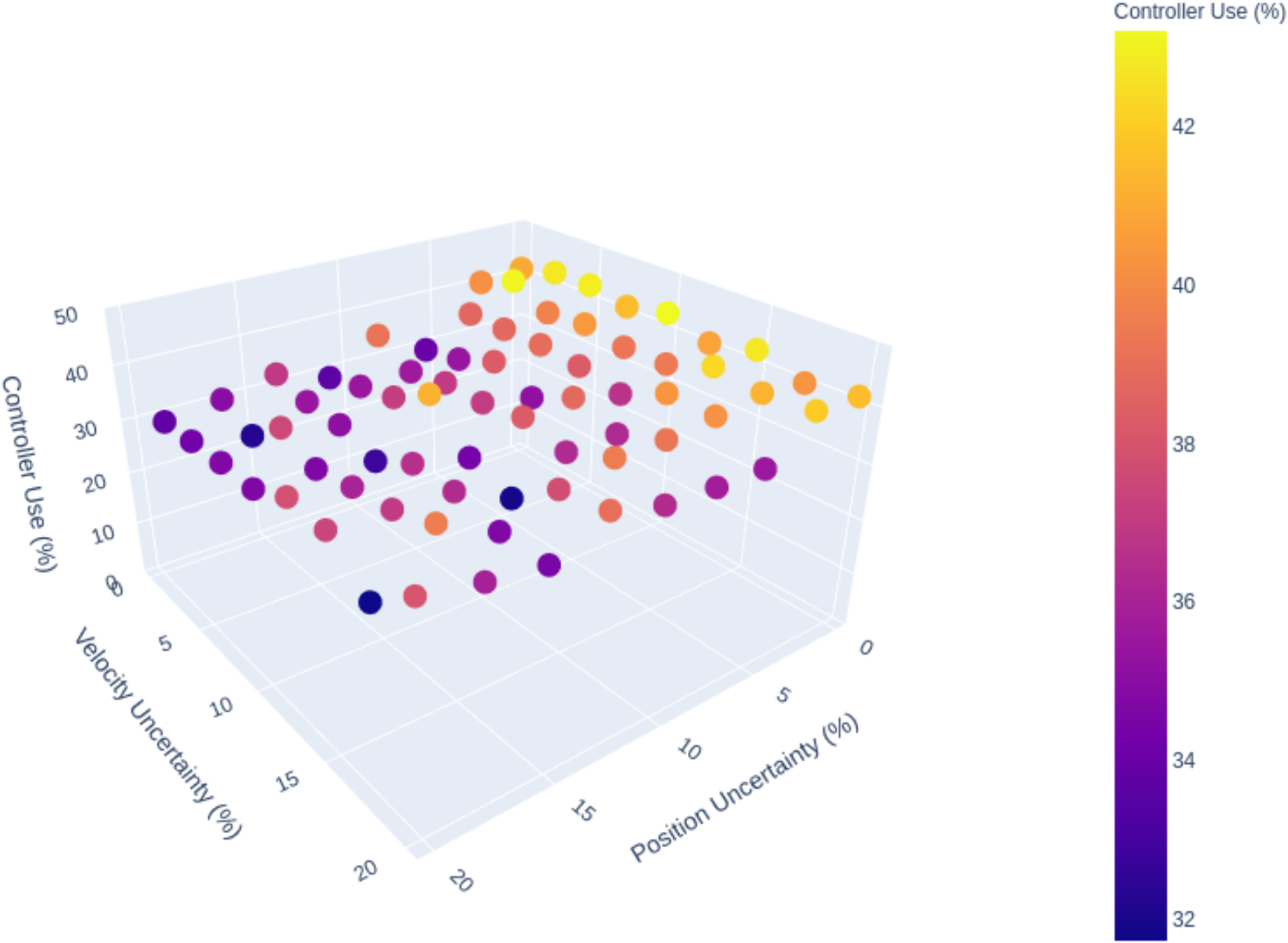}
  \caption{Visualization of the relationship between increasing levels of uncertainty with respect to estimations of the position and velocity of dynamic obstacles, and the use of the complex controller within our simplex regime. The interested reader can interact with the above figure using the following link: \url{https://tinyurl.com/3dcyab7n}.}
  \label{fig:aij_localization_uncertainty3D}
\end{figure}%

In our experiments, we primarily considered uncertainty as it related to the position and velocity of the dynamic obstacles within our environment, as well as uncertainty with respect to the vehicle's measurement of its own position and orientation within the environment. Specifically, we allowed our estimates of the position and velocity of opponent vehicles to lie within intervals. Intuitively, there is always some uncertainty as it relates to dynamic agents within an environment. Furthermore, our localization subsystem was based on a GPU-based particle filtering localization algorithm developed by Walsh et al. \cite{Walsh2017}. In this regime, the position of the vehicle can be defined by a set of position and orientation (pose) estimates, known as particles, that are refined using sensor measurements, a motion model, and odometry data. Rather than using an aggregate measure of these particles as our estimation of the vehicle's pose, we allowed our reachability computations to investigate the set of poses defined by each particle estimate. A visualization of the particles can be seen in \figref{fig:aij_sensor_uncertainty}.

Table~\ref{tab:aij_uncertainty_analysis} displays our analysis of parameter uncertainty in the physical dynamics of the F1/10 model when the particle filter was used for localization. It also includes an analysis of the effects of parameter uncertainty using ground truth data of the vehicle's position within the simulator. While the relationship between uncertainty and the conservativeness of our reachability regime were largely the same in this context, the exponential growth in the size of the reachable sets for the particle filter experiments was more drastic. In fact, when you compare the percent change in the growth of the reachable set, utilizing the zero parameter uncertainty scenario as a baseline, by the time an uncertainty level of 45\% is considered, the growth in the size of the reachsets for the particle filter localization approach are growing at a rate 15.5 times faster than the ground truth experiments. 

Finally, our experiments considering uncertainty with respect to the estimation of the velocity and position of dynamic obstacles within the vehicle's environment are shown in \figref{fig:aij_localization_uncertainty3D}. In these contexts, the impact of uncertainty was much less straightforward. While the size of the reachsets describing the position of other agents within the environment grew significantly, this effect had no material effect unless the ego vehicle was within close-proximity of a dynamic agent. Thus, while the length of time that the complex controller was used during an experiment decreased in general, it was not as significant of a drop as the other experiments. In general, to see a similar decrease in the use of the complex controller as the other experiments, one would need to evaluate the controller within contexts where the proximity between the ego vehicle and dynamic agents is small. However, in our work, due to our design choice of only switching back to the safe controller once a sufficient number of control actions have been determined safe, this is unlikely to occur.

In general, as shown in our experiments, one of the challenges with forward reachability schemes is that while they give strong notions of safety \cite{LeungReach2020}, over long time horizons, and significant uncertainty, this can lead to overly conservative behaviors, which may impede performance. An example of such a scenario are the localization uncertainty experiments. Beyond 40\% uncertainty in the model parameters, the safety controller was utilized 100\% of the time. While backward reachability approaches, such as Hamilton-Jacobi Reachability, are a possible alternative to these schemes, they typically incur a large computational cost. Additionally, we are not aware of any approaches that possess strong real-time guarantees. However, for low dimensional systems they are an attractive framework. We refer readers to the following paper \cite{Akametalu2018} for an in-depth discussion of these methods. 

\section{Discussion and Future Work}

\label{sec:aij_discussion_limitations}

Having evaluated the merits of our approach, both in simulation and on an embedded hardware platform, we now present some observations based on our results. In particular, we briefly focus on real-time considerations and the main limitations of our approach.

\subsection{Real-Time Evaluation and Missed Deadlines} 
The basic requirement for real-time systems is that tasks operate within pre-defined and deterministic time spans. Often, this is accomplished through the use of a real-time operating system (RTOS), which allows for the specification of task priorities to ensure they are executed within established time frames. Our implementation did not make use of an RTOS, thus task management was left to the native Linux implementation. While our experimental evaluation did demonstrate deviations from the specified wall-time, the mean percentage of missed deadlines on the Jetson TX2 was fewer than 2\% across all of our experiments.

\subsection{Limitations}

While the reachability algorithm presented in this work possesses provable guarantees, our architecture does not. Obtaining these guarantees requires developing a formally verified safety controller and switching logic, which was outside the scope of the work presented herein. Therefore, it is possible to enter a state in our framework in which all future trajectories will result in a collision. These states are known as \textit{inevitable collision states} and have been well-studied within the motion planning literature \cite{Fraichard}. In future work, we hope to address this limitation by leveraging approaches such as viability kernels and dynamic safety envelopes that allow for the synthesis of provable safe control regimes \cite{NILSSON}.

One of the challenges that emerged in our experiments was that uncertainty sharply increased the overall conservativeness of the reachable sets derived by our reachability regime. Thus, while the approach was quite successful in being used as part of a safety assurance architecture, sufficient care must be taken in order to minimize the quantity of spurious unsafe determinations that result from the over-approximation of reachable sets. While allocating more wall-time to the reachability regime is a possible solution, it is also worth considering other set representations while maintaining real-time guarantees. Moreover, one may also consider alternate reachability formulations, such as backward reachability regimes. However, these approaches come at the cost of significant computational overhead.

The second challenge is that the real-time reachability regime was designed to reason about relatively short time horizons, and there is an assumption that the control decision remains fixed throughout the reach-set construction. This assumption causes the verification results to be conservative in nature, and in future work we hope to expand this work to consider real-time closed-loop reachability analysis. The difficulty in performing closed-loop reach-set generation lies in developing accurate sensor models. As an example, for end-to-end control based on camera images, it is not clear how to generate camera images based on the state of the system to provide a meaningful and useful reachable set.

Lastly, in recent years, there has been a growth in approaches that perform online parameter estimation for dynamic obstacles within a robot's environment. In this work, we considered a simple two-dimensional kinematic model for the opponent vehicles within the racetrack environment. At low speeds, this model performs quite well; however, at higher speeds, these models would need to incorporate more sophisticated dynamics. In future work, we hope to evaluate online system identification within our framework.

\section{Conclusion}

In this manuscript, we presented a runtime verification framework leveraging real-time reachability and the simplex architecture for the safety assurance of a 1/10 scale autonomous vehicle called the F1/10 platform. The central idea behind our approach lies in computing the set of reachable states of the F1/10 system and ensuring that it never collides with both static and dynamic obstacles within its environment. Rather than analyzing the correctness of the controllers commanding the behavior of the F1/10, the reachability regime is leveraged to focus on the effects of a controller's decisions on the system's future states. In the event of a potential safety violation, a safety controller can be engaged in order to maintain safety. 

One of the key benefits of utilizing reachability regimes for the design of safety assurance frameworks is that they are quite adept in handling uncertainty, and in this work we presented a rigorous analysis of the effects of several classes of uncertainty in reasoning about the correctness of the system. Specifically, we allowed for the consideration of uncertainty with respect to the model of the underlying system, as well as uncertainty with respect to measuring the state of the vehicle's environment. Our experiments, conducted both in simulation and on an embedded hardware platform, validate the real-time aspects of our approach. Moreover, they demonstrate the efficacy of the simplex architecture in ensuring safety in different scenarios. 

Improving the over-conservativeness of the reachability framework, considering closed-loop reach-set generation, evaluating backward-reachability frameworks, making use of real-time operating systems, and incorporating dynamic obstacles into our regime are left for future work. Additionally, we wish to consider online learning applications, as our regime can be applied to such schemes with minimal modifications. Finally, our future work will consider the development of a verified safety controller and switching logic in order to maximize the benefits of the provable guarantees of our reachability framework.

% \begin{equation}
%     \begin{split}
%         \Dot{x} & = v\cos(\theta +\beta)\\
%     \Dot{y} & = v\sin(\theta + \beta)\\
%     \Dot{v} & = -c_av +c_ac_m(u_v-c_h) + d_1\\
%     \end{split}
%     \quad
%     \begin{split}
%         \Dot{\theta} & = \frac{v\cos(\beta)}{l_f+l_r}\tan(\delta) + d_2\\
%     \beta &= \tan^{-1}\Big(\frac{l_r\tan(\delta)}{l_f+l_r}\Big)
%     \end{split}
%     \label{eq:aij_kinematic_bicycle}
% \end{equation}

% %
% \begin{align}%
% \begin{split}
%     \Dot{x} & = v_x\\%
%     \Dot{y} & = v_y%
% \end{split}
% \label{eq:aij_opponent_dynamics}
% \end{align}%
% %

% A hyperplane $\mathcal{H}$ is a set which splits set $\mathbb{R}^{n}$ into two halfspaces and is formally defined as follows:
% \begin{align*}
% \begin{split}
%  \mathcal{H} = \{x \; | \;  a^{T}x = b\} \quad where \quad
%  a \in \mathbb{R}^{n}, b \in \mathbb{R}, a \not= 0 
% \end{split}
% \label{eq:hyperplanes}
% \end{align*} 

\section*{ACKNOWLEDGEMENT}

This material is based upon work supported by the Air Force Office of Scientific Research (AFOSR) under award number FA9550-22-1-0019, the National Science Foundation (NSF) under grant numbers 1918450, 1910017, and 2028001, the Department of Defense (DoD) through the National Defense Science \& Engineering Graduate (NDSEG) Fellowship Program, and the Defense Advanced Research Projects Agency (DARPA) Assured Autonomy program through contract number FA8750-18-C-0089.
Any opinions, finding, and conclusions or recommendations expressed in this material are those of the author(s) and do not necessarily reflect the views of the United States Air Force, DARPA, nor NSF.
\printcredits

\bibliographystyle{cas-model2-names}

\bibliography{cas-sc-template}

\begin{thebibliography}{105}
\expandafter\ifx\csname natexlab\endcsname\relax\def\natexlab#1{#1}\fi
\providecommand{\url}[1]{\texttt{#1}}
\providecommand{\href}[2]{#2}
\providecommand{\path}[1]{#1}
\providecommand{\DOIprefix}{doi:}
\providecommand{\ArXivprefix}{arXiv:}
\providecommand{\URLprefix}{URL: }
\providecommand{\Pubmedprefix}{pmid:}
\providecommand{\doi}[1]{\href{http://dx.doi.org/#1}{\path{#1}}}
\providecommand{\Pubmed}[1]{\href{pmid:#1}{\path{#1}}}
\providecommand{\bibinfo}[2]{#2}
\ifx\xfnm\relax \def\xfnm[#1]{\unskip,\space#1}\fi
%Type = Phdthesis
\bibitem[{Akametalu(2018)}]{Akametalu2018}
\bibinfo{author}{Akametalu, A.}, \bibinfo{year}{2018}.
\newblock \bibinfo{title}{A Learning-Based Approach to Safety for Uncertain
  Robotic Systems}.
\newblock Ph.D. thesis. EECS Department, University of California, Berkeley.
\newblock \URLprefix
  \url{http://www2.eecs.berkeley.edu/Pubs/TechRpts/2018/EECS-2018-41.html}.
%Type = Inproceedings
\bibitem[{Akametalu et~al.(2014)Akametalu, Kaynama, Fisac, Zeilinger, Gillula
  and Tomlin}]{Akametalu2014}
\bibinfo{author}{Akametalu, A.K.}, \bibinfo{author}{Kaynama, S.},
  \bibinfo{author}{Fisac, J.F.}, \bibinfo{author}{Zeilinger, M.N.},
  \bibinfo{author}{Gillula, J.H.}, \bibinfo{author}{Tomlin, C.J.},
  \bibinfo{year}{2014}.
\newblock \bibinfo{title}{Reachability-based safe learning with gaussian
  processes}, in: \bibinfo{booktitle}{Proceedings of the IEEE Conference on
  Decision and Control}, \bibinfo{publisher}{IEEE}, \bibinfo{address}{Los
  Angeles, CA}. pp. \bibinfo{pages}{1424 -- 1431}.
%Type = Inproceedings
\bibitem[{{Allen} et~al.(2014){Allen}, {Clark}, {Starek} and
  {Pavone}}]{Allen2014}
\bibinfo{author}{{Allen}, R.E.}, \bibinfo{author}{{Clark}, A.A.},
  \bibinfo{author}{{Starek}, J.A.}, \bibinfo{author}{{Pavone}, M.},
  \bibinfo{year}{2014}.
\newblock \bibinfo{title}{A machine learning approach for real-time
  reachability analysis}, in: \bibinfo{booktitle}{2014 IEEE/RSJ International
  Conference on Intelligent Robots and Systems}, pp.
  \bibinfo{pages}{2202--2208}.
%Type = Inproceedings
\bibitem[{Althoff(2015)}]{AlthoffCORA2015}
\bibinfo{author}{Althoff, M.}, \bibinfo{year}{2015}.
\newblock \bibinfo{title}{An introduction to cora 2015}, in:
  \bibinfo{editor}{Frehse, G.}, \bibinfo{editor}{Althoff, M.} (Eds.),
  \bibinfo{booktitle}{ARCH14-15. 1st and 2nd International Workshop on Applied
  veRification for Continuous and Hybrid Systems},
  \bibinfo{publisher}{EasyChair}, \bibinfo{address}{Berlin, Germany}. pp.
  \bibinfo{pages}{120--151}.
%Type = Article
\bibitem[{{Althoff} and {Dolan}(2014)}]{Althoff2014}
\bibinfo{author}{{Althoff}, M.}, \bibinfo{author}{{Dolan}, J.M.},
  \bibinfo{year}{2014}.
\newblock \bibinfo{title}{Online verification of automated road vehicles using
  reachability analysis}.
\newblock \bibinfo{journal}{IEEE Transactions on Robotics}
  \bibinfo{volume}{30}, \bibinfo{pages}{903--918}.
%Type = Inproceedings
\bibitem[{Althoff and
  Grebenyuk(2017)}]{ARCH16:Implementation_of_Interval_Arithmetic_CORA}
\bibinfo{author}{Althoff, M.}, \bibinfo{author}{Grebenyuk, D.},
  \bibinfo{year}{2017}.
\newblock \bibinfo{title}{Implementation of interval arithmetic in cora 2016},
  in: \bibinfo{editor}{Frehse, G.}, \bibinfo{editor}{Althoff, M.} (Eds.),
  \bibinfo{booktitle}{ARCH16. 3rd International Workshop on Applied
  Verification for Continuous and Hybrid Systems},
  \bibinfo{publisher}{EasyChair}. pp. \bibinfo{pages}{91--105}.
\newblock \URLprefix \url{https://easychair.org/publications/paper/ZRJ},
  \DOIprefix\doi{10.29007/w19b}.
%Type = Inproceedings
\bibitem[{{Amini} et~al.(2018){Amini}, {Schwarting}, {Rosman}, {Araki},
  {Karaman} and {Rus}}]{VariationalMIT2018}
\bibinfo{author}{{Amini}, A.}, \bibinfo{author}{{Schwarting}, W.},
  \bibinfo{author}{{Rosman}, G.}, \bibinfo{author}{{Araki}, B.},
  \bibinfo{author}{{Karaman}, S.}, \bibinfo{author}{{Rus}, D.},
  \bibinfo{year}{2018}.
\newblock \bibinfo{title}{Variational autoencoder for end-to-end control of
  autonomous driving with novelty detection and training de-biasing}, in:
  \bibinfo{booktitle}{2018 IEEE/RSJ International Conference on Intelligent
  Robots and Systems (IROS)}, pp. \bibinfo{pages}{568--575}.
%Type = Misc
\bibitem[{Anderson(2021)}]{driverlesschallenges}
\bibinfo{author}{Anderson, M.}, \bibinfo{year}{2021}.
\newblock \bibinfo{title}{Surprise! 2020 is not the year for self-driving
  cars}.
\newblock \URLprefix
  \url{https://spectrum.ieee.org/surprise-2020-is-not-the-year-for-selfdriving-cars}.
%Type = Article
\bibitem[{Asarin et~al.(2007)Asarin, Dang, Frehse, Girard, {Le Guernic} and
  Maler}]{Asarin2007}
\bibinfo{author}{Asarin, E.}, \bibinfo{author}{Dang, T.},
  \bibinfo{author}{Frehse, G.}, \bibinfo{author}{Girard, A.},
  \bibinfo{author}{{Le Guernic}, C.}, \bibinfo{author}{Maler, O.},
  \bibinfo{year}{2007}.
\newblock \bibinfo{title}{{Recent progress in continuous and hybrid
  reachability analysis}}.
\newblock \bibinfo{journal}{Proceedings of the 2006 IEEE Conference on Computer
  Aided Control Systems Design, CACSD} ,
  \bibinfo{pages}{1582--1587}\DOIprefix\doi{10.1109/CACSD.2006.285494}.
%Type = Inproceedings
\bibitem[{Asarin et~al.(2003)Asarin, Dang and Girard}]{Asarin2003}
\bibinfo{author}{Asarin, E.}, \bibinfo{author}{Dang, T.},
  \bibinfo{author}{Girard, A.}, \bibinfo{year}{2003}.
\newblock \bibinfo{title}{Reachability analysis of nonlinear systems using
  conservative approximation}, in: \bibinfo{editor}{Maler, O.},
  \bibinfo{editor}{Pnueli, A.} (Eds.), \bibinfo{booktitle}{Hybrid Systems:
  Computation and Control}, \bibinfo{publisher}{Springer Berlin Heidelberg},
  \bibinfo{address}{Berlin, Heidelberg}. pp. \bibinfo{pages}{20--35}.
%Type = Misc
\bibitem[{Babu and Behl(2019)}]{varundev_ros_19}
\bibinfo{author}{Babu, V.S.}, \bibinfo{author}{Behl, M.}, \bibinfo{year}{2019}.
\newblock \bibinfo{title}{Ros f1/10 autonomous racecar simulator}.
%Type = Article
\bibitem[{Bajcsy et~al.(2019)Bajcsy, Bansal, Bronstein, Tolani and
  Tomlin}]{Bajcsy2019Provably}
\bibinfo{author}{Bajcsy, A.}, \bibinfo{author}{Bansal, S.},
  \bibinfo{author}{Bronstein, E.}, \bibinfo{author}{Tolani, V.},
  \bibinfo{author}{Tomlin, C.J.}, \bibinfo{year}{2019}.
\newblock \bibinfo{title}{An efficient reachability-based framework for
  provably safe autonomous navigation in unknown environments}.
\newblock \bibinfo{journal}{Proceedings of the IEEE Conference on Decision and
  Control} \bibinfo{volume}{2019-December}, \bibinfo{pages}{1758 -- 1765}.
%Type = Inproceedings
\bibitem[{{Bak} et~al.(2009){Bak}, {Chivukula}, {Adekunle}, {Sun}, {Caccamo}
  and {Sha}}]{Bak2009Simplex}
\bibinfo{author}{{Bak}, S.}, \bibinfo{author}{{Chivukula}, D.K.},
  \bibinfo{author}{{Adekunle}, O.}, \bibinfo{author}{{Sun}, M.},
  \bibinfo{author}{{Caccamo}, M.}, \bibinfo{author}{{Sha}, L.},
  \bibinfo{year}{2009}.
\newblock \bibinfo{title}{The system-level simplex architecture for improved
  real-time embedded system safety}, in: \bibinfo{booktitle}{2009 15th IEEE
  Real-Time and Embedded Technology and Applications Symposium},
  \bibinfo{publisher}{IEEE}, \bibinfo{address}{San Francisco, CA}. pp.
  \bibinfo{pages}{99--107}.
%Type = Inproceedings
\bibitem[{{Bak} et~al.(2014){Bak}, {Johnson}, {Caccamo} and {Sha}}]{Bak2014}
\bibinfo{author}{{Bak}, S.}, \bibinfo{author}{{Johnson}, T.T.},
  \bibinfo{author}{{Caccamo}, M.}, \bibinfo{author}{{Sha}, L.},
  \bibinfo{year}{2014}.
\newblock \bibinfo{title}{Real-time reachability for verified simplex design},
  in: \bibinfo{booktitle}{2014 IEEE Real-Time Systems Symposium},
  \bibinfo{publisher}{IEEE}, \bibinfo{address}{Rome, Italy}. pp.
  \bibinfo{pages}{138--148}.
%Type = Inproceedings
\bibitem[{Ballester and Araujo(2016)}]{BallesterGoogLeNet}
\bibinfo{author}{Ballester, P.}, \bibinfo{author}{Araujo, R.M.},
  \bibinfo{year}{2016}.
\newblock \bibinfo{title}{On the performance of googlenet and alexnet applied
  to sketches}, in: \bibinfo{booktitle}{Proceedings of the Thirtieth AAAI
  Conference on Artificial Intelligence}, \bibinfo{publisher}{AAAI Press},
  \bibinfo{address}{Phoenix, Arizona}. pp. \bibinfo{pages}{1124--1128}.
%Type = Article
\bibitem[{Bansal et~al.(2019)Bansal, Bajcsy, Ratner, Dragan and
  Tomlin}]{bansal2020hamiltonjacobi}
\bibinfo{author}{Bansal, S.}, \bibinfo{author}{Bajcsy, A.},
  \bibinfo{author}{Ratner, E.}, \bibinfo{author}{Dragan, A.D.},
  \bibinfo{author}{Tomlin, C.J.}, \bibinfo{year}{2019}.
\newblock \bibinfo{title}{A hamilton-jacobi reachability-based framework for
  predicting and analyzing human motion for safe planning}.
\newblock \bibinfo{journal}{CoRR} \bibinfo{volume}{abs/1910.13369}.
\newblock \URLprefix \url{http://arxiv.org/abs/1910.13369},
  \href{http://arxiv.org/abs/1910.13369}{\tt arXiv:1910.13369}.
%Type = Article
\bibitem[{{Beg} et~al.(2017){Beg}, {Abbas}, {Johnson} and {Davoudi}}]{Beg2017}
\bibinfo{author}{{Beg}, O.A.}, \bibinfo{author}{{Abbas}, H.},
  \bibinfo{author}{{Johnson}, T.T.}, \bibinfo{author}{{Davoudi}, A.},
  \bibinfo{year}{2017}.
\newblock \bibinfo{title}{Model validation of pwm dc–dc converters}.
\newblock \bibinfo{journal}{IEEE Transactions on Industrial Electronics}
  \bibinfo{volume}{64}, \bibinfo{pages}{7049--7059}.
\newblock \DOIprefix\doi{10.1109/TIE.2017.2688961}.
%Type = Incollection
\bibitem[{Bernardo(2016)}]{BERNARDO20162151}
\bibinfo{author}{Bernardo, F.P.}, \bibinfo{year}{2016}.
\newblock \bibinfo{title}{Model analysis and optimization under uncertainty
  using highly efficient integration techniques}, in:
  \bibinfo{editor}{Kravanja, Z.}, \bibinfo{editor}{Bogataj, M.} (Eds.),
  \bibinfo{booktitle}{26th European Symposium on Computer Aided Process
  Engineering}. \bibinfo{publisher}{Elsevier}. volume~\bibinfo{volume}{38} of
  \textit{\bibinfo{series}{Computer Aided Chemical Engineering}}, pp.
  \bibinfo{pages}{2151--2156}.
\newblock \URLprefix
  \url{https://www.sciencedirect.com/science/article/pii/B9780444634283503635},
  \DOIprefix\doi{https://doi.org/10.1016/B978-0-444-63428-3.50363-5}.
%Type = Inproceedings
\bibitem[{Bojarski et~al.(2018)Bojarski, Choromanska, Choromanski, Firner,
  Ackel, Muller, Yeres and Zieba}]{bojarski2016end}
\bibinfo{author}{Bojarski, M.}, \bibinfo{author}{Choromanska, A.},
  \bibinfo{author}{Choromanski, K.}, \bibinfo{author}{Firner, B.},
  \bibinfo{author}{Ackel, L.J.}, \bibinfo{author}{Muller, U.},
  \bibinfo{author}{Yeres, P.}, \bibinfo{author}{Zieba, K.},
  \bibinfo{year}{2018}.
\newblock \bibinfo{title}{Visualbackprop: Efficient visualization of cnns for
  autonomous driving}, in: \bibinfo{booktitle}{Proceedings - IEEE International
  Conference on Robotics and Automation}, \bibinfo{publisher}{IEEE},
  \bibinfo{address}{Brisbane, Australia}. pp. \bibinfo{pages}{4701 -- 4708}.
%Type = Inproceedings
\bibitem[{{Boursinos} and {Koutsoukos}(2020)}]{boursinos2020trusted}
\bibinfo{author}{{Boursinos}, D.}, \bibinfo{author}{{Koutsoukos}, X.},
  \bibinfo{year}{2020}.
\newblock \bibinfo{title}{Trusted confidence bounds for learning enabled
  cyber-physical systems}, in: \bibinfo{booktitle}{2020 IEEE Security and
  Privacy Workshops (SPW)}, \bibinfo{publisher}{IEEE}, \bibinfo{address}{San
  Francisco, CA}. pp. \bibinfo{pages}{228--233}.
\newblock \DOIprefix\doi{10.1109/SPW50608.2020.00053}.
%Type = Article
\bibitem[{Bu et~al.(2011)Bu, Wang, Chen, Wang, Zhang, Zhao and Li}]{Bu2011}
\bibinfo{author}{Bu, L.}, \bibinfo{author}{Wang, Q.}, \bibinfo{author}{Chen,
  X.}, \bibinfo{author}{Wang, L.}, \bibinfo{author}{Zhang, T.},
  \bibinfo{author}{Zhao, J.}, \bibinfo{author}{Li, X.}, \bibinfo{year}{2011}.
\newblock \bibinfo{title}{Toward online hybrid systems model checking of
  cyber-physical systems' time-bounded short-run behavior}.
\newblock \bibinfo{journal}{SIGBED Rev.} \bibinfo{volume}{8},
  \bibinfo{pages}{7–10}.
\newblock \URLprefix \url{https://doi.org/10.1145/2000367.2000368},
  \DOIprefix\doi{10.1145/2000367.2000368}.
%Type = Article
\bibitem[{{Bu} et~al.(2020){Bu}, {Wang}, {Ren}, {Xing} and {Li}}]{BU2020}
\bibinfo{author}{{Bu}, L.}, \bibinfo{author}{{Wang}, Q.},
  \bibinfo{author}{{Ren}, X.}, \bibinfo{author}{{Xing}, S.},
  \bibinfo{author}{{Li}, X.}, \bibinfo{year}{2020}.
\newblock \bibinfo{title}{Scenario-based online reachability validation for cps
  fault prediction}.
\newblock \bibinfo{journal}{IEEE Transactions on Computer-Aided Design of
  Integrated Circuits and Systems} \bibinfo{volume}{39},
  \bibinfo{pages}{2081--2094}.
\newblock \DOIprefix\doi{10.1109/TCAD.2019.2935062}.
%Type = Article
\bibitem[{Butler and Finelli(1993)}]{Butler1993}
\bibinfo{author}{Butler, R.W.}, \bibinfo{author}{Finelli, G.B.},
  \bibinfo{year}{1993}.
\newblock \bibinfo{title}{The infeasibility of quantifying the reliability of
  life-critical real-time software}.
\newblock \bibinfo{journal}{IEEE Trans. Softw. Eng.} \bibinfo{volume}{19},
  \bibinfo{pages}{3–12}.
\newblock \DOIprefix\doi{10.1109/32.210303}.
%Type = Inproceedings
\bibitem[{{Chen} et~al.(2015){Chen}, {Hu}, {Mackin}, {Fisac} and
  {Tomlin}}]{Chen2016}
\bibinfo{author}{{Chen}, M.}, \bibinfo{author}{{Hu}, Q.},
  \bibinfo{author}{{Mackin}, C.}, \bibinfo{author}{{Fisac}, J.F.},
  \bibinfo{author}{{Tomlin}, C.J.}, \bibinfo{year}{2015}.
\newblock \bibinfo{title}{Safe platooning of unmanned aerial vehicles via
  reachability}, in: \bibinfo{booktitle}{2015 54th IEEE Conference on Decision
  and Control (CDC)}, \bibinfo{publisher}{IEEE}, \bibinfo{address}{Osaka,
  Japan}. pp. \bibinfo{pages}{4695--4701}.
\newblock \DOIprefix\doi{10.1109/CDC.2015.7402951}.
%Type = Inproceedings
\bibitem[{{Chen} et~al.(2012){Chen}, {Ábrahám} and
  {Sankaranarayanan}}]{Chen2012}
\bibinfo{author}{{Chen}, X.}, \bibinfo{author}{{Ábrahám}, E.},
  \bibinfo{author}{{Sankaranarayanan}, S.}, \bibinfo{year}{2012}.
\newblock \bibinfo{title}{Taylor model flowpipe construction for non-linear
  hybrid systems}, in: \bibinfo{booktitle}{2012 IEEE 33rd Real-Time Systems
  Symposium}, pp. \bibinfo{pages}{183--192}.
\newblock \DOIprefix\doi{10.1109/RTSS.2012.70}.
%Type = Techreport
\bibitem[{Clark et~al.(2013)Clark, Koutsoukos, Porter, Kumar, Pappas, Sokolsky,
  Lee and Pike}]{Clark2013}
\bibinfo{author}{Clark, M.}, \bibinfo{author}{Koutsoukos, X.},
  \bibinfo{author}{Porter, J.}, \bibinfo{author}{Kumar, R.},
  \bibinfo{author}{Pappas, G.J.}, \bibinfo{author}{Sokolsky, O.},
  \bibinfo{author}{Lee, I.}, \bibinfo{author}{Pike, L.}, \bibinfo{year}{2013}.
\newblock \bibinfo{title}{A Study on Run Time Assurance for Complex Cyber
  Physical Systems}.
\newblock \bibinfo{type}{Technical Report}. Aerospace Systems Directorate, Air
  Force Research Lab. \bibinfo{address}{Wright-Patterson Air Force Base}.
%Type = Inbook
\bibitem[{Clarke et~al.(2012)Clarke, Klieber, Nov{\'a}{\v{c}}ek and
  Zuliani}]{Valmari1998}
\bibinfo{author}{Clarke, E.M.}, \bibinfo{author}{Klieber, W.},
  \bibinfo{author}{Nov{\'a}{\v{c}}ek, M.}, \bibinfo{author}{Zuliani, P.},
  \bibinfo{year}{2012}.
\newblock \bibinfo{title}{Model Checking and the State Explosion Problem}.
  \bibinfo{publisher}{Springer Berlin Heidelberg}, \bibinfo{address}{Berlin,
  Heidelberg}. chapter~\bibinfo{chapter}{1}.
\newblock pp. \bibinfo{pages}{1--30}.
\newblock \URLprefix \url{https://doi.org/10.1007/978-3-642-35746-6_1},
  \DOIprefix\doi{10.1007/978-3-642-35746-6_1}.
%Type = Techreport
\bibitem[{Coulter(1992)}]{Coulter-1992-13338}
\bibinfo{author}{Coulter, R.C.}, \bibinfo{year}{1992}.
\newblock \bibinfo{title}{Implementation of the Pure Pursuit Path Tracking
  Algorithm}.
\newblock \bibinfo{type}{Technical Report} \bibinfo{number}{CMU-RI-TR-92-01}.
  Carnegie Mellon University. \bibinfo{address}{Pittsburgh, PA}.
%Type = Inproceedings
\bibitem[{{Crenshaw} et~al.(2007){Crenshaw}, {Gunter}, {Robinson}, {Sha} and
  {Kumar}}]{Crenshaw2007}
\bibinfo{author}{{Crenshaw}, T.L.}, \bibinfo{author}{{Gunter}, E.},
  \bibinfo{author}{{Robinson}, C.L.}, \bibinfo{author}{{Sha}, L.},
  \bibinfo{author}{{Kumar}, P.R.}, \bibinfo{year}{2007}.
\newblock \bibinfo{title}{The simplex reference model: Limiting
  fault-propagation due to unreliable components in cyber-physical system
  architectures}, in: \bibinfo{booktitle}{28th IEEE International Real-Time
  Systems Symposium (RTSS 2007)}, \bibinfo{publisher}{IEEE},
  \bibinfo{address}{Tucson, AZ}. pp. \bibinfo{pages}{400--412}.
%Type = Inproceedings
\bibitem[{Dang and Maler(1998)}]{DangMaler1998}
\bibinfo{author}{Dang, T.}, \bibinfo{author}{Maler, O.}, \bibinfo{year}{1998}.
\newblock \bibinfo{title}{Reachability analysis via face lifting}, in:
  \bibinfo{booktitle}{Proceedings of the First International Workshop on Hybrid
  Systems: Computation and Control}, \bibinfo{publisher}{Springer-Verlag},
  \bibinfo{address}{Berlin, Heidelberg}. p. \bibinfo{pages}{96–109}.
%Type = Phdthesis
\bibitem[{Dang(2000)}]{dang2000}
\bibinfo{author}{Dang, T.X.T.}, \bibinfo{year}{2000}.
\newblock \bibinfo{title}{{Verification and Synthesis of Hybrid Systems}}.
\newblock \bibinfo{type}{Theses}. {Institut National Polytechnique de Grenoble
  - INPG}.
%Type = Inproceedings
\bibitem[{Daws and Tripakis(1998)}]{Daws1998}
\bibinfo{author}{Daws, C.}, \bibinfo{author}{Tripakis, S.},
  \bibinfo{year}{1998}.
\newblock \bibinfo{title}{Model checking of real-time reachability properties
  using abstractions}, in: \bibinfo{editor}{Steffen, B.} (Ed.),
  \bibinfo{booktitle}{Tools and Algorithms for the Construction and Analysis of
  Systems}, \bibinfo{publisher}{Springer Berlin Heidelberg},
  \bibinfo{address}{Berlin, Heidelberg}. pp. \bibinfo{pages}{313--329}.
%Type = Inproceedings
\bibitem[{Desai et~al.(2017a)Desai, Dreossi and Seshia}]{DesaiRV2017}
\bibinfo{author}{Desai, A.}, \bibinfo{author}{Dreossi, T.},
  \bibinfo{author}{Seshia, S.A.}, \bibinfo{year}{2017}a.
\newblock \bibinfo{title}{Combining model checking and runtime verification for
  safe robotics}, in: \bibinfo{editor}{Lahiri, S.}, \bibinfo{editor}{Reger, G.}
  (Eds.), \bibinfo{booktitle}{Runtime Verification},
  \bibinfo{publisher}{Springer International Publishing},
  \bibinfo{address}{Cham}. pp. \bibinfo{pages}{172--189}.
%Type = Inproceedings
\bibitem[{Desai et~al.(2019)Desai, Ghosh, Seshia, Shankar and
  Tiwari}]{Desai2018}
\bibinfo{author}{Desai, A.}, \bibinfo{author}{Ghosh, S.},
  \bibinfo{author}{Seshia, S.A.}, \bibinfo{author}{Shankar, N.},
  \bibinfo{author}{Tiwari, A.}, \bibinfo{year}{2019}.
\newblock \bibinfo{title}{Soter: A runtime assurance framework for programming
  safe robotics systems}, in: \bibinfo{booktitle}{Proceedings - 49th Annual
  IEEE/IFIP International Conference on Dependable Systems and Networks, DSN
  2019}, \bibinfo{publisher}{IEEE}, \bibinfo{address}{Portland, OR}. pp.
  \bibinfo{pages}{138 -- 150}.
\newblock \DOIprefix\doi{10.1109/DSN.2019.00027}.
%Type = Inproceedings
\bibitem[{Desai et~al.(2017b)Desai, Saha, Yang, Qadeer and Seshia}]{Desai2017}
\bibinfo{author}{Desai, A.}, \bibinfo{author}{Saha, I.}, \bibinfo{author}{Yang,
  J.}, \bibinfo{author}{Qadeer, S.}, \bibinfo{author}{Seshia, S.A.},
  \bibinfo{year}{2017}b.
\newblock \bibinfo{title}{Drona: A framework for safe distributed mobile
  robotics}, in: \bibinfo{booktitle}{Proceedings of the 8th International
  Conference on Cyber-Physical Systems}, \bibinfo{publisher}{Association for
  Computing Machinery}, \bibinfo{address}{New York, NY, USA}. p.
  \bibinfo{pages}{239–248}.
\newblock \URLprefix \url{https://doi.org/10.1145/3055004.3055022},
  \DOIprefix\doi{10.1145/3055004.3055022}.
%Type = Article
\bibitem[{Deshmukh et~al.(2017)Deshmukh, Donzé, Ghosh, Jin, Juniwal and
  Seshia}]{Deshmukh2015}
\bibinfo{author}{Deshmukh, J.V.}, \bibinfo{author}{Donzé, A.},
  \bibinfo{author}{Ghosh, S.}, \bibinfo{author}{Jin, X.},
  \bibinfo{author}{Juniwal, G.}, \bibinfo{author}{Seshia, S.A.},
  \bibinfo{year}{2017}.
\newblock \bibinfo{title}{Robust online monitoring of signal temporal logic}.
\newblock \bibinfo{journal}{Formal Methods in System Design}
  \bibinfo{volume}{51}.
\newblock \DOIprefix\doi{10.1007/s10703-017-0286-7}.
%Type = Article
\bibitem[{{Desoer} and {Wing}(1961)}]{Desoer}
\bibinfo{author}{{Desoer}, C.}, \bibinfo{author}{{Wing}, J.},
  \bibinfo{year}{1961}.
\newblock \bibinfo{title}{A minimal time discrete system}.
\newblock \bibinfo{journal}{IRE Transactions on Automatic Control}
  \bibinfo{volume}{6}, \bibinfo{pages}{111--125}.
\newblock \DOIprefix\doi{10.1109/TAC.1961.1105183}.
%Type = Article
\bibitem[{Devi et~al.(2020)Devi, Malarvezhi, Dayana and
  Vadivukkarasi}]{Devi2020ACS}
\bibinfo{author}{Devi, S.}, \bibinfo{author}{Malarvezhi, P.},
  \bibinfo{author}{Dayana, R.}, \bibinfo{author}{Vadivukkarasi, K.},
  \bibinfo{year}{2020}.
\newblock \bibinfo{title}{A comprehensive survey on autonomous driving cars: A
  perspective view}.
\newblock \bibinfo{journal}{Wirel. Pers. Commun.} \bibinfo{volume}{114},
  \bibinfo{pages}{2121--2133}.
%Type = Inproceedings
\bibitem[{Dhinakaran et~al.(2018)Dhinakaran, Chen, Chou, Shih and
  Tomlin}]{dhinakaran2017hybrid}
\bibinfo{author}{Dhinakaran, A.}, \bibinfo{author}{Chen, M.},
  \bibinfo{author}{Chou, G.}, \bibinfo{author}{Shih, J.C.},
  \bibinfo{author}{Tomlin, C.J.}, \bibinfo{year}{2018}.
\newblock \bibinfo{title}{A hybrid framework for multi-vehicle collision
  avoidance}, in: \bibinfo{booktitle}{2017 IEEE 56th Annual Conference on
  Decision and Control, CDC 2017}, \bibinfo{publisher}{IEEE},
  \bibinfo{address}{Melbourne, Australia}. pp. \bibinfo{pages}{2979 -- 2984}.
\newblock \DOIprefix\doi{10.1109/CDC.2017.8264092}.
%Type = Inproceedings
\bibitem[{Doty et~al.()Doty, Doty, Camberos and Yerkes}]{DotyUncertainty}
\bibinfo{author}{Doty, A.}, \bibinfo{author}{Doty, J.},
  \bibinfo{author}{Camberos, J.}, \bibinfo{author}{Yerkes, K.}, .
\newblock \bibinfo{title}{Nonlinear uncertainty quantification, sensitivity
  analysis, and uncertainty propagation of a dynamic electrical circuit}, in:
  \bibinfo{booktitle}{51st AIAA Aerospace Sciences Meeting including the New
  Horizons Forum and Aerospace Exposition}, pp. \bibinfo{pages}{13110--13234}.
\newblock \URLprefix \url{https://arc.aiaa.org/doi/abs/10.2514/6.2013-883},
  \DOIprefix\doi{10.2514/6.2013-883},
  \href{http://arxiv.org/abs/https://arc.aiaa.org/doi/pdf/10.2514/6.2013-883}{\tt
  arXiv:https://arc.aiaa.org/doi/pdf/10.2514/6.2013-883}.
%Type = Book
\bibitem[{Doyen et~al.(2018)Doyen, Frehse, Pappas and Platzer}]{Doyen2018}
\bibinfo{author}{Doyen, L.}, \bibinfo{author}{Frehse, G.},
  \bibinfo{author}{Pappas, G.J.}, \bibinfo{author}{Platzer, A.},
  \bibinfo{year}{2018}.
\newblock \bibinfo{title}{Verification of Hybrid Systems}.
\newblock \bibinfo{publisher}{Springer International Publishing},
  \bibinfo{address}{Cham}.
\newblock \DOIprefix\doi{10.1007/978-3-319-10575-8_30}.
%Type = Inproceedings
\bibitem[{Dunlap et~al.(2022)Dunlap, Mote, Delsing and Hobbs}]{dunlap2021Safe}
\bibinfo{author}{Dunlap, K.}, \bibinfo{author}{Mote, M.},
  \bibinfo{author}{Delsing, K.}, \bibinfo{author}{Hobbs, K.L.},
  \bibinfo{year}{2022}.
\newblock \bibinfo{title}{Run time assured reinforcement learning for safe
  satellite docking}, in: \bibinfo{booktitle}{2022 AIAA SciTech Forum}, pp.
  \bibinfo{pages}{1--20}.
%Type = Inproceedings
\bibitem[{Ferrando et~al.(2020)Ferrando, Cardoso, Fisher, Ancona, Franceschini
  and Mascardi}]{Angelo2020}
\bibinfo{author}{Ferrando, A.}, \bibinfo{author}{Cardoso, R.C.},
  \bibinfo{author}{Fisher, M.}, \bibinfo{author}{Ancona, D.},
  \bibinfo{author}{Franceschini, L.}, \bibinfo{author}{Mascardi, V.},
  \bibinfo{year}{2020}.
\newblock \bibinfo{title}{Rosmonitoring: A runtime verification framework for
  ros}, in: \bibinfo{editor}{Mohammad, A.}, \bibinfo{editor}{Dong, X.},
  \bibinfo{editor}{Russo, M.} (Eds.), \bibinfo{booktitle}{Towards Autonomous
  Robotic Systems}, \bibinfo{publisher}{Springer International Publishing},
  \bibinfo{address}{Cham}. pp. \bibinfo{pages}{387--399}.
%Type = Article
\bibitem[{Fisac et~al.(2019)Fisac, Akametalu, Zeilinger, Kaynama, Gillula and
  Tomlin}]{Fisac2017}
\bibinfo{author}{Fisac, J.F.}, \bibinfo{author}{Akametalu, A.K.},
  \bibinfo{author}{Zeilinger, M.N.}, \bibinfo{author}{Kaynama, S.},
  \bibinfo{author}{Gillula, J.}, \bibinfo{author}{Tomlin, C.J.},
  \bibinfo{year}{2019}.
\newblock \bibinfo{title}{A general safety framework for learning-based control
  in uncertain robotic systems}.
\newblock \bibinfo{journal}{IEEE Transactions on Automatic Control}
  \bibinfo{volume}{64}, \bibinfo{pages}{2737 -- 2752}.
\newblock \DOIprefix\doi{10.1109/TAC.2018.2876389}.
%Type = Inproceedings
\bibitem[{{Fraichard}(2007)}]{Fraichard2007}
\bibinfo{author}{{Fraichard}, T.}, \bibinfo{year}{2007}.
\newblock \bibinfo{title}{A short paper about motion safety}, in:
  \bibinfo{booktitle}{Proceedings 2007 IEEE International Conference on
  Robotics and Automation}, pp. \bibinfo{pages}{1140--1145}.
\newblock \DOIprefix\doi{10.1109/ROBOT.2007.363138}.
%Type = Inproceedings
\bibitem[{Fraichard and Asama(2003)}]{Fraichard}
\bibinfo{author}{Fraichard, T.}, \bibinfo{author}{Asama, H.},
  \bibinfo{year}{2003}.
\newblock \bibinfo{title}{Inevitable collision states. a step towards safer
  robots?}, in: \bibinfo{booktitle}{Proceedings 2003 IEEE/RSJ International
  Conference on Intelligent Robots and Systems (IROS 2003) (Cat.
  No.03CH37453)}, pp. \bibinfo{pages}{388--393 vol.1}.
\newblock \DOIprefix\doi{10.1109/IROS.2003.1250659}.
%Type = Inproceedings
\bibitem[{Girão et~al.(2016)Girão, Asvadi, Peixoto and Nunes}]{Girao2016}
\bibinfo{author}{Girão, P.}, \bibinfo{author}{Asvadi, A.},
  \bibinfo{author}{Peixoto, P.}, \bibinfo{author}{Nunes, U.},
  \bibinfo{year}{2016}.
\newblock \bibinfo{title}{3d object tracking in driving environment: A short
  review and a benchmark dataset}, in: \bibinfo{booktitle}{2016 IEEE 19th
  International Conference on Intelligent Transportation Systems (ITSC)}, pp.
  \bibinfo{pages}{7--12}.
\newblock \DOIprefix\doi{10.1109/ITSC.2016.7795523}.
%Type = Misc
\bibitem[{Gonzalez et~al.(2020)Gonzalez, Collins, Geretti, Bresolin and
  Villa}]{Gonzalez2020}
\bibinfo{author}{Gonzalez, S.Z.}, \bibinfo{author}{Collins, P.},
  \bibinfo{author}{Geretti, L.}, \bibinfo{author}{Bresolin, D.},
  \bibinfo{author}{Villa, T.}, \bibinfo{year}{2020}.
\newblock \bibinfo{title}{Higher order method for differential inclusions}.
\newblock \URLprefix \url{https://arxiv.org/abs/2001.11330},
  \DOIprefix\doi{10.48550/ARXIV.2001.11330}.
%Type = Inproceedings
\bibitem[{{Gurriet} et~al.(2018){Gurriet}, {Mote}, {Ames} and
  {Feron}}]{Gurriet2018}
\bibinfo{author}{{Gurriet}, T.}, \bibinfo{author}{{Mote}, M.},
  \bibinfo{author}{{Ames}, A.D.}, \bibinfo{author}{{Feron}, E.},
  \bibinfo{year}{2018}.
\newblock \bibinfo{title}{An online approach to active set invariance}, in:
  \bibinfo{booktitle}{2018 IEEE Conference on Decision and Control (CDC)},
  \bibinfo{publisher}{IEEE}, \bibinfo{address}{Miami, FL}. pp.
  \bibinfo{pages}{3592--3599}.
\newblock \DOIprefix\doi{10.1109/CDC.2018.8619139}.
%Type = Inproceedings
\bibitem[{Herbert et~al.(2019)Herbert, Bansal, Ghosh and Tomlin}]{Herbert2019}
\bibinfo{author}{Herbert, S.L.}, \bibinfo{author}{Bansal, S.},
  \bibinfo{author}{Ghosh, S.}, \bibinfo{author}{Tomlin, C.J.},
  \bibinfo{year}{2019}.
\newblock \bibinfo{title}{Reachability-based safety guarantees using efficient
  initializations}, in: \bibinfo{booktitle}{Proceedings of the IEEE Conference
  on Decision and Control}, \bibinfo{publisher}{IEEE}, \bibinfo{address}{Nice,
  France}. pp. \bibinfo{pages}{4810 -- 4816}.
%Type = Misc
\bibitem[{Holmes et~al.(2020)Holmes, Kousik, Zhang, Raz, Barbalata,
  Johnson-Roberson and Vasudevan}]{holmes2020reachable}
\bibinfo{author}{Holmes, P.}, \bibinfo{author}{Kousik, S.},
  \bibinfo{author}{Zhang, B.}, \bibinfo{author}{Raz, D.},
  \bibinfo{author}{Barbalata, C.}, \bibinfo{author}{Johnson-Roberson, M.},
  \bibinfo{author}{Vasudevan, R.}, \bibinfo{year}{2020}.
\newblock \bibinfo{title}{Reachable sets for safe, real-time manipulator
  trajectory design}.
\newblock \href{http://arxiv.org/abs/2002.01591}{\tt arXiv:2002.01591}.
%Type = Inproceedings
\bibitem[{Huang et~al.(2014)Huang, Erdogan, Zhang, Moore, Luo, Sundaresan and
  Rosu}]{Huang2014}
\bibinfo{author}{Huang, J.}, \bibinfo{author}{Erdogan, C.},
  \bibinfo{author}{Zhang, Y.}, \bibinfo{author}{Moore, B.},
  \bibinfo{author}{Luo, Q.}, \bibinfo{author}{Sundaresan, A.},
  \bibinfo{author}{Rosu, G.}, \bibinfo{year}{2014}.
\newblock \bibinfo{title}{Rosrv: Runtime verification for robots}, in:
  \bibinfo{editor}{Bonakdarpour, B.}, \bibinfo{editor}{Smolka, S.A.} (Eds.),
  \bibinfo{booktitle}{Runtime Verification}, \bibinfo{publisher}{Springer
  International Publishing}, \bibinfo{address}{Cham}. pp.
  \bibinfo{pages}{247--254}.
%Type = Article
\bibitem[{H{\"u}llermeier and Waegeman(2021)}]{Eyke2021}
\bibinfo{author}{H{\"u}llermeier, E.}, \bibinfo{author}{Waegeman, W.},
  \bibinfo{year}{2021}.
\newblock \bibinfo{title}{Aleatoric and epistemic uncertainty in machine
  learning: an introduction to concepts and methods}.
\newblock \bibinfo{journal}{Machine Learning} \bibinfo{volume}{110},
  \bibinfo{pages}{457--506}.
\newblock \URLprefix \url{https://doi.org/10.1007/s10994-021-05946-3},
  \DOIprefix\doi{10.1007/s10994-021-05946-3}.
%Type = Article
\bibitem[{Hussein et~al.(2017)Hussein, Gaber, Elyan and
  Jayne}]{Hussein2017ImitationL}
\bibinfo{author}{Hussein, A.}, \bibinfo{author}{Gaber, M.},
  \bibinfo{author}{Elyan, E.}, \bibinfo{author}{Jayne, C.},
  \bibinfo{year}{2017}.
\newblock \bibinfo{title}{Imitation learning}.
\newblock \bibinfo{journal}{ACM Computing Surveys (CSUR)} \bibinfo{volume}{50},
  \bibinfo{pages}{1 -- 35}.
%Type = Inbook
\bibitem[{Ivanov et~al.(2020)Ivanov, Carpenter, Weimer, Alur, Pappas and
  Lee}]{ivanov2020case}
\bibinfo{author}{Ivanov, R.}, \bibinfo{author}{Carpenter, T.J.},
  \bibinfo{author}{Weimer, J.}, \bibinfo{author}{Alur, R.},
  \bibinfo{author}{Pappas, G.J.}, \bibinfo{author}{Lee, I.},
  \bibinfo{year}{2020}.
\newblock \bibinfo{title}{Case Study: Verifying the Safety of an Autonomous
  Racing Car with a Neural Network Controller}. \bibinfo{publisher}{Association
  for Computing Machinery}, \bibinfo{address}{New York, NY, USA}.
  chapter~\bibinfo{chapter}{28}.
\newblock pp. \bibinfo{pages}{1--7}.
\newblock \URLprefix \url{https://doi.org/10.1145/3365365.3382216}.
%Type = Inproceedings
\bibitem[{Jha et~al.(2020)Jha, Rushby and Shankar}]{Jha2020}
\bibinfo{author}{Jha, S.}, \bibinfo{author}{Rushby, J.},
  \bibinfo{author}{Shankar, N.}, \bibinfo{year}{2020}.
\newblock \bibinfo{title}{Model-centered assurance for autonomous systems}, in:
  \bibinfo{editor}{Casimiro, A.}, \bibinfo{editor}{Ortmeier, F.},
  \bibinfo{editor}{Bitsch, F.}, \bibinfo{editor}{Ferreira, P.} (Eds.),
  \bibinfo{booktitle}{Computer Safety, Reliability, and Security},
  \bibinfo{publisher}{Springer International Publishing},
  \bibinfo{address}{Cham}. pp. \bibinfo{pages}{228--243}.
%Type = Article
\bibitem[{Johnson et~al.(2016)Johnson, Bak, Caccamo and Sha}]{Johnson2016}
\bibinfo{author}{Johnson, T.T.}, \bibinfo{author}{Bak, S.},
  \bibinfo{author}{Caccamo, M.}, \bibinfo{author}{Sha, L.},
  \bibinfo{year}{2016}.
\newblock \bibinfo{title}{Real-time reachability for verified simplex design}.
\newblock \bibinfo{journal}{ACM Trans. Embed. Comput. Syst.}
  \bibinfo{volume}{15}.
\newblock \URLprefix \url{https://doi.org/10.1145/2723871},
  \DOIprefix\doi{10.1145/2723871}.
%Type = Inproceedings
\bibitem[{{Kerr} and {Nickels}(2012)}]{ROS}
\bibinfo{author}{{Kerr}, J.}, \bibinfo{author}{{Nickels}, K.},
  \bibinfo{year}{2012}.
\newblock \bibinfo{title}{Robot operating systems: Bridging the gap between
  human and robot}, in: \bibinfo{booktitle}{Proceedings of the 2012 44th
  Southeastern Symposium on System Theory (SSST)}, pp.
  \bibinfo{pages}{99--104}.
%Type = Inproceedings
\bibitem[{{Koenig} and {Howard}(2004)}]{Gazebo}
\bibinfo{author}{{Koenig}, N.}, \bibinfo{author}{{Howard}, A.},
  \bibinfo{year}{2004}.
\newblock \bibinfo{title}{Design and use paradigms for gazebo, an open-source
  multi-robot simulator}, in: \bibinfo{booktitle}{2004 IEEE/RSJ International
  Conference on Intelligent Robots and Systems (IROS) (IEEE Cat.
  No.04CH37566)}, pp. \bibinfo{pages}{2149--2154 vol.3}.
%Type = Article
\bibitem[{Krizhevsky et~al.(2017)Krizhevsky, Sutskever and
  Hinton}]{AlexNet2012}
\bibinfo{author}{Krizhevsky, A.}, \bibinfo{author}{Sutskever, I.},
  \bibinfo{author}{Hinton, G.E.}, \bibinfo{year}{2017}.
\newblock \bibinfo{title}{Imagenet classification with deep convolutional
  neural networks}.
\newblock \bibinfo{journal}{Commun. ACM} \bibinfo{volume}{60},
  \bibinfo{pages}{84–90}.
\newblock \DOIprefix\doi{10.1145/3065386}.
%Type = Article
\bibitem[{{Kwon} and {Hwang}(2018)}]{Kwon2018}
\bibinfo{author}{{Kwon}, C.}, \bibinfo{author}{{Hwang}, I.},
  \bibinfo{year}{2018}.
\newblock \bibinfo{title}{Reachability analysis for safety assurance of
  cyber-physical systems against cyber attacks}.
\newblock \bibinfo{journal}{IEEE Transactions on Automatic Control}
  \bibinfo{volume}{63}, \bibinfo{pages}{2272--2279}.
%Type = Article
\bibitem[{Lafferriere et~al.(1999)Lafferriere, Pappas and Yovine}]{LAFFERRIERE}
\bibinfo{author}{Lafferriere, G.}, \bibinfo{author}{Pappas, G.J.},
  \bibinfo{author}{Yovine, S.}, \bibinfo{year}{1999}.
\newblock \bibinfo{title}{Reachability computation for linear hybrid systems}.
\newblock \bibinfo{journal}{IFAC Proceedings Volumes} \bibinfo{volume}{32},
  \bibinfo{pages}{2137--2142}.
\newblock \DOIprefix\doi{https://doi.org/10.1016/S1474-6670(17)56362-4}.
  \bibinfo{note}{14th IFAC World Congress 1999, Beijing, Chia, 5-9 July}.
%Type = Article
\bibitem[{Leung et~al.(2020)Leung, Schmerling, Zhang, Chen, Talbot, Gerdes and
  Pavone}]{LeungReach2020}
\bibinfo{author}{Leung, K.}, \bibinfo{author}{Schmerling, E.},
  \bibinfo{author}{Zhang, M.}, \bibinfo{author}{Chen, M.},
  \bibinfo{author}{Talbot, J.}, \bibinfo{author}{Gerdes, J.C.},
  \bibinfo{author}{Pavone, M.}, \bibinfo{year}{2020}.
\newblock \bibinfo{title}{On infusing reachability-based safety assurance
  within probabilistic planning frameworks for human-robot vehicle
  interactions}.
\newblock \bibinfo{journal}{The International Journal of Robotics Research}
  \bibinfo{volume}{39}, \bibinfo{pages}{1326--1345}.
%Type = Article
\bibitem[{{Li} et~al.(2014){Li}, {Tan}, {Wang}, {Bu}, {Cao} and {Liu}}]{Ti2014}
\bibinfo{author}{{Li}, T.}, \bibinfo{author}{{Tan}, F.},
  \bibinfo{author}{{Wang}, Q.}, \bibinfo{author}{{Bu}, L.},
  \bibinfo{author}{{Cao}, J.}, \bibinfo{author}{{Liu}, X.},
  \bibinfo{year}{2014}.
\newblock \bibinfo{title}{From offline toward real time: A hybrid systems model
  checking and cps codesign approach for medical device plug-and-play
  collaborations}.
\newblock \bibinfo{journal}{IEEE Transactions on Parallel and Distributed
  Systems} \bibinfo{volume}{25}, \bibinfo{pages}{642--652}.
\newblock \DOIprefix\doi{10.1109/TPDS.2013.50}.
%Type = Inproceedings
\bibitem[{Lin et~al.(2020)Lin, Chen, Khurana and Dolan}]{Lin2020}
\bibinfo{author}{Lin, Q.}, \bibinfo{author}{Chen, X.},
  \bibinfo{author}{Khurana, A.}, \bibinfo{author}{Dolan, J.},
  \bibinfo{year}{2020}.
\newblock \bibinfo{title}{Reachflow: An online safety assurance framework for
  waypoint-following of self-driving cars}, in:
  \bibinfo{booktitle}{International Conference on Intelligent Robots and
  systems (IROS)}, \bibinfo{publisher}{IEEE}, \bibinfo{address}{Las Vegas,
  Nevada}. pp. \bibinfo{pages}{6627 -- 6632}.
%Type = Article
\bibitem[{Liu et~al.(2019)Liu, Arnon, Lazarus, Barrett and
  Kochenderfer}]{Liu2019}
\bibinfo{author}{Liu, C.}, \bibinfo{author}{Arnon, T.},
  \bibinfo{author}{Lazarus, C.}, \bibinfo{author}{Barrett, C.W.},
  \bibinfo{author}{Kochenderfer, M.J.}, \bibinfo{year}{2019}.
\newblock \bibinfo{title}{Algorithms for verifying deep neural networks}.
\newblock \bibinfo{journal}{CoRR} \bibinfo{volume}{abs/1903.06758}.
\newblock \href{http://arxiv.org/abs/1903.06758}{\tt arXiv:1903.06758}.
%Type = Inproceedings
\bibitem[{Liu et~al.(1991)Liu, Lin, Shih, Yu, Chung and Zhao}]{Liu1991}
\bibinfo{author}{Liu, J.W.S.}, \bibinfo{author}{Lin, K.J.},
  \bibinfo{author}{Shih, W.K.}, \bibinfo{author}{Yu, A.C.},
  \bibinfo{author}{Chung, J.Y.}, \bibinfo{author}{Zhao, W.},
  \bibinfo{year}{1991}.
\newblock \bibinfo{title}{Algorithms for scheduling imprecise computations},
  in: \bibinfo{editor}{van Tilborg, A.M.}, \bibinfo{editor}{Koob, G.M.} (Eds.),
  \bibinfo{booktitle}{Foundations of Real-Time Computing: Scheduling and
  Resource Management}, \bibinfo{publisher}{Springer US},
  \bibinfo{address}{Boston, MA}. pp. \bibinfo{pages}{203--249}.
\newblock \DOIprefix\doi{10.1007/978-1-4615-3956-8_8}.
%Type = Article
\bibitem[{Ljung et~al.(2004)Ljung, Zhang, Lindskog and Juditski}]{LJUNG2004399}
\bibinfo{author}{Ljung, L.}, \bibinfo{author}{Zhang, Q.},
  \bibinfo{author}{Lindskog, P.}, \bibinfo{author}{Juditski, A.},
  \bibinfo{year}{2004}.
\newblock \bibinfo{title}{Estimation of grey box and black box models for
  non-linear circuit data}.
\newblock \bibinfo{journal}{IFAC Proceedings Volumes} \bibinfo{volume}{37},
  \bibinfo{pages}{399--404}.
\newblock \URLprefix
  \url{https://www.sciencedirect.com/science/article/pii/S1474667017312569},
  \DOIprefix\doi{https://doi.org/10.1016/S1474-6670(17)31256-9}.
  \bibinfo{note}{6th IFAC Symposium on Nonlinear Control Systems 2004 (NOLCOS
  2004), Stuttgart, Germany, 1-3 September, 2004}.
%Type = Article
\bibitem[{Macfarlane and Stroila(2016)}]{Macfarlane2016}
\bibinfo{author}{Macfarlane, J.}, \bibinfo{author}{Stroila, M.},
  \bibinfo{year}{2016}.
\newblock \bibinfo{title}{Addressing the uncertainties in autonomous driving}.
\newblock \bibinfo{journal}{SIGSPATIAL Special} \bibinfo{volume}{8},
  \bibinfo{pages}{35–40}.
\newblock \URLprefix \url{https://doi.org/10.1145/3024087.3024092},
  \DOIprefix\doi{10.1145/3024087.3024092}.
%Type = Inproceedings
\bibitem[{Majumdar and Pavone(2020)}]{Majumdar2017}
\bibinfo{author}{Majumdar, A.}, \bibinfo{author}{Pavone, M.},
  \bibinfo{year}{2020}.
\newblock \bibinfo{title}{How should a robot assess risk? towards an axiomatic
  theory of risk in robotics}, in: \bibinfo{editor}{Amato, N.M.},
  \bibinfo{editor}{Hager, G.}, \bibinfo{editor}{Thomas, S.},
  \bibinfo{editor}{Torres-Torriti, M.} (Eds.), \bibinfo{booktitle}{Robotics
  Research}, \bibinfo{publisher}{Springer International Publishing},
  \bibinfo{address}{Cham}. pp. \bibinfo{pages}{75--84}.
%Type = Inproceedings
\bibitem[{Maler(2013)}]{Maler14}
\bibinfo{author}{Maler, O.}, \bibinfo{year}{2013}.
\newblock \bibinfo{title}{Algorithmic verification of continuous and hybrid
  systems}, in: \bibinfo{editor}{Hol{\'{\i}}k, L.}, \bibinfo{editor}{Clemente,
  L.} (Eds.), \bibinfo{booktitle}{Proceedings 15th International Workshop on
  Verification of Infinite-State Systems, {INFINITY} 2013, Hanoi, Vietnam, 14th
  October 2013}, pp. \bibinfo{pages}{48--69}.
\newblock \DOIprefix\doi{10.4204/EPTCS.140.4}.
%Type = Inproceedings
\bibitem[{Masson et~al.(2018)Masson, Guiochet, Waeselynck, Cabrera, Cassel and
  T{\"o}rngren}]{Masson2018}
\bibinfo{author}{Masson, L.}, \bibinfo{author}{Guiochet, J.},
  \bibinfo{author}{Waeselynck, H.}, \bibinfo{author}{Cabrera, K.},
  \bibinfo{author}{Cassel, S.}, \bibinfo{author}{T{\"o}rngren, M.},
  \bibinfo{year}{2018}.
\newblock \bibinfo{title}{Tuning permissiveness of active safety monitors for
  autonomous systems}, in: \bibinfo{editor}{Dutle, A.},
  \bibinfo{editor}{Mu{\~{n}}oz, C.}, \bibinfo{editor}{Narkawicz, A.} (Eds.),
  \bibinfo{booktitle}{NASA Formal Methods}, \bibinfo{publisher}{Springer
  International Publishing}, \bibinfo{address}{Cham}. pp.
  \bibinfo{pages}{333--348}.
%Type = Article
\bibitem[{Mitsch and Platzer(2016)}]{mitsch}
\bibinfo{author}{Mitsch, S.}, \bibinfo{author}{Platzer, A.},
  \bibinfo{year}{2016}.
\newblock \bibinfo{title}{Modelplex: verified runtime validation of verified
  cyber-physical system models}.
\newblock \bibinfo{journal}{Formal Methods in System Design}
  \bibinfo{volume}{49}, \bibinfo{pages}{33--74}.
\newblock \DOIprefix\doi{10.1007/s10703-016-0241-z}.
%Type = Article
\bibitem[{{Muratore} et~al.(2021){Muratore}, {Gienger} and
  {Peters}}]{Muratore2019}
\bibinfo{author}{{Muratore}, F.}, \bibinfo{author}{{Gienger}, M.},
  \bibinfo{author}{{Peters}, J.}, \bibinfo{year}{2021}.
\newblock \bibinfo{title}{Assessing transferability from simulation to reality
  for reinforcement learning}.
\newblock \bibinfo{journal}{IEEE Transactions on Pattern Analysis and Machine
  Intelligence} \bibinfo{volume}{43}, \bibinfo{pages}{1172--1183}.
\newblock \DOIprefix\doi{10.1109/TPAMI.2019.2952353}.
%Type = Inproceedings
\bibitem[{Musau et~al.(2022)Musau, Hamilton, Manzanas~Lopez, Robinette and
  Johnson}]{Musau2022}
\bibinfo{author}{Musau, P.}, \bibinfo{author}{Hamilton, N.},
  \bibinfo{author}{Manzanas~Lopez, D.}, \bibinfo{author}{Robinette, P.},
  \bibinfo{author}{Johnson, T.}, \bibinfo{year}{2022}.
\newblock \bibinfo{title}{On using real-time reachability for the safety
  assurance of machine learning controllers}, in:
  \bibinfo{booktitle}{Proceedings 2022 IEEE International Conference on Assured
  Autonomy}, pp. \bibinfo{pages}{1--10}.
%Type = Article
\bibitem[{Nilsson et~al.(2014)Nilsson, Fredriksson and Ödblom}]{NILSSON}
\bibinfo{author}{Nilsson, J.}, \bibinfo{author}{Fredriksson, J.},
  \bibinfo{author}{Ödblom, A.C.}, \bibinfo{year}{2014}.
\newblock \bibinfo{title}{Verification of collision avoidance systems using
  reachability analysis}.
\newblock \bibinfo{journal}{IFAC Proceedings Volumes} \bibinfo{volume}{47},
  \bibinfo{pages}{10676--10681}.
\newblock \DOIprefix\doi{https://doi.org/10.3182/20140824-6-ZA-1003.01567}.
  \bibinfo{note}{19th IFAC World Congress}.
%Type = Article
\bibitem[{O'Kelly et~al.(2019)O'Kelly, Sukhil, Abbas, Harkins, Kao, Pant,
  Mangharam, Agarwal, Behl, Burgio and Bertogna}]{F1102019}
\bibinfo{author}{O'Kelly, M.}, \bibinfo{author}{Sukhil, V.},
  \bibinfo{author}{Abbas, H.}, \bibinfo{author}{Harkins, J.},
  \bibinfo{author}{Kao, C.}, \bibinfo{author}{Pant, Y.V.},
  \bibinfo{author}{Mangharam, R.}, \bibinfo{author}{Agarwal, D.},
  \bibinfo{author}{Behl, M.}, \bibinfo{author}{Burgio, P.},
  \bibinfo{author}{Bertogna, M.}, \bibinfo{year}{2019}.
\newblock \bibinfo{title}{{F1/10:} an open-source autonomous cyber-physical
  platform}.
\newblock \bibinfo{journal}{CoRR} \bibinfo{volume}{abs/1901.08567}.
\newblock \href{http://arxiv.org/abs/1901.08567}{\tt arXiv:1901.08567}.
%Type = Inproceedings
\bibitem[{O'Kelly et~al.(2020)O'Kelly, Zheng, Karthik and
  Mangharam}]{okelly2020}
\bibinfo{author}{O'Kelly, M.}, \bibinfo{author}{Zheng, H.},
  \bibinfo{author}{Karthik, D.}, \bibinfo{author}{Mangharam, R.},
  \bibinfo{year}{2020}.
\newblock \bibinfo{title}{F1tenth: An open-source evaluation environment for
  continuous control and reinforcement learning}, in:
  \bibinfo{editor}{Escalante, H.J.}, \bibinfo{editor}{Hadsell, R.} (Eds.),
  \bibinfo{booktitle}{Proceedings of the NeurIPS 2019 Competition and
  Demonstration Track}, \bibinfo{publisher}{PMLR}. pp. \bibinfo{pages}{77--89}.
\newblock \URLprefix \url{https://proceedings.mlr.press/v123/o-kelly20a.html}.
%Type = Article
\bibitem[{Orellana et~al.(2021)Orellana, Coronel, Carvajal, Delgado, Escárate
  and Agüero}]{ORELLANA2021589}
\bibinfo{author}{Orellana, R.}, \bibinfo{author}{Coronel, M.},
  \bibinfo{author}{Carvajal, R.}, \bibinfo{author}{Delgado, R.A.},
  \bibinfo{author}{Escárate, P.}, \bibinfo{author}{Agüero, J.C.},
  \bibinfo{year}{2021}.
\newblock \bibinfo{title}{On the uncertainty modelling for linear
  continuous-time systems utilising sampled data and gaussian mixture
  models⁎⁎this work was supported by anid/doctorado nacional/2017-21170804,
  anid-fondecyt grants 1211630 and 11201187, dpp at universidad técnica
  federico santa maría and the advanced center for electrical and electronic
  engineering, ac3e, basal project fb0008, anid, chile.}
\newblock \bibinfo{journal}{IFAC-PapersOnLine} \bibinfo{volume}{54},
  \bibinfo{pages}{589--594}.
\newblock \URLprefix
  \url{https://www.sciencedirect.com/science/article/pii/S2405896321011988},
  \DOIprefix\doi{https://doi.org/10.1016/j.ifacol.2021.08.424}.
  \bibinfo{note}{19th IFAC Symposium on System Identification SYSID 2021}.
%Type = Misc
\bibitem[{Otterness(2019)}]{otterness_2019}
\bibinfo{author}{Otterness, N.}, \bibinfo{year}{2019}.
\newblock \bibinfo{title}{The "disparity extender" algorithm, and f1/tenth}.
%Type = Article
\bibitem[{Pettersson(2005)}]{PETTERSSON200573}
\bibinfo{author}{Pettersson, O.}, \bibinfo{year}{2005}.
\newblock \bibinfo{title}{Execution monitoring in robotics: A survey}.
\newblock \bibinfo{journal}{Robotics and Autonomous Systems}
  \bibinfo{volume}{53}.
\newblock \DOIprefix\doi{10.1016/j.robot.2005.09.004}.
%Type = Inproceedings
\bibitem[{Phan et~al.(2020)Phan, Grosu, Jansen, Paoletti, Smolka and
  Stoller}]{Phan2020}
\bibinfo{author}{Phan, D.T.}, \bibinfo{author}{Grosu, R.},
  \bibinfo{author}{Jansen, N.}, \bibinfo{author}{Paoletti, N.},
  \bibinfo{author}{Smolka, S.A.}, \bibinfo{author}{Stoller, S.D.},
  \bibinfo{year}{2020}.
\newblock \bibinfo{title}{Neural simplex architecture}, in:
  \bibinfo{editor}{Lee, R.}, \bibinfo{editor}{Jha, S.},
  \bibinfo{editor}{Mavridou, A.} (Eds.), \bibinfo{booktitle}{NASA Formal
  Methods}, \bibinfo{publisher}{Springer International Publishing},
  \bibinfo{address}{Cham}. pp. \bibinfo{pages}{97--114}.
%Type = Inproceedings
\bibitem[{Puri and Varaiya(1994)}]{Puri1994}
\bibinfo{author}{Puri, A.}, \bibinfo{author}{Varaiya, P.},
  \bibinfo{year}{1994}.
\newblock \bibinfo{title}{Decidability of hybrid systems with rectangular
  differential inclusions}, in: \bibinfo{editor}{Dill, D.L.} (Ed.),
  \bibinfo{booktitle}{Computer Aided Verification},
  \bibinfo{publisher}{Springer Berlin Heidelberg}, \bibinfo{address}{Berlin,
  Heidelberg}. pp. \bibinfo{pages}{95--104}.
%Type = Book
\bibitem[{Rajamani(2012)}]{Rajamani2012}
\bibinfo{author}{Rajamani, R.}, \bibinfo{year}{2012}.
\newblock \bibinfo{title}{Lateral Vehicle Dynamics}.
\newblock \bibinfo{publisher}{Springer US}, \bibinfo{address}{Boston, MA}.
\newblock \DOIprefix\doi{10.1007/978-1-4614-1433-9_2}.
%Type = Article
\bibitem[{{Rakovic} et~al.(2006){Rakovic}, {Kerrigan}, {Mayne} and
  {Lygeros}}]{Rakovic}
\bibinfo{author}{{Rakovic}, S.V.}, \bibinfo{author}{{Kerrigan}, E.C.},
  \bibinfo{author}{{Mayne}, D.Q.}, \bibinfo{author}{{Lygeros}, J.},
  \bibinfo{year}{2006}.
\newblock \bibinfo{title}{Reachability analysis of discrete-time systems with
  disturbances}.
\newblock \bibinfo{journal}{IEEE Transactions on Automatic Control}
  \bibinfo{volume}{51}, \bibinfo{pages}{546--561}.
\newblock \DOIprefix\doi{10.1109/TAC.2006.872835}.
%Type = Inproceedings
\bibitem[{Sculley et~al.(2015)Sculley, Holt, Golovin, Davydov, Phillips, Ebner,
  Chaudhary, Young, Crespo and Dennison}]{Sculley2015}
\bibinfo{author}{Sculley, D.}, \bibinfo{author}{Holt, G.},
  \bibinfo{author}{Golovin, D.}, \bibinfo{author}{Davydov, E.},
  \bibinfo{author}{Phillips, T.}, \bibinfo{author}{Ebner, D.},
  \bibinfo{author}{Chaudhary, V.}, \bibinfo{author}{Young, M.},
  \bibinfo{author}{Crespo, J.F.}, \bibinfo{author}{Dennison, D.},
  \bibinfo{year}{2015}.
\newblock \bibinfo{title}{Hidden technical debt in machine learning systems},
  in: \bibinfo{editor}{Cortes, C.}, \bibinfo{editor}{Lawrence, N.},
  \bibinfo{editor}{Lee, D.}, \bibinfo{editor}{Sugiyama, M.},
  \bibinfo{editor}{Garnett, R.} (Eds.), \bibinfo{booktitle}{Advances in Neural
  Information Processing Systems}, \bibinfo{publisher}{Curran Associates,
  Inc.}. pp. \bibinfo{pages}{1--9}.
\newblock \URLprefix
  \url{https://proceedings.neurips.cc/paper/2015/file/86df7dcfd896fcaf2674f757a2463eba-Paper.pdf}.
%Type = Article
\bibitem[{Seshia and Sadigh(2016)}]{SeshiaTowards2016}
\bibinfo{author}{Seshia, S.A.}, \bibinfo{author}{Sadigh, D.},
  \bibinfo{year}{2016}.
\newblock \bibinfo{title}{Towards verified artificial intelligence}.
\newblock \bibinfo{journal}{CoRR} \bibinfo{volume}{abs/1606.08514}.
\newblock \URLprefix \url{http://arxiv.org/abs/1606.08514},
  \href{http://arxiv.org/abs/1606.08514}{\tt arXiv:1606.08514}.
%Type = Techreport
\bibitem[{Seto et~al.(2000)Seto, Ferriera and Marz}]{SetoCaseStudy2000}
\bibinfo{author}{Seto, D.}, \bibinfo{author}{Ferriera, E.},
  \bibinfo{author}{Marz, T.}, \bibinfo{year}{2000}.
\newblock \bibinfo{title}{Case Study: Development of a Baseline Controller for
  Automatic Landing of an F-16 Aircraft Using Linear Matrix Inequalities
  (LMIs)}.
\newblock \bibinfo{type}{Technical Report} \bibinfo{number}{CMU/SEI-99-TR-020}.
  Software Engineering Institute, Carnegie Mellon University.
  \bibinfo{address}{Pittsburgh, PA}.
%Type = Inproceedings
\bibitem[{Seto et~al.(1998)Seto, Krogh, Sha and Chutinan}]{SetoSimplex}
\bibinfo{author}{Seto, D.}, \bibinfo{author}{Krogh, B.}, \bibinfo{author}{Sha,
  L.}, \bibinfo{author}{Chutinan, A.}, \bibinfo{year}{1998}.
\newblock \bibinfo{title}{The simplex architecture for safe online control
  system upgrades}, in: \bibinfo{booktitle}{Proceedings of the 1998 American
  Control Conference. ACC (IEEE Cat. No.98CH36207)}, pp.
  \bibinfo{pages}{3504--3508 vol.6}.
\newblock \DOIprefix\doi{10.1109/ACC.1998.703255}.
%Type = Inproceedings
\bibitem[{Simonyan and Zisserman(2015)}]{SimonyanVeryDeep}
\bibinfo{author}{Simonyan, K.}, \bibinfo{author}{Zisserman, A.},
  \bibinfo{year}{2015}.
\newblock \bibinfo{title}{Very deep convolutional networks for large-scale
  image recognition}, in: \bibinfo{editor}{Bengio, Y.}, \bibinfo{editor}{LeCun,
  Y.} (Eds.), \bibinfo{booktitle}{3rd International Conference on Learning
  Representations, {ICLR} 2015, San Diego, CA, USA, May 7-9, 2015, Conference
  Track Proceedings}, pp. \bibinfo{pages}{1--14}.
\newblock \URLprefix \url{http://arxiv.org/abs/1409.1556}.
%Type = Article
\bibitem[{Sohlberg and Jacobsen(2008)}]{SOHLBERG200811415}
\bibinfo{author}{Sohlberg, B.}, \bibinfo{author}{Jacobsen, E.},
  \bibinfo{year}{2008}.
\newblock \bibinfo{title}{Grey box modelling branches and experiences}.
\newblock \bibinfo{journal}{IFAC Proceedings Volumes} \bibinfo{volume}{41},
  \bibinfo{pages}{11415--11420}.
\newblock \URLprefix
  \url{https://www.sciencedirect.com/science/article/pii/S1474667016408025},
  \DOIprefix\doi{https://doi.org/10.3182/20080706-5-KR-1001.01934}.
  \bibinfo{note}{17th IFAC World Congress}.
%Type = Article
\bibitem[{Sokolsky and Ro{\c{s}}u(2012)}]{Sokolsky2012}
\bibinfo{author}{Sokolsky, O.}, \bibinfo{author}{Ro{\c{s}}u, G.},
  \bibinfo{year}{2012}.
\newblock \bibinfo{title}{Introduction to the special issue on runtime
  verification}.
\newblock \bibinfo{journal}{Formal Methods in System Design}
  \bibinfo{volume}{41}, \bibinfo{pages}{233--235}.
\newblock \URLprefix \url{https://doi.org/10.1007/s10703-012-0174-0},
  \DOIprefix\doi{10.1007/s10703-012-0174-0}.
%Type = Inproceedings
\bibitem[{Thrun(2006)}]{DARPA_Challenge}
\bibinfo{author}{Thrun, S.}, \bibinfo{year}{2006}.
\newblock \bibinfo{title}{Winning the darpa grand challenge}, in:
  \bibinfo{editor}{F{\"u}rnkranz, J.}, \bibinfo{editor}{Scheffer, T.},
  \bibinfo{editor}{Spiliopoulou, M.} (Eds.), \bibinfo{booktitle}{Machine
  Learning: ECML 2006}, \bibinfo{publisher}{Springer Berlin Heidelberg},
  \bibinfo{address}{Berlin, Heidelberg}. pp. \bibinfo{pages}{4--4}.
%Type = Article
\bibitem[{{Tomlin} et~al.(2003){Tomlin}, {Mitchell}, {Bayen} and
  {Oishi}}]{Tomlin2003}
\bibinfo{author}{{Tomlin}, C.J.}, \bibinfo{author}{{Mitchell}, I.},
  \bibinfo{author}{{Bayen}, A.M.}, \bibinfo{author}{{Oishi}, M.},
  \bibinfo{year}{2003}.
\newblock \bibinfo{title}{Computational techniques for the verification of
  hybrid systems}.
\newblock \bibinfo{journal}{Proceedings of the IEEE} \bibinfo{volume}{91},
  \bibinfo{pages}{986--1001}.
%Type = Article
\bibitem[{Tran et~al.(2019a)Tran, Cai, Diego, Musau, Johnson and
  Koutsoukos}]{Tran2019}
\bibinfo{author}{Tran, H.D.}, \bibinfo{author}{Cai, F.},
  \bibinfo{author}{Diego, M.L.}, \bibinfo{author}{Musau, P.},
  \bibinfo{author}{Johnson, T.T.}, \bibinfo{author}{Koutsoukos, X.},
  \bibinfo{year}{2019}a.
\newblock \bibinfo{title}{Safety verification of cyber-physical systems with
  reinforcement learning control}.
\newblock \bibinfo{journal}{ACM Trans. Embed. Comput. Syst.}
  \bibinfo{volume}{18}.
\newblock \URLprefix \url{https://doi.org/10.1145/3358230},
  \DOIprefix\doi{10.1145/3358230}.
%Type = Inproceedings
\bibitem[{Tran et~al.()Tran, Nguyen, Hamilton, Xiang and
  Johnson}]{TranSimulation2019}
\bibinfo{author}{Tran, H.D.}, \bibinfo{author}{Nguyen, L.V.},
  \bibinfo{author}{Hamilton, N.}, \bibinfo{author}{Xiang, W.},
  \bibinfo{author}{Johnson, T.T.}, .
\newblock \bibinfo{title}{Reachability analysis for high-index linear
  differential algebraic equations}, in: \bibinfo{booktitle}{Formal Modeling
  and Analysis of Timed Systems}, \bibinfo{publisher}{Springer-Verlag},
  \bibinfo{address}{Berlin, Heidelberg}. p. \bibinfo{pages}{160–177}.
\newblock \URLprefix \url{https://doi.org/10.1007/978-3-030-29662-9_10},
  \DOIprefix\doi{10.1007/978-3-030-29662-9_10}.
%Type = Inproceedings
\bibitem[{Tran et~al.(2019b)Tran, Nguyen, Musau, Xiang and Johnson}]{Tran2020}
\bibinfo{author}{Tran, H.D.}, \bibinfo{author}{Nguyen, L.V.},
  \bibinfo{author}{Musau, P.}, \bibinfo{author}{Xiang, W.},
  \bibinfo{author}{Johnson, T.T.}, \bibinfo{year}{2019}b.
\newblock \bibinfo{title}{Decentralized real-time safety verification for
  distributed cyber-physical systems}, in: \bibinfo{editor}{P{\'e}rez, J.A.},
  \bibinfo{editor}{Yoshida, N.} (Eds.), \bibinfo{booktitle}{Formal Techniques
  for Distributed Objects, Components, and Systems},
  \bibinfo{publisher}{Springer International Publishing},
  \bibinfo{address}{Cham}. pp. \bibinfo{pages}{261--277}.
%Type = Article
\bibitem[{Urban and Min{\'{e}}(2021)}]{UrbanFormalMethodsML2021}
\bibinfo{author}{Urban, C.}, \bibinfo{author}{Min{\'{e}}, A.},
  \bibinfo{year}{2021}.
\newblock \bibinfo{title}{A review of formal methods applied to machine
  learning}.
\newblock \bibinfo{journal}{CoRR} \bibinfo{volume}{abs/2104.02466}.
\newblock \URLprefix \url{https://arxiv.org/abs/2104.02466},
  \href{http://arxiv.org/abs/2104.02466}{\tt arXiv:2104.02466}.
%Type = Article
\bibitem[{Walsh and Karaman(2017)}]{Walsh2017}
\bibinfo{author}{Walsh, C.H.}, \bibinfo{author}{Karaman, S.},
  \bibinfo{year}{2017}.
\newblock \bibinfo{title}{{CDDT:} fast approximate 2d ray casting for
  accelerated localization}.
\newblock \bibinfo{journal}{CoRR} \bibinfo{volume}{abs/1705.01167}.
\newblock \URLprefix \url{http://arxiv.org/abs/1705.01167},
  \href{http://arxiv.org/abs/1705.01167}{\tt arXiv:1705.01167}.
%Type = Article
\bibitem[{Wilhelm et~al.(2008)Wilhelm, Engblom, Ermedahl, Holsti, Thesing,
  Whalley, Bernat, Ferdinand, Heckmann, Mitra, Mueller, Puaut, Puschner,
  Staschulat and Stenstr\"{o}m}]{Reinhard2008}
\bibinfo{author}{Wilhelm, R.}, \bibinfo{author}{Engblom, J.},
  \bibinfo{author}{Ermedahl, A.}, \bibinfo{author}{Holsti, N.},
  \bibinfo{author}{Thesing, S.}, \bibinfo{author}{Whalley, D.},
  \bibinfo{author}{Bernat, G.}, \bibinfo{author}{Ferdinand, C.},
  \bibinfo{author}{Heckmann, R.}, \bibinfo{author}{Mitra, T.},
  \bibinfo{author}{Mueller, F.}, \bibinfo{author}{Puaut, I.},
  \bibinfo{author}{Puschner, P.}, \bibinfo{author}{Staschulat, J.},
  \bibinfo{author}{Stenstr\"{o}m, P.}, \bibinfo{year}{2008}.
\newblock \bibinfo{title}{The worst-case execution-time problem—overview of
  methods and survey of tools}.
\newblock \bibinfo{journal}{ACM Trans. Embed. Comput. Syst.}
  \bibinfo{volume}{7}.
\newblock \DOIprefix\doi{10.1145/1347375.1347389}.
%Type = Article
\bibitem[{Xiang et~al.(2018)Xiang, Musau, Wild, Lopez, Hamilton, Yang,
  Rosenfeld and Johnson}]{xiang20118survey}
\bibinfo{author}{Xiang, W.}, \bibinfo{author}{Musau, P.},
  \bibinfo{author}{Wild, A.A.}, \bibinfo{author}{Lopez, D.M.},
  \bibinfo{author}{Hamilton, N.}, \bibinfo{author}{Yang, X.},
  \bibinfo{author}{Rosenfeld, J.A.}, \bibinfo{author}{Johnson, T.T.},
  \bibinfo{year}{2018}.
\newblock \bibinfo{title}{Verification for machine learning, autonomy, and
  neural networks survey}.
\newblock \bibinfo{journal}{CoRR} \bibinfo{volume}{abs/1810.01989}.
\newblock \href{http://arxiv.org/abs/1810.01989}{\tt arXiv:1810.01989}.
%Type = Article
\bibitem[{Xiang et~al.(2021)Xiang, Tran, Yang and Johnson}]{Xiang2021TNLS}
\bibinfo{author}{Xiang, W.}, \bibinfo{author}{Tran, H.D.},
  \bibinfo{author}{Yang, X.}, \bibinfo{author}{Johnson, T.T.},
  \bibinfo{year}{2021}.
\newblock \bibinfo{title}{Reachable set estimation for neural network control
  systems: A simulation-guided approach}.
\newblock \bibinfo{journal}{IEEE Transactions on Neural Networks and Learning
  Systems} \bibinfo{volume}{32}, \bibinfo{pages}{1821--1830}.
\newblock \DOIprefix\doi{10.1109/TNNLS.2020.2991090}.
%Type = Inproceedings
\bibitem[{Yang et~al.(2017)Yang, Islam, Murthy, Smolka and Stoller}]{Yang2017}
\bibinfo{author}{Yang, J.}, \bibinfo{author}{Islam, M.A.},
  \bibinfo{author}{Murthy, A.}, \bibinfo{author}{Smolka, S.A.},
  \bibinfo{author}{Stoller, S.D.}, \bibinfo{year}{2017}.
\newblock \bibinfo{title}{A simplex architecture for hybrid systems using
  barrier certificates}, in: \bibinfo{editor}{Tonetta, S.},
  \bibinfo{editor}{Schoitsch, E.}, \bibinfo{editor}{Bitsch, F.} (Eds.),
  \bibinfo{booktitle}{Computer Safety, Reliability, and Security},
  \bibinfo{publisher}{Springer International Publishing},
  \bibinfo{address}{Cham}. pp. \bibinfo{pages}{117--131}.
%Type = Article
\bibitem[{Yilmaz et~al.(2006)Yilmaz, Javed and Shah}]{YilmazObjectTracking}
\bibinfo{author}{Yilmaz, A.}, \bibinfo{author}{Javed, O.},
  \bibinfo{author}{Shah, M.}, \bibinfo{year}{2006}.
\newblock \bibinfo{title}{Object tracking: A survey}.
\newblock \bibinfo{journal}{ACM Comput. Surv.} \bibinfo{volume}{38},
  \bibinfo{pages}{13–es}.
\newblock \DOIprefix\doi{10.1145/1177352.1177355}.
%Type = Article
\bibitem[{Yurtsever et~al.(2020)Yurtsever, Lambert, Carballo and
  Takeda}]{Yurtsever2019}
\bibinfo{author}{Yurtsever, E.}, \bibinfo{author}{Lambert, J.},
  \bibinfo{author}{Carballo, A.}, \bibinfo{author}{Takeda, K.},
  \bibinfo{year}{2020}.
\newblock \bibinfo{title}{A survey of autonomous driving: Common practices and
  emerging technologies}.
\newblock \bibinfo{journal}{IEEE Access} \bibinfo{volume}{8},
  \bibinfo{pages}{58443--58469}.
\newblock \DOIprefix\doi{10.1109/ACCESS.2020.2983149}.

\end{thebibliography}



\end{document}